\documentclass[11pt,a4paper]{article}
\usepackage[T1]{fontenc}
\usepackage[utf8]{inputenc}
\usepackage{lmodern}
\usepackage{microtype}
\usepackage[margin=2.5cm]{geometry}
\usepackage{amsmath,amssymb}
\usepackage{graphicx}
\usepackage{booktabs}
\usepackage{longtable}
\usepackage{array}
\usepackage{calc}
\usepackage{caption}
\usepackage[table]{xcolor}
\usepackage{textcomp}
\usepackage{CJKutf8}
\usepackage[numbers,sort&compress]{natbib}
\usepackage[hidelinks]{hyperref}
\hypersetup{
  pdftitle  = {Phoneme-identity depth profiles do not reliably predict LoRA placement for dysarthric ASR},
  pdfauthor = {Bernard Muller and LaVonne Roberts},
  pdfsubject= {Parameter-efficient adaptation of ASR foundation models to dysarthric speech},
  pdfkeywords = {dysarthric speech recognition, LoRA, adapter placement, layer-wise probing, Qwen3-ASR, Whisper}
}
\DeclareUnicodeCharacter{2212}{\ensuremath{-}}
\DeclareUnicodeCharacter{2248}{\ensuremath{\approx}}
\DeclareUnicodeCharacter{2264}{\ensuremath{\leq}}
\DeclareUnicodeCharacter{2265}{\ensuremath{\geq}}
\DeclareUnicodeCharacter{03B1}{\ensuremath{\alpha}}
\DeclareUnicodeCharacter{0394}{\ensuremath{\Delta}}
\DeclareUnicodeCharacter{2113}{\ensuremath{\ell}}
\DeclareUnicodeCharacter{2713}{\ensuremath{\checkmark}}
\DeclareUnicodeCharacter{2075}{\textsuperscript{5}}
\DeclareUnicodeCharacter{2076}{\textsuperscript{6}}

\newcounter{none}

\providecommand{\tightlist}{\setlength{\itemsep}{0pt}\setlength{\parskip}{0pt}}

\title{\textbf{Phoneme-identity depth profiles do not reliably predict LoRA placement for dysarthric ASR}}
\author{%
  Bernard Muller\textsuperscript{1,2} \and LaVonne Roberts\textsuperscript{1,2}\\[2pt]
  \small\textsuperscript{1}The Scott-Morgan Foundation, United Kingdom\\
  \small\textsuperscript{2}SMF Labs, 28 rue du Colonel Pierre Avia, 75015 Paris, France\\[2pt]
  \small Correspondence: \texttt{lavonne@scottmorganfoundation.org}
}
\date{}

\begin{document}
\maketitle

\begin{abstract}
Parameter-efficient adaptation of a speech foundation model to dysarthric speech requires a rarely
justified choice: which encoder layers to adapt. A recently proposed strategy is to probe the encoder,
find where phonetic information becomes most decodable, and put the adapters there. We test that
inference across two ASR models (Qwen3-ASR, Whisper-small) and two dysarthric corpora (Mandarin CDSD,
English UASpeech), holding recipe and seed discipline constant throughout, and split and gate constant
within each corpus.

\textbf{The inference fails.} Measured under one pipeline rather than inherited from published figures, all
four phoneme-error curves minimise between 83\% and 92\% of relative depth, each model's minimising layer
moving no further between corpora than between probe seeds. The best adaptation band is not similarly
constant. None of three natural mappings from the curve to a band --- steepest improvement, lowest
band-mean error, or knee location --- recovers the best tested configuration. We do not claim no statistic
of the curve could work, only that these three do not.

\textbf{Placement still matters, but is model- and corpus-specific.} On CDSD the models disagree --- Qwen3-ASR
prefers its upper third (29.88\% CER, three-seed mean), Whisper-small all-block LoRA (53.56\%) --- while on
UASpeech neither favours the upper band. Retraining the CDSD sweep at three seeds leaves upper-versus-all
resolved (−1.31 pp, CI [−2.16, −0.25]) but not upper-versus-middle (−0.83 pp, CI [−2.06, +1.86]) --- a gap
of that size is within what a training seed moves. The two models' preferences differ as a directly
tested interaction (−4.71 pp, CI [−6.17, −2.89]). We designate this the primary claim by its role
--- it is what our first contribution asserts and the question the paper sets out to answer --- and we
record that the designation was made after an earlier version nominated a different contrast
(§6.1.1). It resolves under micro CER, under speaker-equal weighting and under an exact paired test
alike (unadjusted exact p = 0.031); Qwen3-ASR's own preference holds under micro-averaging only. A budget-matched control
finds no statistically resolved benefit from additional capacity for either placement, but does not
establish that capacity is irrelevant: at matched budget the placement contrast itself is not
resolvable at this sample size, and the resolved result is the equal-rank one. On Qwen3-ASR × UASpeech, adapting
near the input is reliably worse than not adapting at all (+9.90 pp CER). We therefore recommend choosing
the band by a short adaptation sweep on the model and corpus in hand, not by reading it off a probe.
\end{abstract}

\section{1. Introduction}\label{introduction}

Automatic speech recognition has become usable for typical speech and remains unusable for many people
with dysarthria. The dominant remedy is adaptation: take a large pre-trained model and fine-tune it, in
whole or in part, on whatever disordered speech is available. Because that data is scarce and because
per-speaker adaptation is increasingly the deployment model, parameter-efficient methods --- low-rank
adapters (LoRA; \citep{hu2022lora}) in particular --- have become common. The dysarthric-ASR literature bears
this out: \citet{jiang2024perceiverprompt} fine-tune Whisper with LoRA before adding speaker prompts for Chinese
disordered speech; \citet{tan2025cbawhisper} use curriculum-scheduled AdaLoRA for low-resource dysarthric
recognition; \citet{wagner2025personalized} report AdaLoRA outperforming both full fine-tuning and standard LoRA for
per-speaker personalisation; and \citet{hu2025moespeaker} route between mixture-of-experts speaker adapters on
UASpeech. \textbf{What none of them settles is placement.} \citet{hu2025moespeaker} is the closest to our question and
still orthogonal to it: it selects \emph{which expert} adapts a given speaker, not \emph{where in the encoder} the
adaptation sits. Each inherits or searches an adapter configuration without measuring where the relevant
information sits, which makes a design question unavoidable: \textbf{where in the encoder should the adapter
go?}

The question invites an appeal to mechanism: if one can measure where in an encoder's depth the phonetic
information relevant to the target domain is formed, one ought to be able to place adapters there. \citet{muthu2026layerwise} made exactly this argument for dysarthric ASR. Probing a 12-block Whisper-small encoder on
Mandarin dysarthric speech, they found that phoneme error rate drops steeply through the early-to-middle
layers and plateaus after layer 6, inferred that ``these layers should be the most productive targets for
dysarthric adaptation'', and confirmed it: a single adapter at layer 7 and the 5--8 band both land close to
full-encoder adaptation, while their upper band 9--12 is the weaker contiguous group. (Their v2 reports 68.64 and 66.21 CER
against 64.35 for all blocks, i.e.~6.67\% and 2.89\% relative. Their v1 abstract gave 3.5\% and 2.48\%, which
did not follow from its own table; the v2 abstract corrects this. \textbf{We cite v2 throughout} --- it also
revised the unadapted baseline from 105.08 to 89.98 CER, which matters for §5.) That is a clean result, and it invites a reading we think is unsafe: \emph{dysarthric
adaptation belongs mid-encoder}.

\textbf{The confound is that a layer index means different things in different encoders.} Depth, time
resolution, training objective and what sits downstream all vary across ASR model families. An encoder
feeding a Whisper-style text decoder is under different pressure at its top than an encoder feeding a
projector into a large language model. If the depth at which phonetic information matures differs, the
band where adaptation is productive should differ with it --- and layer-placement advice inherited from the
literature would be actively misleading.

\textbf{That was our hypothesis, and the second half of it turned out to be wrong.} Placement advice \emph{is}
model-specific on the Mandarin corpus: swapping the encoder while holding everything else fixed changes
which band wins. It is not model-specific everywhere --- running the same contrast in all four cells of a
model × corpus design shows the two models agreeing on English isolated words (§6.8, §6.9), so the
difference is a property of one corpus rather than of the two architectures. But the proposed mechanism ---
differing maturation depth --- does not survive measurement in either case. Measured under our own pipeline
rather than cited, Whisper-small's phoneme-error minimum lands at 83\% of relative depth rather than the
50\% the published curve suggests, essentially where Qwen3-ASR's sits, so the depth profile cannot be what
selects the band.

We test this along two orthogonal axes. \textbf{First, hold the data constant and change the encoder.} We use
the same corpus as Muthu et al.~--- the Chinese Dysarthria Speech Database (CDSD; \citep{wan2024cdsd}) --- and
probe the 24-block audio encoder of Qwen3-ASR-1.7B \citep{shi2026qwen3asr}, whose decoder is a language model
rather than a Whisper-style text decoder. Differences in the depth profile are therefore attributable to
the encoder rather than to corpus, language or task --- though not to any single architectural factor, since
the two systems differ simultaneously in depth, frame stride, pre-training data and objective, scale and
decoder type. \textbf{Second, hold the encoder constant and change the corpus.} We repeat the identical
adaptation sweep on \textbf{UASpeech} \citep{kim2008uaspeech}, which differs from CDSD on every axis available to
us: English rather than Mandarin, 449 isolated words rather than read news sentences, and clinical
severity labels rather than CER-derived pseudo-labels. Both depth probes are run on \textbf{both} corpora
(§5.5, §5.6), so the second axis tests the mechanism proposed to explain the ordering and not only the
prediction's outcome --- and finds that the probe curves barely move while the placement outcome does.

\textbf{Two things this paper is not.} It is not a claim that one foundation model is better than another for
dysarthric speech: that comparison is not run here, and the one published head-to-head we are aware of
\citep{huber2026adapting} reaches the opposite conclusion for adaptability, on a single speaker (§2). Nor is it
a claim that probing-before-adapting is a new idea --- \citet{pasad2023comparative}, \citet{beheshti2026probingguided} and \citet{muthu2026layerwise} all state some form of it. Our contribution is narrower: \textbf{a controlled demonstration that a
LoRA placement result obtained on one dysarthric ASR model cannot safely be transferred to another, and
evidence that three natural phoneme-PER heuristics fail to explain the placement difference we observe.}

\textbf{Contributions.}

\begin{enumerate}
\def\labelenumi{\arabic{enumi}.}
\item
  \textbf{A same-corpus, same-recipe, cross-architecture adaptation comparison showing that optimal adapter
  placement differs between these two models on one of the two corpora tested.} Holding corpus, split,
  integrity gate, LoRA recipe, per-model parameter budget and seed discipline constant and swapping the
  model, CDSD resolves the \texttt{upper} − \texttt{all} contrast in \textbf{opposite} directions, and the interaction
  between them is resolved under every estimator we apply (§6.9.1) --- though Qwen3-ASR's own main effect
  holds under micro-averaged CER and not under speaker-equal weighting (§6.10): Qwen3-ASR is best served
  by its upper band, which beats adapting all encoder blocks at one third the parameters, whereas
  Whisper-small is best served by all-block LoRA and ranks \texttt{upper} worst of the four (§6.1, §6.6). The prior
  literature could not make this comparison, because it was run across papers rather than within one.
  \textbf{Running the same contrast in all four cells of a model × corpus design shows the reversal is specific
  to CDSD:} on UASpeech neither model favours \texttt{upper} (§6.8, §6.9). A placement × capacity 2×2 finds no resolved benefit from total
  budget (§6.7): tripling either placement's rank changes its CER by no
  resolved margin.
\item
  \textbf{A negative result: three phoneme-identity PER heuristics fail to predict the best LoRA band across
  two ASR models and two corpora.} Only phoneme-identity PER is tested as the placement signal; a frozen
  CTC readout asks whether phone identity is \emph{recoverable by a CTC readout} at a layer, whereas LoRA
  adaptation asks which perturbation location most improves a generative sequence model, so the result is
  best read as: \textbf{CTC-readout recoverability of phone identity is not equivalent to local adaptation
  utility.} Measured under our own pipeline rather than cited, Whisper-small's phoneme-error minimum
  sits at 83\% of relative depth and Qwen3-ASR's at 83--92\% --- both late, yet the two models prefer
  different bands. \emph{Adapt where PER improves most steeply} is contradicted on Qwen3-ASR, where the
  steepest band is the worst target; \emph{adapt where band-mean PER is lowest} selects the winning target in
  one of four cells; and \emph{adapt where the knee sits} --- a reconciliation we
  proposed ourselves before measuring the second architecture --- predicts \texttt{upper} for Whisper-small,
  where it ranks worst of the four. \textbf{We withdraw that
  rule on the strength of our own measurement.} Repeating both probes on UASpeech (§5.6) closes the
  alternative explanation that the curve is corpus-dependent: all four minima fall between \textbf{83\% and
  92\%} of relative depth, all four order the three proportional bands identically, and each model's
  argmin moves no further between corpora than between probe seeds within one.
\item
  \textbf{A second-corpus replication that separates what generalises from what does not.} Two contrasts hold
  with intervals excluding zero on \emph{both} corpora --- \texttt{upper} outperforms \texttt{lower}, and \texttt{middle} outperforms
  \texttt{lower}. A third, \texttt{upper} over \texttt{middle}, resolves on UASpeech but not on CDSD once training seeds are
  resampled alongside speakers (§6.1.1), while pointing the same way in every run; we report it as
  directional. No contrast is resolved in opposite directions on the two corpora. \texttt{upper}-versus-\texttt{all} is the one that does not carry: UASpeech resolves
  it in neither direction, so it and the parameter-matched 2×2 built on it are reported as CDSD-only. The
  second corpus also settles a question the first could not: on CDSD no adaptation-versus-baseline
  contrast reached significance, whereas on UASpeech all four do, establishing that adaptation helps (up
  to −9.20 pp CER) and that on this model and corpus lower-band adaptation is \textbf{reliably worse than not adapting at all}
  (+9.90 pp CER). Three matched seeds per band put seed variance at 0.12--0.38 pp, well inside
  the speaker-sampling uncertainty.
\item
  \textbf{Two methodological results about layer probing itself, which explain how the discarded claim arose.}
  \textbf{Probe-readout sensitivity to training volume is depth-dependent and architecture-specific} (at a
  fixed epoch budget, so volume and optimiser steps move together --- §5.5): increasing probe
  training data 5.4× (6,200 → 33,617 utterances) improves Whisper-small's deepest block by 9.15 pp,
  1.87× its own stack mean, against 2.25 pp and 0.74× for Qwen3-ASR's. A low-volume probe therefore
  manufactures an apparent late-stack difference between the two architectures --- 6.56 against 1.85 pp of
  post-minimum rise --- that shrinks to 2.07 against 1.82 at matched volume. And \textbf{we cannot reproduce the
  mid-stack plateau reported for Whisper-small}: probe configuration matches the prior work item by
  item, and the discrepancy survives both padding treatments and matched training volume. We can no
  longer rule out unadapted baseline competence as a contributor: the prior work's v2 reports 89.98 CER
  against our corrected 77.57 (§5). We report the discrepancy as unexplained, with competence among the
  live candidates.
\item
  \textbf{An architectural precondition for the method, reported as a finding rather than a gap.} The
  Qwen3-ASR audio encoder emits one frame per \textasciitilde77 ms against Whisper-small's 20 ms, and against this
  corpus's forced alignments (median phone 80 ms) 49.4\% of phones are shorter than a single encoder frame,
  versus 1.3\% at 20 ms. We therefore do not compare frame-level boundary F1 across the two encoders: that
  would confound representational information with quantisation imposed by the output grid. We do \textbf{not}
  claim the representation cannot encode sub-frame timing; what is limited is where a framewise probe can
  place a boundary.
\item
  \textbf{A documented failure mode of shallow adaptation.} Damage from adapting near the input is
  concentrated on acoustically atypical speakers, and the errors are \textbf{consistent with} language-prior
  takeover --- a mechanistic hypothesis rather than a demonstrated causal pathway, since the decoder is
  frozen and adaptation changes its conditioning representation rather than any language-model weights:
  the adapted model emits fluent, corpus-typical text unrelated to the audio while the unadapted model
  transcribes correctly. We also report a negative result that bounds how far this generalises.
\end{enumerate}

\begin{center}\rule{0.5\linewidth}{0.5pt}\end{center}

\section{2. Related work}\label{related-work}

\subsection{2.1 Layer-wise analysis of speech representations}\label{layer-wise-analysis-of-speech-representations}

\citet{pasad2021layerwise} established layer-wise analysis of self-supervised speech encoders, showing an
autoencoder-like behaviour in wav2vec 2.0 in which middle layers carry the most linguistic content while
top layers move back toward the input. Their follow-up \citep{pasad2023comparative} states two things this paper
tests in a new setting: that layer-wise properties ``evolve across layers differently depending on the
model'', and that layer-wise analysis ``provide[s] reliable guidance to choose layers of interest for
downstream tasks''. \textbf{What has not been tested is whether they hold for the choice of \emph{adaptation} depth in
a supervised ASR foundation encoder under pathological speech.} \citet{ashihara2024crossmodel} perform the
analogous cross-model-family comparison for self-supervised speech and speaker models, and are the
methodological precedent for holding data constant while varying the encoder.

Two recent studies probe layer-wise structure on atypical speech. \citet{yue2026probing} probe Whisper-Medium
for dysarthria detection and severity assessment, locating the most informative representations around
layers 13--15 and finding that fine-tuning induces only modest layer-wise change; \citet{mojarad2026ccr}
probe wav2vec 2.0 and Whisper for consonant-cluster reduction in African American English. Neither
connects the probe to adapter placement.

Mechanistic-interpretability work on ASR converges on the same region of the model from a different
direction. \citet{glazer2026beyond} find contextual bias emerging \emph{within the encoder} and overriding acoustic
evidence --- closely related to the failure mode we document in §7.2; \citet{pluth2026sae} report that causal
steering of Whisper encoder features ``is more reliable for higher-level features than lower-level ones'';
and \citet{ma2026behind} describe a ``delayed specialisation'' in a LoRA-adapted Whisper. All three are
consistent with the direction of our adaptation result and are not, individually, evidence for it.

\subsection{2.2 Layer-wise probing tied to adapter placement}\label{layer-wise-probing-tied-to-adapter-placement}

\citet{muthu2026layerwise} is the direct antecedent of this paper and the work we contradict in part. They probe
a Whisper-small encoder on CDSD under three transcript-matched conditions, report a task-dependent
hierarchy across depth, and use it to motivate mid-band adapter placement, which their sweep over single
layers and the bands 1--4 / 5--8 / 9--12 / all then confirms. Their setting is a broken-baseline regime:
unadapted Whisper-small reaches 89.98\% CER on dysarthric CDSD and 64.35\% after full-encoder LoRA
(their v2; v1 reported 105.08 and 68.85 respectively). We
reuse their probe designs, adapter configuration and band structure unchanged, so that what differs
between the two studies is the encoder and not the protocol --- and we \textbf{re-ran their architecture rather
than only citing it}: both the depth probe (§5.5) and the four-band sweep (§6.6, §6.8) were repeated on
Whisper-small under our own corpus, split, gate, recipe and scoring, which converts the paper's central
comparison from cross-paper to internally controlled. Their probe result does not reproduce under our
pipeline, for reasons we have been unable to identify (§5.5), while their band result partially does.

\subsection{2.3 Where to adapt, outside dysarthria --- an established and contested question}\label{where-to-adapt-outside-dysarthria-an-established-and-contested-question}

The general proposition that the middle is not automatically the right place to adapt is \textbf{not novel},
and the literature does not agree with itself.

\begin{itemize}
\tightlist
\item
  \textbf{A U-shape.} DAMA \citep{xiao2026dama} studies depth-aware adaptation for multilingual ASR in
  low-resource languages and reports that early \emph{and} late layers are language-specific and need
  adaptation while intermediate layers retain shared structure, reaching an 80\% reduction in trainable
  parameters at equal or better accuracy across 18 languages.
\item
  \textbf{Layer-selective LoRA in ASR already exists.} \citet{han2024lalora} apply LoRA to an automatically
  selected subset of layers of a large ASR model for low-resource multilingual adaptation, establishing
  the general move --- do not adapt every layer --- in ASR specifically. It does not address which depth suits
  \emph{pathological} speech, nor tie the choice to a measured phonetic-maturity profile, but it removes
  ``selective placement in ASR'' from anything we can claim as novel.
\item
  \textbf{Probing-guided selection.} \citet{beheshti2026probingguided} rank layers with cheap probes \emph{before any
  task-specific model is trained}, reporting that informative layers ``cluster in depth zones rather than
  at uniquely optimal positions'', with ``backbone-specific selections rather than a fixed layer recipe''.
  They select layers to \emph{read from} rather than to \emph{adapt}, but this is the closest existing statement of
  the principle we test.
\item
  \textbf{The contrary result.} \citet{lin2024peftspeech} searched layer-wise PEFT placement in self-supervised speech
  encoders and found it \textbf{did not} beat uniform insertion into all layers. Our finding that \texttt{upper} beats
  \texttt{all} on CDSD × Qwen3-ASR --- at one third the parameter count (§6.1) ---
  is a counterexample in a different model family, and the disagreement is itself evidence for this
  paper's thesis.
\item
  \textbf{Transfer from other modalities.} \citet{ogawa2026layerwiselora} select LoRA layers by CKA similarity and halve
  trainable parameters at parity across encoder-only, decoder-only and multimodal models; \citet{xu2026layercard}
  propose a theory-driven ``Layer Card'' diagnostic for placement in large language models.
\item
  \textbf{Learning the layer choice instead of reading it off a probe.} Gumbel-BEARD \citep{wang2026gumbelbeard}
  replaces manual layer search with an end-to-end hard Gumbel-Softmax selector over Whisper encoder
  layers, trained against a BEST-RQ self-supervised objective, and reports that the selected depth varies
  by domain. It is the closest contemporary alternative to this paper's practical recommendation: where we
  argue for a short sweep because a phoneme probe does not transfer, that work argues for learning the
  choice from the adaptation objective itself. Both share the premise the paper defends --- that the useful
  depth is a property of the model and the task, not something readable from a generic diagnostic.
\item
  \textbf{Encoder depth and LoRA in ASR.} \citet{kolluri2026encoderdepth} prune Whisper encoder layers and combine
  pruning with LoRA, finding that LoRA partly compensates for removed encoder capacity via language-model
  priors, less well in low-resource settings --- a close relative of the failure mode in §7.2.
\item
  \textbf{Steering the encoder rather than the decoder.} SALSA \citep{yegorova2026salsa} reports that steering a
  speech encoder, ``particularly the later layers, is more effective than steering the LLM backbone''.
\end{itemize}

\subsection{2.4 Foundation-model choice for dysarthric speech}\label{foundation-model-choice-for-dysarthric-speech}

\citet{huber2026adapting} compare Whisper-large-v3 against Qwen3-ASR-1.7B as bases for dysarthric adaptation
and conclude in favour of Whisper, reporting 9.7\% WER with all available adaptation data against 14--16\%
for Qwen3-ASR fine-tuning and LoRA, and noting that LoRA underperformed full fine-tuning; their stated
hypothesis is that dysarthric speech is too far from the base distribution for low-rank capacity to
absorb. That study is \textbf{n = 1 speaker}, with 92 h of read speech plus 8.8 h of in-the-wild corrections
from a single user, and it measures adaptability under generous per-speaker data rather than
representational structure or adapter placement.

We record it because a reader will encounter it, and because it cuts both ways. It is a reason not to
phrase any part of this work as a claim about which base model is better --- we make no such claim. But it
does not undercut what we do claim: our between-band contrasts are internal to one encoder and hold
whatever the encoder's absolute standing, and our own result that no adapted variant differs significantly
from the unadapted model on the CDSD test set (§6.1) is if anything consistent with their scepticism.

\subsection{2.5 Corpus and tooling}\label{corpus-and-tooling}

CDSD \citep{wan2024cdsd} is the Mandarin dysarthric corpus used here and by Muthu et al.~Alignments are
produced with the Montreal Forced Aligner \citep{mcauliffe2017mfa}; the phoneme-identity probe is a CTC
probe \citep{graves2006ctc}; the adapters are LoRA \citep{hu2022lora}.

\begin{center}\rule{0.5\linewidth}{0.5pt}\end{center}

\section{3. Encoder geometry: a precondition, and a finding}\label{encoder-geometry-a-precondition-and-a-finding}

Before any probe is defined, the encoder's time resolution has to be measured against the phenomenon
being probed. This step is normally silent, and on this architecture it decides which probes are
admissible.

The Qwen3-ASR technical report \citep{shi2026qwen3asr} specifies 8× downsampling of 128-dimensional filterbank
features, a nominal token rate of 12.5 Hz. Measured directly across five CDSD utterances spanning
0.40--9.40 s, the released encoder emits \textbf{12.5--13.2 frames per second, i.e.~\textasciitilde77 ms per frame}; we use the
measured value throughout, and the nominal 80 ms would make every figure below slightly more extreme.
Whisper-small, the encoder probed by Muthu et al., runs at \textbf{20 ms} \citep{radford2023whisper}, so the
Qwen3-ASR encoder is roughly \textbf{4× coarser in time}, and 2× deeper (24 blocks against 12) with d\_model
1024 against 768. The encoder returns 25 hidden states: index 0 is the post-convolutional encoder input
and 1--24 are block outputs, matching the layer-0-is-input convention of the prior work. Hidden state 0 is
already post-convolution, so no earlier tap exists at a finer resolution.

Against the corpus's forced alignments (21,251 phones, 2,640 utterances, 44 speakers):

\textbf{Table 1.} Phone-duration distribution in CDSD, and its relation to encoder frame period.

{\def\LTcaptype{none} % do not increment counter
\begin{longtable}[]{@{}ll@{}}
\toprule\noalign{}
Statistic & ms \\
\midrule\noalign{}
\endhead
\bottomrule\noalign{}
\endlastfoot
p5 / p10 / p25 & 30 / 30 / 30 \\
\textbf{median} & \textbf{80} \\
p75 / p90 / p95 & 140 / 230 / 300 \\
mean & 110.4 \\
\end{longtable}
}

{\def\LTcaptype{none} % do not increment counter
\begin{longtable}[]{@{}
  >{\raggedright\arraybackslash}p{(\linewidth - 6\tabcolsep) * \real{0.2500}}
  >{\raggedright\arraybackslash}p{(\linewidth - 6\tabcolsep) * \real{0.2500}}
  >{\raggedright\arraybackslash}p{(\linewidth - 6\tabcolsep) * \real{0.2500}}
  >{\raggedright\arraybackslash}p{(\linewidth - 6\tabcolsep) * \real{0.2500}}@{}}
\toprule\noalign{}
\begin{minipage}[b]{\linewidth}\raggedright
Frame period
\end{minipage} & \begin{minipage}[b]{\linewidth}\raggedright
Phones \textless{} 1 frame
\end{minipage} & \begin{minipage}[b]{\linewidth}\raggedright
Phones \textless{} 2 frames
\end{minipage} & \begin{minipage}[b]{\linewidth}\raggedright
Mean frames per phone
\end{minipage} \\
\midrule\noalign{}
\endhead
\bottomrule\noalign{}
\endlastfoot
20 ms (Whisper-small) & \textbf{1.3\%} & 28.6\% & 5.52 \\
\textbf{77 ms (Qwen3-ASR, measured)} & \textbf{49.4\%} & 77.8\% & \textbf{1.43} \\
\end{longtable}
}

\textbf{Consequence.} Frame-level phoneme-boundary detection --- ``is frame \emph{t} a boundary?'' --- is not a
measurement that can be made on this encoder. Half of all phones are shorter than the sampling interval,
so any F1 obtained would characterise the frame grid, not the representation. This is a different
statement from the prior work's finding that boundary information stays weak at every layer for
dysarthric speech; that finding is about a representation, and we cannot test it here. Nor is the figure
sensitive to the 77-versus-80 ms question, since the corpus's median phone duration is itself 80 ms.

We report the constraint because it is the clearest instance of the paper's general point: \textbf{a probing
protocol is not architecture-portable, and its low-level probes can silently stop being well-posed when
the encoder changes.} Anyone porting a layer-probing study to a new encoder should measure frame period
against the target phenomenon's duration distribution first. This matters beyond methodology: dysarthria
distorts sub-phonemic timing, so an encoder at this resolution is structurally unable to represent some of
the phenomena clinicians care about. The remaining probes are unaffected: phoneme identity via a CTC probe
is feasible (99.9\% of utterances satisfy T ≥ L, median T/L = 3.47), utterance-level difficulty needs no
time resolution, and the cross-condition and adaptation analyses are pooled.

\begin{center}\rule{0.5\linewidth}{0.5pt}\end{center}

\section{4. Experimental setup}\label{experimental-setup}

\subsection{4.1 Corpus and integrity gate}\label{corpus-and-integrity-gate}

The probing experiments of §5.1--§5.5 and the adaptation sweep of §6.1--§6.4 use CDSD. The replication
sweeps of §6.5 and §6.8 and the replicated depth probes of §5.6 use UASpeech \citep{kim2008uaspeech} under an
otherwise identical recipe; its corpus, split and scoring details are given in §6.5, and the two protocol
departures forced on the UASpeech probes in §5.6.

CDSD \citep{wan2024cdsd}: 44 dysarthric Mandarin speakers reading sentence material drawn from news text.
The corpus ships two subset directories: the \textbf{1h} subset holds 55,851 utterances from all 44 speakers
(35.06 h), the \textbf{10h} subset 92,730 utterances from 8 speakers (64.32 h). \textbf{6,103 (speaker, utterance-id)
pairs name one file in each subset, and in every one of those cases the two files carry different
transcripts.} The prior work reports \textasciitilde55,850 utterances, which identifies the 1h subset, and we use it for
the same reason it is the right choice independently: it retains all 44 speakers. \textbf{On corpus size}, the
two subsets sum to 99.38 h against the 133 h reported in the corpus publication; we have no account of the
difference, and all results here derive from the \textbf{1h subset only}.

Utterance identifiers are keyed \texttt{speaker/subset/utt\_id} throughout, and every retained row's alignment
transcript is verified against its manifest transcript, with mismatches dropped. Forced alignments were
produced with the Montreal Forced Aligner \citep{mcauliffe2017mfa}; utterances whose phone tier contains
no real phones are excluded, as are those with no TextGrid at all. Of the 55,851 utterances in the 1h
subset, \textbf{8,164 (14.6\%) are removed} --- 7,720 for having no real phones and 444 for a missing TextGrid ---
leaving 47,687.

\textbf{What the gate removes, and whether it makes the evaluation easier.} Forced-alignment failure is not
random in pathological speech, so an integrity gate can quietly select an easier evaluation population.
We measured this rather than assuming it away.

{\def\LTcaptype{none} % do not increment counter
\begin{longtable}[]{@{}lll@{}}
\toprule\noalign{}
Stratum & Excluded & Rate \\
\midrule\noalign{}
\endhead
\bottomrule\noalign{}
\endlastfoot
mild & 323 / 3,211 & 10.1\% \\
moderate & 3,256 / 22,778 & 14.3\% \\
severe & 4,585 / 29,862 & \textbf{15.4\%} \\
kept utterances & --- & mean duration \textbf{2.37 s} (median 2.06) \\
excluded utterances & --- & mean duration \textbf{1.59 s} (median 1.28) \\
\end{longtable}
}

\textbf{There is a severity gradient and it runs in the direction the concern predicts} --- 10.1\% of mild
material is removed against 15.4\% of severe --- so the retained set is modestly easier than the raw subset.
The gradient spans 5.3 pp across the full severity range, which we judge small relative to the effects in
§6 but do not dismiss. \textbf{The dominant factor is duration, not severity:} excluded utterances are on
average 0.78 s shorter than retained ones, and 94.6\% of exclusions are ``no real phones'', the signature of
very short or effectively empty recordings. Per-speaker exclusion rates range from 3.5\% to 32.5\%.

Because the gate is applied identically to every band and to the unadapted baseline, it does not
\emph{differentially assign} utterances by condition, so it cannot manufacture a between-band difference out of
nothing. \textbf{It does, however, change the population the contrasts are estimated on, and could alter their
magnitude if placement effects are heterogeneous in the characteristic being selected on} --- which is not
a hypothetical concern here, since §7 argues that shallow adaptation harms acoustically atypical speakers
specifically, and the gate removes short and somewhat more severe utterances at a higher rate. What the
gate certainly affects is the absolute CER level and any comparison against work using a different gate ---
including the cross-paper Whisper-small comparison, where the antecedent applied no equivalent filter.
LoRA training and CER scoring do not themselves require alignments, so re-running the key CDSD conditions
on the full valid-audio subset would settle this; we have not done it, and flag it as an open sensitivity
check (§9).

\textbf{Splits are speaker-disjoint: 31 train / 7 validation / 6 test speakers}, and the same partition is used
for the probes and for the adaptation sweep so that the two are commensurable. The probe dataset is 8,800
utterances (200 per speaker); the adaptation manifest is 47,687 utterances (33,617 train / 8,682
validation / 5,388 test), 14.6\% smaller than the prior work's because it inherits the gate.

\textbf{Tone units are derived from the transcript, not from the alignments.} The alignment phone labels retain
tone letters, but T1 (high level) does not occur once in 43,137 phone tokens --- a data defect rather than a
fact about Mandarin, since T1 is roughly a fifth of Mandarin syllables. Text-derived pinyin over the same
transcripts gives the expected distribution (T1 18.3\% / T2 19.3\% / T3 19.4\% / T4 33.2\% / neutral 9.9\%), so
we follow the prior work exactly: pinyin initial/final/tone units from text (\texttt{tong2} → \{t, ong, T2\}), with
alignments used only for segment timing.

\subsection{4.2 Model and layer indexing}\label{model-and-layer-indexing}

The probed encoder is the AuT audio tower of \textbf{Qwen3-ASR-1.7B} \citep{shi2026qwen3asr}, read from the
released Apache-2.0 checkpoint: 24 transformer blocks, d\_model 1024, 16 attention heads, FFN width 4096,
projected to 2048 for the language-model decoder. Following the prior work, probe layer 0 is the encoder
input and layers 1--24 are block outputs; adaptation bands are named by \textbf{block} index, 1-based, so
\texttt{upper} = blocks 17--24, and probe tap 17 corresponds to block 17.

Loading this checkpoint under a current \texttt{transformers} release requires an explicit key remapping and a
sub-config rebuild; the naive path yields zero key overlap and a randomly initialised model behind a
warning (§10).

\subsection{4.3 Probes}\label{probes}

We follow the prior work's probe designs unchanged, so that the comparison is about the encoder rather
than about probe capacity.

\begin{itemize}
\item
  \textbf{Probe B --- phoneme identity.} A CTC probe \citep{graves2006ctc}: layer normalisation → 128-dim
  bottleneck → residual temporal convolution → linear classifier; dropout 0.3, blank bias 0.0, CTC loss,
  Adam lr 1e-3, batch 256, 10 epochs, gradient clip 1.0; trained independently per layer on frame-level
  states. Metric is phoneme error rate, tone-sensitive and tone-agnostic, with model selection on
  tone-sensitive PER. Training uses 6,200 utterances (200 × 31 train speakers) and evaluation the 1,200
  utterances of the 6 unseen test speakers, so absolute PER is not comparable to the prior work's \textasciitilde39,000
  utterances; only the shape across depth is. The same probe is run on UASpeech in §5.6 over ARPAbet
  phones from a pronunciation dictionary.
\item
  \textbf{Probe C --- base-model difficulty prediction.} Named for what it measures: the labels are derived from
  the probed model's own CER, so the probe asks where in its encoder this model's eventual recognition
  failure becomes recoverable by the trained CTC readout --- which carries a bottleneck and a temporal
  convolution and is therefore not a linear probe --- not where an independently defined notion of difficulty or clinical
  severity lives. A linear softmax classifier over mean-pooled encoder states (cross-entropy, Adam lr
  1e-3, batch 4096, 10 epochs), with \textbf{speaker-disjoint} 5-fold cross-validation --- a random fold split
  would let the probe memorise speaker identity. Metric is macro-F1 over three difficulty tiers; chance is
  0.333. Two deviations must be stated. First, difficulty labels come from decoding with the \textbf{base model
  itself}: having no stronger \emph{uncontaminated} Mandarin decoder, we define difficulty as ``hard for a
  general-purpose ASR model''. Second, the tier construction is forced: the CER distribution has two large
  mass points (2,267 utterances at exactly 0.0 and 2,853 at exactly 1.0), so plain percentile thresholds
  collapse the tiers to 2,943 / 5,503 / 354. Binning at the natural breakpoints (high = CER ≥ 1.0, low =
  CER \textless{} 0.2) gives 2,813 / 2,780 / 3,207 with well-separated tier means (0.024 / 0.531 / 1.061).
\item
  \textbf{Probe A --- phoneme boundary.} Not run: at this output stride the frame-level formulation is not
  comparable to the 20 ms case without redefining the task (§3). This bounds what the \emph{probe} can measure
  at frame resolution; it is not a claim that sub-frame timing is absent from the representation, which a
  pooled or tolerance-based formulation could still test.
\end{itemize}

\subsection{4.4 Cross-condition divergence, and an explicit design weakness}\label{cross-condition-divergence-and-an-explicit-design-weakness}

The prior work compares three transcript-matched conditions per utterance: (d) original dysarthric
speech, (r) speaker-conditioned zero-shot TTS resynthesis of the same transcript, and (u) unconditioned
TTS. We reproduce the conditions but \textbf{not the per-utterance matching}. Our synthetic pools were
generated for severity-matched augmentation, which has no reason to preserve who-said-what: of the 13,627
dysarthric-conditioned synthesis rows built on CDSD speakers, 93.0\% use a text that exists somewhere in
CDSD but only \textbf{3.0\% (411 rows)} use a text the conditioning speaker actually said. That is a pairing
problem, not a volume problem.

We therefore run a \textbf{matched-distribution} design --- 37 speakers with dysarthric-conditioned synthesis for
(r), 39 healthy-conditioned speakers for (u), all 44 real CDSD speakers for (d), with per-speaker mean
pooling over \textasciitilde400 utterances each --- validated against the \textbf{411 strictly paired rows} as an exact-match
consistency check. This is weaker than a per-utterance match, which removes lexical confounds by
construction rather than in expectation. Divergence is cosine distance (1 − cos) between condition means
at each layer. A second weakness: condition (u) is conditioned on healthy control speakers from a
different corpus, so the d--u contrast mixes a speaker-set difference with a condition difference. \textbf{d--r
is the clean contrast.}

\subsection{4.5 Adaptation sweep}\label{adaptation-sweep}

LoRA \citep{hu2022lora} is applied to the encoder self-attention projections \texttt{q\_proj}, \texttt{k\_proj}, \texttt{v\_proj},
\texttt{out\_proj} of the selected blocks, with \textbf{everything else frozen, including the entire decoder} --- the
prior work's configuration, unchanged: r = 64, α = 128, dropout 0.1, AdamW at lr 3e-5, effective batch
128, 6 epochs, cosine schedule, weight decay 0.03. Adaptation starts from the \textbf{unadapted base model};
starting from an existing Mandarin dysarthric fine-tune would have measured nothing. Throughout, \textbf{\texttt{all}
means all-block LoRA} --- low-rank adapters on every encoder block --- \textbf{not} full-weight fine-tuning, and
no experiment in this paper updates base weights directly.

\textbf{Table 2.} Adaptation variants. The three eight-block bands share an identical trainable-parameter
budget; \texttt{all} uses three times as many. Module counts were verified by a dry run before training, and
merge diffs after training confirmed that each adapter touched exactly its own band and nothing else ---
ruling out the failure mode in which a silently broken adapter attach exports a copy of the base model
and the bands then score alike for a spurious reason.

{\def\LTcaptype{none} % do not increment counter
\begin{longtable}[]{@{}
  >{\raggedright\arraybackslash}p{(\linewidth - 12\tabcolsep) * \real{0.1428}}
  >{\raggedright\arraybackslash}p{(\linewidth - 12\tabcolsep) * \real{0.1428}}
  >{\raggedright\arraybackslash}p{(\linewidth - 12\tabcolsep) * \real{0.1428}}
  >{\raggedright\arraybackslash}p{(\linewidth - 12\tabcolsep) * \real{0.1428}}
  >{\raggedright\arraybackslash}p{(\linewidth - 12\tabcolsep) * \real{0.1428}}
  >{\raggedright\arraybackslash}p{(\linewidth - 12\tabcolsep) * \real{0.1428}}
  >{\raggedright\arraybackslash}p{(\linewidth - 12\tabcolsep) * \real{0.1428}}@{}}
\toprule\noalign{}
\begin{minipage}[b]{\linewidth}\raggedright
Variant
\end{minipage} & \begin{minipage}[b]{\linewidth}\raggedright
Encoder blocks
\end{minipage} & \begin{minipage}[b]{\linewidth}\raggedright
Modules
\end{minipage} & \begin{minipage}[b]{\linewidth}\raggedright
Trainable params
\end{minipage} & \begin{minipage}[b]{\linewidth}\raggedright
\% of model
\end{minipage} & \begin{minipage}[b]{\linewidth}\raggedright
Train hours
\end{minipage} & \begin{minipage}[b]{\linewidth}\raggedright
Merge diff (tensors)
\end{minipage} \\
\midrule\noalign{}
\endhead
\bottomrule\noalign{}
\endlastfoot
\texttt{lower} & 1--8 & 32 & 4,194,304 & 0.205\% & 3.25 & 32 \\
\texttt{middle} & 9--16 & 32 & 4,194,304 & 0.205\% & 3.16 & 32 \\
\textbf{\texttt{upper}} & \textbf{17--24} & \textbf{32} & \textbf{4,194,304} & \textbf{0.205\%} & \textbf{3.12} & \textbf{32} \\
\texttt{all} & 1--24 & 96 & 12,582,912 & 0.614\% & 3.54 & 96 \\
base & --- & 0 & 0 & 0\% & --- & --- \\
\end{longtable}
}

\texttt{middle} (blocks 9--16) is the proportional analogue of the prior work's winning 5--8 band on a 12-block
encoder; \texttt{upper} (17--24) is the analogue of their weaker 9--12 band, and \texttt{lower} (1--8) of their 1--4. Every
one of the 33,617 training utterances read successfully in all four runs.

\subsection{4.6 Evaluation and statistics}\label{evaluation-and-statistics}

Adapters are merged and decoded through the model's own packaged inference path (§10) on the 5,388
held-out utterances of the 6 unseen test speakers. \textbf{Mandarin has no orthographic word boundaries, so the
scorer tokenises by character and the edit rate over those tokens IS the character error rate; we
therefore report CER only for Mandarin and do not label a character-tokenised edit rate ``WER.''} Because
six test speakers is a small N, we report per-speaker CER, micro- and macro-averages, \textbf{paired speaker
bootstrap 95\% confidence intervals (B = 10,000)} on every contrast, and a leave-one-speaker-out
sensitivity table. \textbf{Every CDSD adaptation configuration was trained at three seeds (42/43/44)} and all
CDSD intervals are speaker × seed (§6.1.1); the CDSD \emph{probe} curves of §5.1 remain single-run (§9).

\begin{center}\rule{0.5\linewidth}{0.5pt}\end{center}

\section{5. Probing results}\label{probing-results}

\subsection{5.1 Phoneme identity matures late, not mid-stack}\label{phoneme-identity-matures-late-not-mid-stack}

\textbf{The band was selected from the validation speakers, not the test speakers.} Probes are trained on the
31 training speakers, the depth profile that chooses the adaptation band is read off the \textbf{7 validation
speakers}, and the 6 test speakers are touched only once, by the adaptation sweep in §6.

\textbf{Table 3.} Tone-sensitive PER (\%) by encoder layer, base model. \textbf{VAL} (7 unseen speakers) is the
selection basis; \textbf{TEST} (6 unseen speakers) is reported for description only. Selected taps; full
25-tap curves in the released CSV.

{\def\LTcaptype{none} % do not increment counter
\begin{longtable}[]{@{}
  >{\raggedright\arraybackslash}p{(\linewidth - 16\tabcolsep) * \real{0.1111}}
  >{\raggedright\arraybackslash}p{(\linewidth - 16\tabcolsep) * \real{0.1111}}
  >{\raggedright\arraybackslash}p{(\linewidth - 16\tabcolsep) * \real{0.1111}}
  >{\raggedright\arraybackslash}p{(\linewidth - 16\tabcolsep) * \real{0.1111}}
  >{\raggedright\arraybackslash}p{(\linewidth - 16\tabcolsep) * \real{0.1111}}
  >{\raggedright\arraybackslash}p{(\linewidth - 16\tabcolsep) * \real{0.1111}}
  >{\raggedright\arraybackslash}p{(\linewidth - 16\tabcolsep) * \real{0.1111}}
  >{\raggedright\arraybackslash}p{(\linewidth - 16\tabcolsep) * \real{0.1111}}
  >{\raggedright\arraybackslash}p{(\linewidth - 16\tabcolsep) * \real{0.1111}}@{}}
\toprule\noalign{}
\begin{minipage}[b]{\linewidth}\raggedright
Layer
\end{minipage} & \begin{minipage}[b]{\linewidth}\raggedright
0
\end{minipage} & \begin{minipage}[b]{\linewidth}\raggedright
4
\end{minipage} & \begin{minipage}[b]{\linewidth}\raggedright
8
\end{minipage} & \begin{minipage}[b]{\linewidth}\raggedright
12
\end{minipage} & \begin{minipage}[b]{\linewidth}\raggedright
16
\end{minipage} & \begin{minipage}[b]{\linewidth}\raggedright
20
\end{minipage} & \begin{minipage}[b]{\linewidth}\raggedright
\textbf{21}
\end{minipage} & \begin{minipage}[b]{\linewidth}\raggedright
24
\end{minipage} \\
\midrule\noalign{}
\endhead
\bottomrule\noalign{}
\endlastfoot
PER --- \textbf{VAL} (selection) & 71.17 & 62.66 & 54.20 & 47.04 & 39.29 & 34.06 & \textbf{33.01} & 36.56 \\
PER --- TEST (descriptive) & 75.14 & 66.96 & 57.64 & 49.46 & 41.07 & 34.39 & \textbf{33.58} & 35.43 \\
PER tone-agnostic (TEST) & 78.07 & 67.08 & 55.54 & 45.92 & 37.85 & 33.41 & 33.27 & 32.95 \\
\end{longtable}
}

\textbf{The two splits agree on every feature the selection depends on.} On the validation speakers PER
decreases monotonically from layer 0 to layer 21 and then flattens, the minimum sits at \textbf{layer 21} on
both splits, and from the midpoint (layer 12, 47.04) to that minimum the curve gains a further
\textbf{14.03 pp} on VAL against 15.88 pp on TEST. Mean PER over blocks 9--16 is \textbf{46.00} on VAL; over blocks
17--24 it is \textbf{34.88}.

\textbf{Table 4.} Per-band summary of the base-model tone-sensitive PER curve, on the validation split
(selection basis; TEST values in parentheses). Note that the band with the \emph{steepest} improvement is not
the band with the \emph{lowest} PER, and §6 shows the two orderings make opposite predictions.

{\def\LTcaptype{none} % do not increment counter
\begin{longtable}[]{@{}
  >{\raggedright\arraybackslash}p{(\linewidth - 8\tabcolsep) * \real{0.2000}}
  >{\raggedright\arraybackslash}p{(\linewidth - 8\tabcolsep) * \real{0.2000}}
  >{\raggedright\arraybackslash}p{(\linewidth - 8\tabcolsep) * \real{0.2000}}
  >{\raggedright\arraybackslash}p{(\linewidth - 8\tabcolsep) * \real{0.2000}}
  >{\raggedright\arraybackslash}p{(\linewidth - 8\tabcolsep) * \real{0.2000}}@{}}
\toprule\noalign{}
\begin{minipage}[b]{\linewidth}\raggedright
Band
\end{minipage} & \begin{minipage}[b]{\linewidth}\raggedright
Blocks
\end{minipage} & \begin{minipage}[b]{\linewidth}\raggedright
Mean PER
\end{minipage} & \begin{minipage}[b]{\linewidth}\raggedright
PER at band entry → exit
\end{minipage} & \begin{minipage}[b]{\linewidth}\raggedright
Drop across band
\end{minipage} \\
\midrule\noalign{}
\endhead
\bottomrule\noalign{}
\endlastfoot
\texttt{lower} & 1--8 & 61.08 (64.66) & 71.17 → 54.20 & \textbf{16.97 pp (steepest)} \\
\texttt{middle} & 9--16 & 46.00 (48.17) & 54.20 → 39.29 & 14.91 pp \\
\texttt{upper} & 17--24 & \textbf{34.88 (35.20) --- lowest} & 39.29 → 36.56 & 2.72 pp (flattest) \\
\end{longtable}
}

The curve's knee --- the depth beyond which PER stops improving --- sits at layer 21, inside the \texttt{upper}
band. Muthu et al.~report PER plateauing after their layer 6 of 12, i.e.~at \textbf{50\% of relative depth};
here it continues to improve to \textbf{≈85--88\%}.

\textbf{That contrast is between our measurement and their published figure, and it does not survive measuring
both ourselves.} §5.5 places Whisper-small's minimum, under this paper's own pipeline, at layer 10 of 12
--- \textbf{83\% of relative depth}, not 50\% --- while at matched probe-training volume the Qwen3-ASR minimum
resolves to a flat plateau across layers 20--23 (a spread of 0.27 pp). \textbf{Both probes reach their minima
late in the stack}, so the apparent depth-profile difference in the published comparison is not a
property of the two architectures. Because the argmin layer sits inside the noise band of a single-seed
estimate, only the \emph{band} is claimable in either architecture.

\textbf{The prediction this licensed was documented before the adaptation sweep ran.} The prior work justifies
mid-band placement by mid-encoder phoneme maturation; on Qwen3-ASR that premise is false, so the
productive band should be the upper one rather than the proportional 9--16 analogue. §6 is that
prediction's test, and \textbf{it is confirmed on Qwen3-ASR} --- but its \emph{rationale} does not survive, since
Whisper-small's own knee also sits in its upper band, where the same reasoning is wrong (§6.6).

\textbf{A secondary observation, with a caveat that limits it.} A Mandarin dysarthric fine-tune of the same
base model tracks base to within \textasciitilde1 pp for layers 0--12 and then diverges steadily (layer 16: 41.07 vs
37.80; layer 20: 34.39 vs 25.67) --- fine-tuning reshaped the top third of the encoder and left the lower
two thirds untouched. That fine-tune had, however, seen 81.9\% of the probe utterances and all six test
speakers during its own training. \textbf{The layer localisation is informative; the magnitude is not
interpretable as generalisation.} All headline claims rest on the base model, which has never seen this
corpus.

\textbf{Tone.} The tone-sensitive minus tone-agnostic gap is negative for layers 0--4 (−2.93 pp at layer 0),
crosses over at layer 5, peaks at +3.5 to +3.8 pp over layers 10--16, and narrows at the top. From layer 5
upward this reproduces the prior work's finding that requiring tone costs accuracy. The early-layer
inversion is a normalisation artefact: at 75--78\% PER the probe is near chance, and stripping tone tokens
shortens the reference by roughly a third, so a similar error count is divided by a smaller denominator.

\subsection{5.2 Cross-condition divergence peaks mid-encoder and then falls}\label{cross-condition-divergence-peaks-mid-encoder-and-then-falls}

\textbf{Figure 1, upper panel}, plots the curves discussed in this section; its lower panel belongs to §5.3.

\begin{figure}[htbp]
\centering
\includegraphics[width=\linewidth]{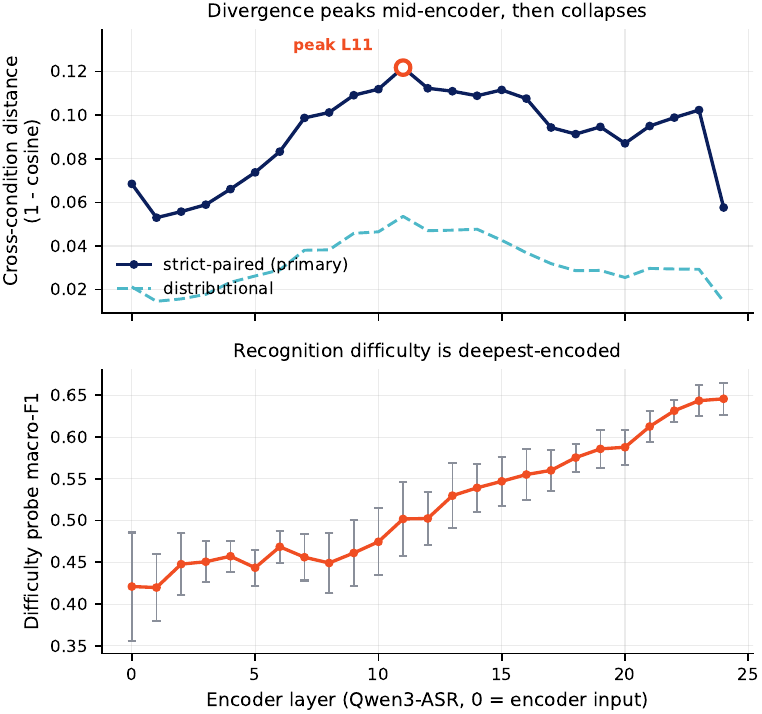}

\caption{Upper: cross-condition representational distance (1 − cosine) between dysarthric and resynthesised speech by Qwen3-ASR encoder layer, strict-paired as primary with the distributional variant shown for comparison. It peaks at layer 11 of 24 and then collapses toward the output, where the Whisper-small result reported in prior work rises with depth. Lower: recognition-difficulty probe macro-F1 by layer, speaker-disjoint 5-fold cross-validation, error bars ±1 SD across folds. Difficulty is deepest-encoded — the one prior-work finding that reproduces here.}
\label{fig:fig1_divergence_difficulty}%

\end{figure}

\textbf{Alt text (Figure 1).} Two stacked line charts sharing an x-axis of Qwen3-ASR encoder layer index, 0 to 24. Upper panel: cross-condition representational distance, one minus cosine similarity, between dysarthric and speaker-conditioned resynthesised speech. The curve rises from near zero at the input, peaks in the middle of the stack, then falls steadily through the upper third, so the two conditions look most alike again at the top of the encoder. Two curves are plotted --- the strictly paired primary analysis as a solid line with markers, and the distributional variant as a dashed line --- and both share that rise-then-fall shape, peaking at layer 11; Table 5 reports the two further sensitivity variants numerically rather than plotting them. Lower panel: the difficulty probe across all 25 taps with per-fold error bars, rising with depth to its maximum macro-F1 of 0.646 at layer 24.

\textbf{Table 5.} Cross-condition cosine distance: peak location and collapse toward the output, across the
primary analysis and three sensitivity variants. (d) real dysarthric speech, (r) speaker-conditioned
resynthesis, (u) unconditioned synthesis. The full per-layer curves are in Figure 1 and the released CSV;
this table reports what the figure cannot --- that all four analyses agree on the peak and its magnitude.

{\def\LTcaptype{none} % do not increment counter
\begin{longtable}[]{@{}
  >{\raggedright\arraybackslash}p{(\linewidth - 12\tabcolsep) * \real{0.1428}}
  >{\raggedright\arraybackslash}p{(\linewidth - 12\tabcolsep) * \real{0.1428}}
  >{\raggedright\arraybackslash}p{(\linewidth - 12\tabcolsep) * \real{0.1428}}
  >{\raggedright\arraybackslash}p{(\linewidth - 12\tabcolsep) * \real{0.1428}}
  >{\raggedright\arraybackslash}p{(\linewidth - 12\tabcolsep) * \real{0.1428}}
  >{\raggedright\arraybackslash}p{(\linewidth - 12\tabcolsep) * \real{0.1428}}
  >{\raggedright\arraybackslash}p{(\linewidth - 12\tabcolsep) * \real{0.1428}}@{}}
\toprule\noalign{}
\begin{minipage}[b]{\linewidth}\raggedright
Analysis
\end{minipage} & \begin{minipage}[b]{\linewidth}\raggedright
Pair
\end{minipage} & \begin{minipage}[b]{\linewidth}\raggedright
n
\end{minipage} & \begin{minipage}[b]{\linewidth}\raggedright
Peak layer
\end{minipage} & \begin{minipage}[b]{\linewidth}\raggedright
Peak distance
\end{minipage} & \begin{minipage}[b]{\linewidth}\raggedright
At L24
\end{minipage} & \begin{minipage}[b]{\linewidth}\raggedright
peak/L24
\end{minipage} \\
\midrule\noalign{}
\endhead
\bottomrule\noalign{}
\endlastfoot
\textbf{Primary, paired} & \textbf{d--r} & \textbf{411} & \textbf{11} & \textbf{0.1218} & 0.0577 & \textbf{2.11×} \\
sensitivity --- distributional, per speaker & d--r & 37 & 11 & 0.0764 & 0.0232 & 3.29× \\
sensitivity --- distributional, global & d--r & 8,800 & 11 & 0.0536 & 0.0145 & 3.69× \\
sensitivity --- distributional, global & d--u & 8,800 & 11 & 0.0567 & 0.0242 & 2.35× \\
\end{longtable}
}

\textbf{The primary analysis is the 411-row strictly paired (d, r) subset}, in which each pair is the same
speaker saying the same sentence under the two conditions, so lexical content and speaker identity are
matched by construction. The distributional rows are sensitivity analyses; the d--u row confounds condition
with source corpus and should carry the least weight.

All four peak at \textbf{layer 11 of 24} (the r--u pair peaks at 15) and then fall by \textbf{2.1--3.7×} toward the
output. Muthu et al.~report divergence rising with depth, and that rise is the evidence for their headline
claim that disordered speech affects high-level representations more than low-level acoustic ones. \textbf{On
this encoder that claim does not hold as stated:} divergence is maximal at ≈46\% of relative depth and the
top third of the encoder pulls the three conditions back together. The three distributional analyses
reproduce the primary subset's shape and its layer-11 peak, which is the condition under which the
matched-distribution design is admissible. Magnitudes are small (cosine distance ≤ 0.12) and \textbf{no
confidence intervals are computed for any curve in this table} --- a real limitation of this section (§9).
r--u divergence is an order of magnitude smaller than d--r or d--u at every layer (peak 0.0116), so the two
synthetic conditions resemble each other far more than either resembles real dysarthric speech.

\textbf{Joint reading with §5.1.} PER keeps improving into the upper layers while cross-condition divergence
falls there, so the upper third of this encoder abstracts away speaker identity and synthesis artefacts
while continuing to sharpen linguistic content --- what one would expect of layers feeding a projector into
a language-model decoder, and which a 12-block encoder with half as many layers either lacks or never
reaches.

\subsection{5.3 Recognition difficulty is deepest-encoded --- this one reproduces}\label{recognition-difficulty-is-deepest-encoded-this-one-reproduces}

\textbf{Figure 1, lower panel}, plots the difficulty probe across all 25 taps with per-fold error bars.
Macro-F1 rises monotonically with depth, from \textbf{0.421} at layer 0 to a maximum of \textbf{0.646} at layer 24
(chance = 0.333, speaker-disjoint 5-fold cross-validation), and the fold standard deviations tighten with
depth (±0.065 at layer 0 against ±0.019 at layer 24). This \textbf{reproduces} the prior work's finding that
recognition difficulty is encoded predominantly in the deepest encoder representations --- on a different
encoder family, backbone size and label source. A fine-tuned model gives a near-identical curve (0.642 at
the final layer), so the signal is a property of the architecture's depth profile rather than of
adaptation.

\subsection{5.4 Scorecard}\label{scorecard}

\textbf{Table 6.} The prior work's testable claims against this encoder, same corpus.

{\def\LTcaptype{none} % do not increment counter
\begin{longtable}[]{@{}
  >{\raggedright\arraybackslash}p{(\linewidth - 4\tabcolsep) * \real{0.3333}}
  >{\raggedright\arraybackslash}p{(\linewidth - 4\tabcolsep) * \real{0.3333}}
  >{\raggedright\arraybackslash}p{(\linewidth - 4\tabcolsep) * \real{0.3333}}@{}}
\toprule\noalign{}
\begin{minipage}[b]{\linewidth}\raggedright
\citet{muthu2026layerwise} finding, on Whisper-small
\end{minipage} & \begin{minipage}[b]{\linewidth}\raggedright
Qwen3-ASR result
\end{minipage} & \begin{minipage}[b]{\linewidth}\raggedright
Verdict
\end{minipage} \\
\midrule\noalign{}
\endhead
\bottomrule\noalign{}
\endlastfoot
Phoneme boundary information weak at every layer & Not measurable --- \textasciitilde77 ms frames, 49.4\% of phones sub-frame (§3) & \textbf{Untestable on this architecture} \\
Phoneme identity plateaus after mid-encoder & Improves monotonically to ≈85\% depth; −14.03 pp past the midpoint on the selection split, −15.88 pp on test (§5.1) & \textbf{Contradicted} \\
Cross-condition divergence grows with depth & Inverted U peaking at layer 11 of 24, then falls 2.1--3.7× (§5.2) & \textbf{Contradicted} \\
Recognition difficulty encoded in the deepest layers & Monotone rise to macro-F1 0.646 at layer 24 (§5.3) & \textbf{Reproduced} \\
Mandarin tone a persistent error source & Tone-sensitive PER worse from layer 5 up, peak gap +3.8 pp (§5.1) & \textbf{Reproduced} \\
\end{longtable}
}

Two entries need a caveat. The rows are scored against the prior work's \textbf{published} Whisper-small
results; where we have since measured the same quantity ourselves (row 2), our own Whisper-small probe
does \emph{not} reproduce their plateau (§5.5), so that row records a disagreement with a published curve
rather than a demonstrated architectural difference. Rows 4 and 5 are reproductions and are unaffected.

\begin{center}\rule{0.5\linewidth}{0.5pt}\end{center}

\subsection{5.5 The Whisper-small profile, measured here rather than cited}\label{the-whisper-small-profile-measured-here-rather-than-cited}

Everything above compares our Qwen3-ASR measurement against Whisper-small results reported elsewhere. We
therefore ran the Whisper-small probe ourselves, on the same CDSD manifest, speaker split, integrity gate,
probe architecture, optimiser (Adam, lr 1e-3, batch 256, 10 epochs, gradient clipping 1.0, dropout 0.3),
unit inventory and model-selection criterion. Whisper-small exposes 13 taps (embedding output plus 12
blocks), the analogue of Qwen3-ASR's 25.

\textbf{The minimum is at layer 10 of 12 --- 83\% of relative depth --- and there is no plateau after layer 6.}
\textbf{Figure 2} plots both architectures against \emph{relative} depth on one axis, which is the clearest
statement of this paper's negative result.

\begin{figure}[htbp]
\centering
\includegraphics[width=\linewidth]{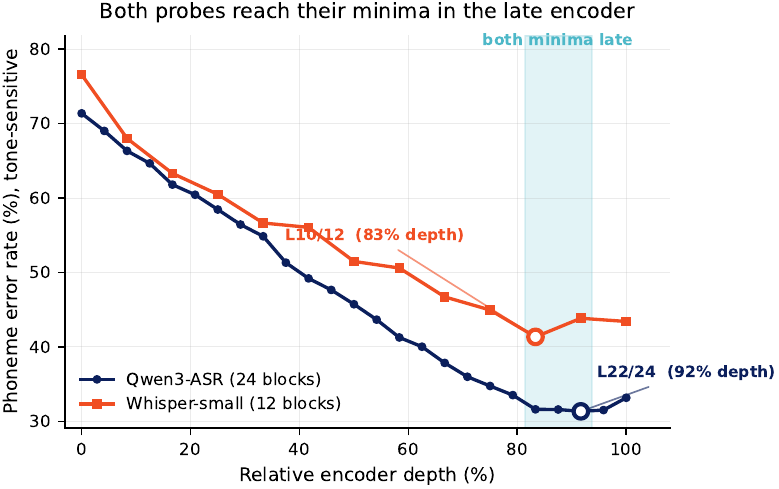}
\caption{Tone-sensitive phoneme error rate against \emph{relative} encoder depth, both architectures at matched probe-training volume (33,617 utterances), same corpus, split, probe and integrity gate. Open markers are each curve's minimum. Whisper-small minimises at layer 10 of 12 (83\% of depth) and Qwen3-ASR at layer 22 of 24 (92\%); the shaded band spans the two. Both minima fall in the late encoder, yet the two models prefer different adaptation bands (Figure 4), which is why no rule reading the band off this curve can work.}
\label{fig:fig2_depth_profiles}%
\end{figure}

\textbf{Alt text (Figure 2).} A line chart of tone-sensitive phoneme error rate, y-axis, against relative encoder depth from 0 at the input to 1 at the output, x-axis, with one curve per architecture. Both curves fall steeply over the first half of the stack. The Qwen3-ASR curve keeps improving to roughly 92 per cent of depth, minimising at layer 22 of 24; the Whisper-small curve also improves into its late stack and minimises at layer 10 of 12, about 83 per cent of depth, gaining a further 10.14 pp between layer 6 and layer 10 rather than flattening at mid-stack. Plotting against relative rather than absolute depth is what makes the two architectures, which have different numbers of blocks, directly comparable.

Between layer 6 and layer 10 the curve gains a further 10.14 pp at matched volume: Whisper-small, under
this pipeline, matures late in exactly the way Qwen3-ASR does.

\textbf{We checked four explanations for the disagreement with the published curve, and none accounts for it.}

\begin{enumerate}
\def\labelenumi{\arabic{enumi}.}
\item
  \emph{Probe configuration.} Matched item by item against the prior work's description: probe architecture,
  optimiser, batch size, epochs, unit inventory and 70/15/15 speaker-disjoint splits.
\item
  \emph{Baseline competence --- no longer ruled out, and the reason is instructive.} An earlier version of
  this section reported that our unadapted baseline agreed with the prior work's to 1.5 pp.~That
  comparison used the figures in \textbf{v1} of their preprint, which we read on 13 August 2026. \textbf{Their v2
  of 15 August revised the results tables}, and the baseline moved from 105.08 to \textbf{89.98} CER
  (all-block LoRA likewise from 68.85 to 64.35). We cite v2 throughout, and against it the agreement
  does not hold: our corrected baseline of 77.57 is \textbf{12.4 pp} better than theirs. \textbf{We therefore
  cannot rule baseline competence out as a contributor to the probe-curve disagreement, and no longer
  claim to.} It joins the list of live candidates rather than the list of eliminated ones.

  Two observations keep this informative rather than merely deflating. First, \textbf{their revision runs in
  the same direction and for plausibly the same reason as ours}: an unadapted Whisper baseline above
  100\% CER was corrected downward by \textasciitilde15 pp, exactly as ours fell from 103.57 to 77.57 once repetition
  control was applied (§6.6.1). Two groups independently retracting an inflated Whisper baseline on the
  same corpus is better evidence that the artefact is real and general than either correction alone.
  Second, \textbf{this check never bore on the probe result anyway}: the probe is a CTC readout trained on
  encoder activations and never runs the autoregressive decoder, so \textbf{no probe result in this paper
  depends on the decode configuration} --- of ours or of theirs.
\item
  \emph{Padding.} Whisper pads every input to a fixed 30 s / 1500-frame window, and the prior work truncates
  to 1500 frames where we slice to the \textasciitilde133 real frames per utterance. A matched-pair extraction wrote
  both a padded and a sliced shard set over the \emph{same} 1,188 utterances and the \emph{same} 7 taps, verified
  numerically so that padding was the only variable. \textbf{Both arms minimise at layer 10.}
\item
  \emph{Probe training volume.} The prior work trains on \textasciitilde39,000 utterances; our first run used 6,200. At
  33,617 the curve shifts down \textasciitilde4.9 pp on average and \textbf{the location of the late minimum is unchanged},
  though the magnitude of the post-minimum rise changes substantially (Table 7).
\end{enumerate}

The layer-10 minimum therefore holds across four configurations spanning a 28× range in sample size and
opposite padding treatments (8,800 sliced; 1,188 sliced; 1,188 padded; 33,617 sliced). \textbf{We report the
discrepancy as unexplained.}

\textbf{An unexpected result about padding belongs in the record.} Padding did not dilute the probe; it helped
it substantially (87.89 → 53.54 PER at the minimum in the matched-pair experiment). Whisper's encoder does
not honour attention masks and attends across the whole 30 s window, so the padded positions carry
contextualised, smeared copies of the speech, giving the CTC probe hundreds of informative positions to
align \textasciitilde15 units against instead of \textasciitilde133. On this reading \textbf{Whisper's padded frames are an
information channel rather than noise} --- a genuine asymmetry with the masked Qwen3-ASR arm (§9). We
offer that as a \textbf{hypothesis}: the experiment establishes that padding changes the probe's readout
substantially and in which direction, not that the proposed information-flow mechanism is the reason.

\textbf{Probe-readout sensitivity to training volume is depth-dependent, and the dependence differs by
architecture.} This is the result that explains how the discarded depth-profile claim arose.

\textbf{A confound we must state before interpreting it.} Both volumes are trained for the same 10 epochs, so
the 33,617-utterance condition receives roughly 5.4× as many optimiser updates as well as 5.4× as many
unique utterances. \textbf{This experiment therefore identifies sensitivity to training volume under a
fixed-epoch recipe, not sensitivity to the number of unique utterances at matched optimisation.} A
deeper representation may need more optimiser steps for this CTC readout rather than more independent
data, and the small condition may be under-trained rather than data-starved. Separating the two needs a
step-matched control --- resampling the 6,200-utterance set to the larger step count, and training the
large set at the smaller step count --- at a few representative depths. We have not run it, so we describe
this result as \textbf{volume sensitivity} throughout and avoid ``data efficiency'' as a mechanism claim (§9).

\textbf{Table 7.} Probe-readout sensitivity to training volume by depth. Gains are the improvement in
tone-sensitive PER when probe training data increases 5.4× (6,200 → 33,617 utterances) at a fixed 10
epochs, on the same speaker-disjoint split.

{\def\LTcaptype{none} % do not increment counter
\begin{longtable}[]{@{}
  >{\raggedright\arraybackslash}p{(\linewidth - 6\tabcolsep) * \real{0.2500}}
  >{\raggedright\arraybackslash}p{(\linewidth - 6\tabcolsep) * \real{0.2500}}
  >{\raggedright\arraybackslash}p{(\linewidth - 6\tabcolsep) * \real{0.2500}}
  >{\raggedright\arraybackslash}p{(\linewidth - 6\tabcolsep) * \real{0.2500}}@{}}
\toprule\noalign{}
\begin{minipage}[b]{\linewidth}\raggedright
Encoder
\end{minipage} & \begin{minipage}[b]{\linewidth}\raggedright
Mean gain across layers, 6,200 → 33,617
\end{minipage} & \begin{minipage}[b]{\linewidth}\raggedright
Deepest layer's gain
\end{minipage} & \begin{minipage}[b]{\linewidth}\raggedright
Ratio to that stack's mean
\end{minipage} \\
\midrule\noalign{}
\endhead
\bottomrule\noalign{}
\endlastfoot
Whisper-small & −4.88 pp & layer 12: \textbf{−9.15 pp} & \textbf{1.87×} \\
Qwen3-ASR & −3.03 pp & layer 24: \textbf{−2.25 pp} & \textbf{0.74×} \\
\end{longtable}
}

Whisper-small's deepest representation is the \textbf{most} volume-sensitive layer in its stack; Qwen3-ASR's
is among the \textbf{least}. \textbf{Figure 3} shows the consequence directly.

\begin{figure}[htbp]
\centering
\includegraphics[width=\linewidth]{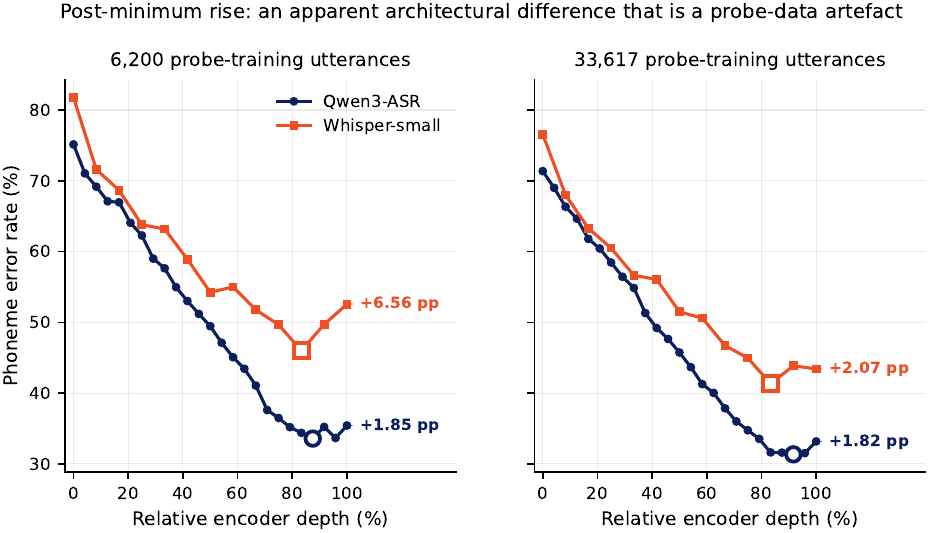}

\caption{The same two depth profiles at two probe-training volumes. Labels give each curve's rise from its own minimum to its final layer. At 6,200 utterances the architectures appear to differ sharply in late-stack behaviour (+6.56 against +1.85 pp); at 33,617 they do not (+2.07 against +1.82 pp). Whisper-small's tail flattens because its deepest block is the most volume-sensitive layer in its stack, while Qwen3-ASR's is essentially volume-invariant (1.85 → 1.82). The layer-10 minimum survives a 5.4× change in training data, which is a robustness check against one specific manipulation---\textbf{not} a substitute for the confidence intervals these curves still lack (§9, Limitation 17).}
\label{fig:fig3_volume_artefact}%

\end{figure}

\textbf{Alt text (Figure 3).} The same two depth profiles as Figure 2, each drawn twice, once per probe-training volume. Within each architecture the two volumes trace curves of the same shape, so the depth at which phoneme identity stops improving does not move when the probe is given more data. Labels give each curve's rise.

The consequence is cautionary. \textbf{Report the probe curve at more than one training volume, or do not
compare late-stack shape across architectures.}

\textbf{The two minima differ in shape, and this is measured rather than asserted.} Both probes were re-run at
matched volume under three seeds (0/1/2), with per-utterance error counts retained so that the paired
speaker × seed bootstrap used for the adaptation contrasts (B = 10,000, contrasts formed within a seed)
applies to the curves.

On Whisper-small, L10 − L6 = −7.67 [−11.92, −3.98], L11 − L10 = +1.83 [+0.49, +3.54] and
L12 − L10 = +1.55 [+0.07, +3.11] --- all three resolved. On Qwen3-ASR, L22 − L16 = −6.20 [−9.34, −2.75] and
L22 − L20 = −0.64 [−1.29, −0.02] resolve, but the terminal contrast L24 − L22 = −0.15 [−2.63, +2.43] spans
zero. \textbf{Whisper-small rises significantly past its minimum; Qwen3-ASR does not} --- its late stack is flat
in a way Whisper-small's is not.

Judged against the raw per-layer seed standard deviations (mean 1.43 pp on Whisper-small), the +1.83 pp
step would look like noise. It is not: pairing within a seed cancels that seed's overall difficulty, which
is why the paired interval excludes zero where a marginal comparison would not. Each claim also sits at
the \textbf{most precisely estimated point on its curve} (seed SD 0.128 pp at Whisper-small's layer 10 against
2.75 pp at its layer 2), while the speaker-bootstrap intervals on individual layers are wide --- restating
on the probe curves what §6.5 found on the adaptation sweeps: \textbf{speaker sampling, not retraining, is the
binding source of uncertainty.}

\begin{center}\rule{0.5\linewidth}{0.5pt}\end{center}

\subsection{5.6 Both depth profiles, measured again on the second corpus}\label{both-depth-profiles-measured-again-on-the-second-corpus}

Everything in §5.1--§5.5 is a CDSD measurement, and the same fixed pair of probe curves stands behind a
placement \emph{difference} on CDSD and placement \emph{agreement} on UASpeech. Either the phoneme-identity curve is
intrinsically non-predictive of the band, or it is \textbf{corpus-dependent} --- in which case placement might
still be readable off a probe measured on the corpus in hand. The second reading would leave this paper's
negative result substantially weaker, and nothing in §5.1--§5.5 distinguishes the two. \textbf{We therefore ran
Probe B on UASpeech --- both models, every tap, three seeds.}

\textbf{Two departures from the CDSD probe protocol bound what this subsection may be used for.}

\begin{enumerate}
\def\labelenumi{\arabic{enumi}.}
\item
  \textbf{Targets are canonical dictionary pronunciations, not forced alignments.} No aligner is installed on
  the machine that holds UASpeech, and UASpeech is not redistributable, so it cannot be moved to compute
  where one is. Probe B is a CTC readout and requires the target \emph{sequence}, not time-aligned boundaries,
  and UASpeech's closed 449-word vocabulary makes a dictionary lookup exact. Targets are therefore CMU
  Pronouncing Dictionary pronunciations \citep{cmudict07b} --- version 0.7b, 135,166 entries --- with
  stress stripped, over a 39-phone ARPAbet inventory. \textbf{This makes the UASpeech probe target the
  canonical phone sequence implied by the word label, not the sequence the speaker actually produced.}
  In a pathology study that distinction matters, because substitutions, deletions and distortions are
  part of the phenomenon being measured: on UASpeech the probe measures decodability of the \emph{canonical}
  phonological sequence, and its absolute PER is therefore not commensurable with CDSD's
  alignment-derived targets. Only the shape across depth is compared. \textbf{This also reconciles the two
  UASpeech test counts}: the adaptation sweep scores all 3,798 held-out utterances, the probe 3,738 ---
  the 60-utterance difference is those whose word has no CMUdict pronunciation and therefore no phone
  target, which the probe must drop and the ASR sweep does not. Coverage
  is \textbf{11,222 of 11,402 utterances (98.4\%)} --- 10,942 direct hits and 280 recovered by compound
  decomposition (\emph{backspace} → \emph{back} + \emph{space}) --- with 180 utterances across 10 word types dropped for
  having no pronunciation.
\item
  \textbf{There is no integrity gate.} The CDSD probe inherits the alignment gate of §4.1, which removes 14.6\%
  of utterances and thereby also removes very short and effectively empty recordings. Nothing equivalent
  exists here, because there are no alignments to fail.
\end{enumerate}

\textbf{Absolute PER is therefore not comparable between the two corpora}: the unit inventories differ, one
corpus is gated and the other is not, and the material is isolated words against read sentences. Only the
\textbf{shape across depth} is comparable, and only shape is used below. Splits are the same speaker-disjoint
8 train / 2 validation / 5 test partition used in §6.5, giving 5,979 / 1,505 / 3,738 utterances; probe
architecture, optimiser, model-selection criterion and per-layer independence are unchanged from §4.3;
every tap of both models was trained three times with seeds 0/1/2.

\textbf{Table 8.} Phoneme-identity PER (\%) by layer on UASpeech, three seeds (0/1/2), mean ± SD. Layer 0 is
the encoder input; layers 1--24 (Qwen3-ASR) and 1--12 (Whisper-small) are block outputs. \textbf{Layer index is
not the comparable axis across the two models} --- relative depth is, and each curve's minimum is in
bold. Band means use the same proportional definitions as the adaptation sweep. Absolute values are
\textbf{not} comparable to the CDSD curves in Table 3 or §5.5.

{\def\LTcaptype{none} % do not increment counter
\begin{longtable}[]{@{}
  >{\raggedright\arraybackslash}p{(\linewidth - 4\tabcolsep) * \real{0.3333}}
  >{\raggedright\arraybackslash}p{(\linewidth - 4\tabcolsep) * \real{0.3333}}
  >{\raggedright\arraybackslash}p{(\linewidth - 4\tabcolsep) * \real{0.3333}}@{}}
\toprule\noalign{}
\begin{minipage}[b]{\linewidth}\raggedright
Layer
\end{minipage} & \begin{minipage}[b]{\linewidth}\raggedright
Qwen3-ASR (24 blocks)
\end{minipage} & \begin{minipage}[b]{\linewidth}\raggedright
Whisper-small (12 blocks)
\end{minipage} \\
\midrule\noalign{}
\endhead
\bottomrule\noalign{}
\endlastfoot
0 & 85.43 ± 0.55 & 93.12 ± 0.18 \\
1 & 83.42 ± 0.81 & 90.44 ± 1.49 \\
2 & 80.77 ± 1.04 & 86.02 ± 0.56 \\
3 & 77.07 ± 0.57 & 81.02 ± 0.50 \\
4 & 74.85 ± 0.37 & 73.01 ± 0.29 \\
5 & 71.26 ± 0.56 & 71.25 ± 3.91 \\
6 & 68.34 ± 1.00 & 69.53 ± 4.96 \\
7 & 65.06 ± 0.46 & 63.51 ± 4.81 \\
8 & 62.05 ± 0.48 & 55.64 ± 0.62 \\
9 & 59.66 ± 0.61 & 53.90 ± 0.45 \\
10 & 56.77 ± 0.16 & 54.46 ± 1.38 \\
11 & 54.33 ± 0.71 & \textbf{53.86 ± 0.81} \\
12 & 52.12 ± 0.44 & 69.79 ± 0.70 \\
13 & 48.52 ± 0.50 & --- \\
14 & 47.14 ± 0.48 & --- \\
15 & 45.58 ± 0.11 & --- \\
16 & 44.93 ± 0.57 & --- \\
17 & 43.49 ± 0.59 & --- \\
18 & 42.06 ± 0.61 & --- \\
19 & 41.58 ± 0.38 & --- \\
20 & \textbf{41.04 ± 1.13} & --- \\
21 & 41.15 ± 0.66 & --- \\
22 & 41.55 ± 0.42 & --- \\
23 & 41.42 ± 0.67 & --- \\
24 & 42.08 ± 0.50 & --- \\
\textbf{band mean --- \texttt{lower}} & 72.85 (blocks 1--8) & 82.62 (blocks 1--4) \\
\textbf{band mean --- \texttt{middle}} & 51.13 (blocks 9--16) & 64.98 (blocks 5--8) \\
\textbf{band mean --- \texttt{upper}} & \textbf{41.80 (blocks 17--24)} & \textbf{58.00 (blocks 9--12)} \\
\end{longtable}
}

\textbf{Both curves minimise late, and both band orderings put \texttt{upper} lowest.} Qwen3-ASR's minimum sits at
\textbf{layer 20 of 24 --- 83\% of relative depth}; Whisper-small's at \textbf{layer 11 of 12 --- 92\%}. Neither shows a
mid-stack plateau. Band means order \texttt{upper} \textless{} \texttt{middle} \textless{} \texttt{lower} on both models (41.80 \textless{} 51.13 \textless{} 72.85
and 58.00 \textless{} 64.98 \textless{} 82.62), which is the ordering both models also give on CDSD.

\textbf{Table 9.} Depth of the phoneme-identity PER minimum in all four cells of the model × corpus design.
Every entry is the argmin of a probe curve measured under this paper's own pipeline; the CDSD entries
are the matched-volume three-seed re-runs of §5.5 and the UASpeech entries are Table 8. Percentages are
the minimising block as a fraction of that stack's depth. PER \emph{values} are not comparable across the two
columns --- only the depth at which each curve turns.

{\def\LTcaptype{none} % do not increment counter
\begin{longtable}[]{@{}
  >{\raggedright\arraybackslash}p{(\linewidth - 4\tabcolsep) * \real{0.3333}}
  >{\raggedright\arraybackslash}p{(\linewidth - 4\tabcolsep) * \real{0.3333}}
  >{\raggedright\arraybackslash}p{(\linewidth - 4\tabcolsep) * \real{0.3333}}@{}}
\toprule\noalign{}
\begin{minipage}[b]{\linewidth}\raggedright
Model
\end{minipage} & \begin{minipage}[b]{\linewidth}\raggedright
CDSD (Mandarin, read sentences)
\end{minipage} & \begin{minipage}[b]{\linewidth}\raggedright
UASpeech (English, isolated words)
\end{minipage} \\
\midrule\noalign{}
\endhead
\bottomrule\noalign{}
\endlastfoot
Qwen3-ASR (24 blocks) & layer 22 --- \textbf{92\%} & layer 20 --- \textbf{83\%} \\
Whisper-small (12 blocks) & layer 10 --- \textbf{83\%} & layer 11 --- \textbf{92\%} \\
\end{longtable}
}

\textbf{All four probe minima fall in an 83--92\% band, and this is measured rather than assumed.} Each model
keeps its late-maturing shape on the second corpus: the curve descends through the stack, turns within the
last sixth of it, and shows no mid-depth plateau in either corpus or either architecture. The argmin does
move between corpora --- Qwen3-ASR from 92\% to 83\%, Whisper-small from 83\% to 92\% --- but by two block
positions and one, \textbf{no further than it moves between seeds on a single corpus}. What is constant is the
band in which the minimum falls, the absence of a mid-stack plateau, and the ordering of the three
proportional bands --- not the minimising layer.

\textbf{What this licenses.} Placement varies across the grid of §6.9 while the probe curves do not vary in
any respect that could track it, and \textbf{a quantity that is near-constant across the design cannot explain
an outcome that changes across it} --- which closes the remaining escape route, that a corpus-matched probe
would have predicted the band where an inherited one did not. These curves also fix the scope of the
strongest candidate rule (§8): \emph{adapt where band-mean PER is lowest} picks \texttt{upper} for both models on
UASpeech, where both prefer \texttt{all} (Tables 14 and 22).

\textbf{Two caveats cut against reading these curves too finely.} \emph{The argmin is unstable across seeds}, which
is what a flat late plateau produces: seed 0 alone places Qwen3-ASR's minimum at layer 23 and
Whisper-small's at layer 9, against three-seed means of 20 and 11 --- further than either moves between
corpora, and why the paper reports \textbf{bands rather than argmin layers} throughout. \emph{And Whisper-small's
late curve is markedly less flat than Qwen3-ASR's}: its layer 12 rises to 69.79 from layer 11's 53.86,
\textbf{+15.93 pp} in a single block, while Qwen3-ASR's entire upper band spans 2.45 pp.~That is the one
respect in which the two curves genuinely differ in shape, and it rescues no placement rule. One
irregularity belongs in the record: Whisper-small's seed SDs are elevated at layers 5--7 (\textbf{3.91, 4.96 and
4.81 pp}) against a median of 0.62 pp elsewhere, so that band's mean (64.98) should be read with that in
mind.

\begin{center}\rule{0.5\linewidth}{0.5pt}\end{center}

\section{6. Adaptation results}\label{adaptation-results}

\subsection{6.1 The four-band sweep}\label{the-four-band-sweep}

\textbf{Table 10.} CER (\%) on 5,388 held-out utterances from 6 unseen speakers, \textbf{mean ± SD over three
training seeds (42/43/44)}; the unadapted base involves no training and is a single deterministic
evaluation, reproduced identically on all three passes. Three bands at an identical
4.19 M trainable-parameter budget; \texttt{all} at 12.58 M. Best in \textbf{bold}. Severity labels are
CER-derived rather than clinical (§9, Limitation 13).

{\def\LTcaptype{none} % do not increment counter
\begin{longtable}[]{@{}
  >{\raggedright\arraybackslash}p{(\linewidth - 10\tabcolsep) * \real{0.1667}}
  >{\raggedright\arraybackslash}p{(\linewidth - 10\tabcolsep) * \real{0.1667}}
  >{\raggedright\arraybackslash}p{(\linewidth - 10\tabcolsep) * \real{0.1667}}
  >{\raggedright\arraybackslash}p{(\linewidth - 10\tabcolsep) * \real{0.1667}}
  >{\raggedright\arraybackslash}p{(\linewidth - 10\tabcolsep) * \real{0.1667}}
  >{\raggedright\arraybackslash}p{(\linewidth - 10\tabcolsep) * \real{0.1667}}@{}}
\toprule\noalign{}
\begin{minipage}[b]{\linewidth}\raggedright
Variant
\end{minipage} & \begin{minipage}[b]{\linewidth}\raggedright
Blocks
\end{minipage} & \begin{minipage}[b]{\linewidth}\raggedright
Params
\end{minipage} & \begin{minipage}[b]{\linewidth}\raggedright
Micro CER
\end{minipage} & \begin{minipage}[b]{\linewidth}\raggedright
Mid-difficulty
\end{minipage} & \begin{minipage}[b]{\linewidth}\raggedright
High-difficulty
\end{minipage} \\
\midrule\noalign{}
\endhead
\bottomrule\noalign{}
\endlastfoot
base (no adaptation) & --- & 0 & 32.06 & 23.91 & 74.55 \\
\texttt{lower} & 1--8 & 4.19 M & 35.52 ± 1.15 & 28.79 ± 1.38 & 70.63 ± 0.35 \\
\texttt{middle} & 9--16 & 4.19 M & 30.71 ± 0.31 & 23.72 ± 0.43 & 67.14 ± 0.36 \\
\textbf{\texttt{upper}} & \textbf{17--24} & \textbf{4.19 M} & \textbf{29.88 ± 0.36} & \textbf{22.76 ± 0.40} & 66.96 ± 0.14 \\
\texttt{all} & 1--24 & 12.58 M & 31.19 ± 0.37 & 24.34 ± 0.41 & \textbf{66.87 ± 0.25} \\
\end{longtable}
}

\textbf{Ordering: \texttt{upper} \textless{} \texttt{middle} \textless{} \texttt{all} \textless{} base \textless{} \texttt{lower}, and \texttt{upper} is the best band in every seed.}
The prediction pre-specified from §5.1 partly holds. \texttt{upper} beats adapting the whole encoder while
training one third as many parameters, with an interval excluding zero (Table 12). It also beats
\texttt{middle} --- the proportional analogue of the band the prior work's probe pointed it to --- in all three
seeds, but \textbf{that margin is not resolved once optimisation stochasticity is included}, and we report it
as directional rather than established (§6.1.1). \textbf{The \texttt{upper}-beats-\texttt{all} sentence does not, on its
own, isolate placement}, since \texttt{upper} and \texttt{all} differ in where the adapters sit \emph{and} in how much
capacity they deploy; §6.7 separates the two with a placement × capacity 2×2.

\textbf{Table 11.} Per-speaker CER (\%), mean ± SD over the same three seeds. Best variant per speaker in
\textbf{bold}.

{\def\LTcaptype{none} % do not increment counter
\begin{longtable}[]{@{}
  >{\raggedright\arraybackslash}p{(\linewidth - 10\tabcolsep) * \real{0.1667}}
  >{\raggedright\arraybackslash}p{(\linewidth - 10\tabcolsep) * \real{0.1667}}
  >{\raggedright\arraybackslash}p{(\linewidth - 10\tabcolsep) * \real{0.1667}}
  >{\raggedright\arraybackslash}p{(\linewidth - 10\tabcolsep) * \real{0.1667}}
  >{\raggedright\arraybackslash}p{(\linewidth - 10\tabcolsep) * \real{0.1667}}
  >{\raggedright\arraybackslash}p{(\linewidth - 10\tabcolsep) * \real{0.1667}}@{}}
\toprule\noalign{}
\begin{minipage}[b]{\linewidth}\raggedright
Speaker
\end{minipage} & \begin{minipage}[b]{\linewidth}\raggedright
base
\end{minipage} & \begin{minipage}[b]{\linewidth}\raggedright
\texttt{lower}
\end{minipage} & \begin{minipage}[b]{\linewidth}\raggedright
\texttt{middle}
\end{minipage} & \begin{minipage}[b]{\linewidth}\raggedright
\texttt{upper}
\end{minipage} & \begin{minipage}[b]{\linewidth}\raggedright
\texttt{all}
\end{minipage} \\
\midrule\noalign{}
\endhead
\bottomrule\noalign{}
\endlastfoot
CDSD\_01 & 13.21 & 10.55 ± 0.32 & 10.56 ± 0.31 & \textbf{9.02 ± 0.11} & 11.46 ± 0.35 \\
CDSD\_03 & 53.29 & 49.74 ± 0.54 & 44.42 ± 0.46 & 44.32 ± 0.53 & \textbf{43.20 ± 0.63} \\
CDSD\_06 & 91.78 & 87.56 ± 0.46 & 85.55 ± 0.79 & \textbf{85.30 ± 0.26} & 86.04 ± 0.64 \\
CDSD\_08 & \textbf{39.42} & 59.06 ± 2.24 & 48.90 ± 1.73 & 51.42 ± 1.85 & 52.47 ± 1.94 \\
CDSD\_29 & 27.38 & 23.33 ± 0.19 & 21.98 ± 0.35 & \textbf{20.03 ± 0.22} & 22.23 ± 0.14 \\
CDSD\_38 & 16.87 & 25.26 ± 3.42 & 15.94 ± 0.34 & \textbf{13.50 ± 0.10} & 14.06 ± 0.47 \\
\textbf{micro} & 32.06 & 35.52 ± 1.15 & 30.71 ± 0.31 & \textbf{29.88 ± 0.36} & 31.19 ± 0.37 \\
\end{longtable}
}

\textbf{Table 12.} Paired \textbf{speaker × seed} bootstrap, B = 10,000: speakers are resampled with replacement
and a training seed is drawn per replicate, with each contrast formed \emph{within} a seed so that a seed's
overall difficulty cancels. This is the same estimator as Tables 15 and 23, and it replaces the
speaker-only bootstrap used in an earlier version of this table --- see §6.1.1 for what changed and why.
``Seed SD'' is the standard deviation of the contrast across the three seeds.

\noindent{\begin{tabular}{@{}
  >{\raggedright\arraybackslash}p{(\linewidth - 8\tabcolsep) * \real{0.2000}}
  >{\raggedright\arraybackslash}p{(\linewidth - 8\tabcolsep) * \real{0.2000}}
  >{\raggedright\arraybackslash}p{(\linewidth - 8\tabcolsep) * \real{0.2000}}
  >{\raggedright\arraybackslash}p{(\linewidth - 8\tabcolsep) * \real{0.2000}}
  >{\raggedright\arraybackslash}p{(\linewidth - 8\tabcolsep) * \real{0.2000}}@{}}
\toprule\noalign{}
\begin{minipage}[b]{\linewidth}\raggedright
Comparison
\end{minipage} & \begin{minipage}[b]{\linewidth}\raggedright
Δ CER (pp)
\end{minipage} & \begin{minipage}[b]{\linewidth}\raggedright
Seed SD
\end{minipage} & \begin{minipage}[b]{\linewidth}\raggedright
95\% CI
\end{minipage} & \begin{minipage}[b]{\linewidth}\raggedright
Verdict
\end{minipage} \\
\midrule\noalign{}
\texttt{upper} − \texttt{lower} & \textbf{−5.64} & 1.27 & [−10.87, −2.43] & excludes 0 \\
\texttt{upper} − \texttt{all} & \textbf{−1.31} & 0.16 & [−2.16, −0.25] & excludes 0 \\
\texttt{middle} − \texttt{all} & −0.48 & 0.65 & [−3.13, +1.10] & \textbf{includes 0} \\
\texttt{upper} − \texttt{middle} & −0.83 & 0.59 & [−2.06, \textbf{+1.86}] & \textbf{includes 0} \\
\texttt{upper} − base & −2.18 & 0.36 & [−6.78, +4.60] & \textbf{includes 0} \\
\texttt{middle} − base & −1.35 & 0.31 & [−5.51, +3.80] & includes 0 \\
\texttt{all} − base & −0.87 & 0.37 & [−5.63, +5.93] & includes 0 \\
\texttt{lower} − base & +3.46 & 1.15 & [−3.74, +12.26] & includes 0 \\
\bottomrule\noalign{}
\end{tabular}\hfill
}

\textbf{These halves say different things, and the paper's claim is narrower than the point estimates
suggest.} \emph{Among adaptation strategies}, \texttt{upper} is the best band in every seed, and two of its three
contrasts resolve: \texttt{upper} − \texttt{all} = −1.31, CI [−2.16, −0.25], and \texttt{upper} − \texttt{lower} = −5.64, CI
[−10.87, −2.43]. \textbf{\texttt{upper} − \texttt{middle} does not resolve}: −0.83, CI [−2.06, +1.86]. It is negative in all
three seeds (−0.26 / −1.43 / −0.81), so the direction is consistent, but the magnitude varies by a factor
of 5.5 and the interval includes zero once the seed is resampled alongside the speakers. \texttt{middle} − \texttt{all}
does not resolve either, and its sign flips across seeds. \emph{Against no adaptation at all}, \textbf{nothing here
is significant}: \texttt{upper} − base is −2.18 pp with a CI spanning [−6.78, +4.60]. The cause is CDSD\_08,
which regresses under every variant and so cancels in adapted-versus-adapted contrasts while dominating
adapted-versus-base ones. A sign test on \texttt{upper} \textless{} \texttt{middle} does \emph{not} reach unanimity: it gives
4/6 speakers at seed 42 and 5/6 at seeds 43 and 44, and 5/6 on the three-seed means, CDSD\_08 being the
dissenter in each case (Table 11). Two-sided that is p = 0.219 by the sign test and p = 0.438 by the
exact paired sign-flip test (Table 26). Unanimity does hold for \texttt{upper} \textless{} \texttt{lower} --- 6/6 in every
seed, p = 0.031 under both tests. Agreement across training seeds is not unanimity across speakers.

\textbf{The defensible claim from CDSD alone is therefore narrow: given that you are going to place a low-rank
adapter in this encoder, the upper band is a better target than the lower band and than all encoder
blocks. It is also better than the middle band in every seed we ran, but that particular margin is not
resolved at this sample size. Whether this adaptation is worth doing at all is not established by this
test set} --- though it is established on the second corpus in §6.5.

\subsubsection{6.1.1 These are three-seed numbers, and one earlier conclusion does not survive them}\label{these-are-three-seed-numbers-and-one-earlier-conclusion-does-not-survive-them}

An earlier version of this section reported a single training run per band, with intervals from a
speaker-only bootstrap. That estimator quantifies which six speakers happened to be held out; it is
silent about which random initialisation and batch order the adapter happened to get. Since the two
headline CDSD contrasts were around 1.5 pp, that omission was load-bearing, and a reviewer was right to
say so.

Each band was therefore retrained at seeds 42, 43 and 44 under an otherwise identical recipe, and every
CDSD interval in this paper now comes from the speaker × seed estimator. \textbf{One conclusion changed.}
\texttt{upper} − \texttt{middle} was previously reported as −1.52 with CI [−2.27, −0.54], excluding zero; it is
\textbf{−0.83 with CI [−2.06, +1.86], which includes zero}. We no longer claim \texttt{upper} beats \texttt{middle} on
CDSD, only that it did so in all three seeds we ran.

The claim the paper's first contribution rests on is unaffected. \texttt{upper} − \texttt{all} moves from −1.47 to
−1.31 and stays resolved. The parameter-matched 2×2 of §6.7 gives the same equal-rank pair as −1.26,
consistent in sign and magnitude; note that the 2×2's own \emph{budget-matched} contrasts do not resolve
(§6.7), so it corroborates this contrast without independently establishing a placement effect.
The seed sensitivities of the two contrasts differ by more than a factor of three --- seed SD 0.16 for
\texttt{upper} − \texttt{all} against 0.59 for \texttt{upper} − \texttt{middle} --- so the contrast that has independent corroboration
elsewhere in the paper is also the one that is stable under retraining, and the contrast that had none is
the one that dissolves. We report that as a caution about single-run band sweeps generally, ours
included: a 1.5 pp difference between two bands is within what a seed can move.

The \texttt{upper} \textless{} \texttt{all} result contradicts \citet{lin2024peftspeech}, who concluded that uniform insertion into all
layers was not beaten in self-supervised speech encoders. Here adapters confined to a third of the blocks
beat insertion into all of them, with a CI excluding zero, stable under leave-one-out --- a second instance of this paper's thesis that whether selective
placement pays is itself encoder-dependent. \textbf{This contrast rests on CDSD alone}; UASpeech does not
resolve it in either direction, so it, the 2×2 built on it, and the cross-model comparison of §6.6 are all
single-corpus.

\subsection{6.2 Leave-one-speaker-out sensitivity}\label{leave-one-speaker-out-sensitivity}

\textbf{Table 13.} Micro CER (\%) recomputed over the remaining five speakers, mean ± SD over seeds 42/43/44,
with the two headline contrasts. The base column involves no training and is a single evaluation.

{\def\LTcaptype{none} % do not increment counter
\begin{longtable}[]{@{}
  >{\raggedright\arraybackslash}p{(\linewidth - 12\tabcolsep) * \real{0.1428}}
  >{\raggedright\arraybackslash}p{(\linewidth - 12\tabcolsep) * \real{0.1428}}
  >{\raggedright\arraybackslash}p{(\linewidth - 12\tabcolsep) * \real{0.1428}}
  >{\raggedright\arraybackslash}p{(\linewidth - 12\tabcolsep) * \real{0.1428}}
  >{\raggedright\arraybackslash}p{(\linewidth - 12\tabcolsep) * \real{0.1428}}
  >{\raggedright\arraybackslash}p{(\linewidth - 12\tabcolsep) * \real{0.1428}}
  >{\raggedright\arraybackslash}p{(\linewidth - 12\tabcolsep) * \real{0.1428}}@{}}
\toprule\noalign{}
\begin{minipage}[b]{\linewidth}\raggedright
Excluded
\end{minipage} & \begin{minipage}[b]{\linewidth}\raggedright
base
\end{minipage} & \begin{minipage}[b]{\linewidth}\raggedright
\texttt{upper}
\end{minipage} & \begin{minipage}[b]{\linewidth}\raggedright
\texttt{middle}
\end{minipage} & \begin{minipage}[b]{\linewidth}\raggedright
\texttt{all}
\end{minipage} & \begin{minipage}[b]{\linewidth}\raggedright
\textbf{\texttt{upper}}\\−base
\end{minipage} & \begin{minipage}[b]{\linewidth}\raggedright
\textbf{\texttt{upper}}\\−\texttt{middle}
\end{minipage} \\
\midrule\noalign{}
\endhead
\bottomrule\noalign{}
\endlastfoot
none & 32.06 & 29.88 ± 0.36 & 30.71 ± 0.31 & 31.19 ± 0.37 & \textbf{−2.18 ± 0.36} & −0.83 ± 0.59 \\
CDSD\_01 & 37.02 & 35.37 ± 0.48 & 36.01 ± 0.33 & 36.38 ± 0.38 & −1.65 ± 0.48 & −0.65 ± 0.69 \\
CDSD\_03 & 30.41 & 28.76 ± 0.35 & 29.65 ± 0.32 & 30.25 ± 0.36 & −1.66 ± 0.35 & −0.89 ± 0.56 \\
CDSD\_06 & 26.23 & 24.47 ± 0.41 & 25.36 ± 0.41 & 25.83 ± 0.41 & −1.76 ± 0.41 & −0.89 ± 0.74 \\
\textbf{\small CDSD\_08} & 30.32 & 24.79 ± 0.04 & 26.42 ± 0.03 & 26.16 ± 0.03 & \textbf{−5.53 ± 0.04} & −1.62 ± 0.02 \\
CDSD\_29 & 33.41 & 32.73 ± 0.43 & 33.24 ± 0.33 & 33.78 ± 0.51 & −0.69 ± 0.43 & −0.51 ± 0.69 \\
CDSD\_38 & 36.23 & 34.37 ± 0.48 & 34.76 ± 0.49 & 35.89 ± 0.57 & −1.86 ± 0.48 & −0.39 ± 0.85 \\
\end{longtable}
}

\textbf{\texttt{upper} − \texttt{middle} keeps its sign under every leave-one-out subset} (−0.39 to −1.62 pp), so no single
speaker drives it --- but its seed SD is comparable to its magnitude in every subset, which is the same
picture the full bootstrap gives (§6.1.1) and the reason we do not claim it as resolved. \texttt{upper} − base
swings from −0.69 to −5.53 pp and is entirely hostage to test-set composition. That is also why CDSD\_08
is retained rather than excluded: dropping it would move the apparent benefit of adaptation from −2.18 to
−5.53 pp, a 2.5× inflation obtained by removing the one speaker that disagrees. Note also that excluding
CDSD\_08 collapses the seed SD to ±0.04 --- nearly all seed-to-seed variation on this test set comes from
that one speaker.

\subsection{\texorpdfstring{6.3 Behavioural corroboration: \texttt{middle} barely moves the model}{6.3 Behavioural corroboration: middle barely moves the model}}\label{behavioural-corroboration-middle-barely-moves-the-model}

The fraction of held-out utterances whose hypothesis is \textbf{byte-identical} to the unadapted model's:

\noindent{\begin{tabular}{@{}lll@{}}
\toprule\noalign{}
Variant & Identical to base & \% \\
\midrule\noalign{}
\texttt{middle} & 1,217 / 5,388 & \textbf{22.6\%} \\
\texttt{all} & 579 / 5,388 & 10.7\% \\
\texttt{upper} & 129 / 5,388 & \textbf{2.4\%} \\
\bottomrule\noalign{}
\end{tabular}\hfill
}

Adapting blocks 9--16 left more than four times as many outputs untouched as adapting 17--24 did --- an
independent, non-metric confirmation that the upper third of this encoder is where the consequential
work happens.

\subsection{6.4 Difficulty tiers are not uniform}\label{difficulty-tiers-are-not-uniform}

On \textbf{moderate} speakers, on three-seed means, \texttt{upper} (22.76) and \texttt{middle} (23.72) both come in under
base (23.91) while \texttt{all} (24.34) does not; none of those gaps is resolved at six speakers, so this is a
numerical ordering rather than a demonstrated effect. On \textbf{severe} speakers all three deep variants
improve substantially over base (66.87--67.14 against 74.55), and \texttt{all} is marginally best. (An earlier
version of this paragraph quoted single-run figures --- 22.96, 24.63, 24.77 --- which the three-seed re-run
of §6.1.1 superseded.) \texttt{upper}'s aggregate
advantage is therefore carried by moderate speakers --- an observation rather than a result, at six test
speakers with CER-derived severity labels.

\subsection{6.5 Second corpus: the sweep replicated on English isolated words}\label{second-corpus-the-sweep-replicated-on-english-isolated-words}

Everything above is one language, one corpus, one content type, and CER-derived severity labels. We
repeated the identical four-band sweep on \textbf{UASpeech} \citep{kim2008uaspeech}, which differs from CDSD on
every one of those axes: English rather than Mandarin, \textbf{449 isolated words} rather than read news
sentences, and \textbf{clinical} severity labels. Recipe, band definitions, parameter budget, LoRA
configuration, integrity gates and decode path are unchanged.

\textbf{Corpus accounting, so the split can be reconstructed.} The original UASpeech release describes 19
dysarthric speakers and 455 distinct words; the current distribution contains 16. \textbf{Our copy contains
15}: F02, F03, F04, F05, M01, M04, M05, M07, M08, M09, M10, M11, M12, M14, M16 --- the current 16 minus
\textbf{M06}. Our manifest covers \textbf{449 distinct word types}, the shortfall following from the missing speaker
and from utterances dropped by the audio-integrity check.

Splits are speaker-disjoint: 8 train / 2 validation / \textbf{5 test} speakers (3,798 utterances), stratified
on severity \textbf{and} gender; an initial severity-only stratification produced an all-male test set and was
corrected before launch. Because every UASpeech speaker reads the same 449-word list, a text-disjoint
split is impossible by construction --- a closed-vocabulary task by design. Scoring is case-folded. Every
adapted condition was trained \textbf{three times} with seeds 42/43/44, reported as the seed mean ± SD.

\textbf{Table 14.} UASpeech, 3,798 held-out utterances from 5 unseen speakers. Adapted rows are the mean of
three seeds ± SD. Best in \textbf{bold}.

{\def\LTcaptype{none} % do not increment counter
\begin{longtable}[]{@{}lllll@{}}
\toprule\noalign{}
Variant & Blocks & Params & CER (mean ± SD) & seed range \\
\midrule\noalign{}
\endhead
\bottomrule\noalign{}
\endlastfoot
base (no adaptation) & --- & 0 & 41.95 (single run) & --- \\
\texttt{lower} & 1--8 & 4.19 M & 51.85 ± 0.38 & 51.60 -- 52.29 \\
\texttt{middle} & 9--16 & 4.19 M & 36.62 ± 0.12 & 36.47 -- 36.70 \\
\texttt{upper} & 17--24 & 4.19 M & 34.41 ± 0.32 & 34.10 -- 34.75 \\
\texttt{all} & 1--24 & 12.58 M & \textbf{32.75 ± 0.30} & 32.50 -- 33.08 \\
\end{longtable}
}

\textbf{Seed variance is small: 0.12--0.38 pp SD, against band separations of 2--17 pp.} Retraining moves a
band's CER by a few tenths of a point; changing which band is adapted moves it by several points.

\textbf{Table 15.} UASpeech contrasts with BOTH sources of uncertainty. ``Seed range'' is the paired contrast
recomputed per seed with all 5 speakers held fixed; ``speaker × seed'' resamples speakers and seeds
together (B = 10,000). Contrasts are formed within a seed before combining, so a seed's overall
difficulty cancels.

\noindent{\begin{tabular}{@{}
  >{\raggedright\arraybackslash}p{(\linewidth - 8\tabcolsep) * \real{0.2000}}
  >{\raggedright\arraybackslash}p{(\linewidth - 8\tabcolsep) * \real{0.2000}}
  >{\raggedright\arraybackslash}p{(\linewidth - 8\tabcolsep) * \real{0.2000}}
  >{\raggedright\arraybackslash}p{(\linewidth - 8\tabcolsep) * \real{0.2000}}
  >{\raggedright\arraybackslash}p{(\linewidth - 8\tabcolsep) * \real{0.2000}}@{}}
\toprule\noalign{}
\begin{minipage}[b]{\linewidth}\raggedright
Contrast
\end{minipage} & \begin{minipage}[b]{\linewidth}\raggedright
Mean (pp)
\end{minipage} & \begin{minipage}[b]{\linewidth}\raggedright
Seed range
\end{minipage} & \begin{minipage}[b]{\linewidth}\raggedright
Speaker × seed 95\% CI
\end{minipage} & \begin{minipage}[b]{\linewidth}\raggedright
Sign across seeds
\end{minipage} \\
\midrule\noalign{}
\texttt{upper} − \texttt{middle} & \textbf{−2.21} & [−2.58, −1.73] & \textbf{[−4.37, −0.39]} excludes 0 & stable \\
\texttt{upper} − \texttt{lower} & \textbf{−17.44} & [−17.91, −16.85] & \textbf{[−24.44, −10.05]} excludes 0 & stable \\
\texttt{lower} − \texttt{middle} & \textbf{+15.23} & [+14.99, +15.59] & \textbf{[+8.79, +21.33]} excludes 0 & stable \\
\texttt{all} − \texttt{middle} & \textbf{−3.87} & [−4.20, −3.59] & \textbf{[−6.72, −1.24]} excludes 0 & stable \\
\texttt{upper} − \texttt{all} & +1.66 & [+1.01, +2.09] & [−0.51, +4.46] spans 0 & stable \\
\bottomrule\noalign{}
\end{tabular}\hfill
}

\textbf{The primary contrast survives the addition of seed variance:} \texttt{upper} beats \texttt{middle} by 2.21 pp with a
speaker × seed interval excluding zero, and the sign is the same in all three seeds. \textbf{\texttt{upper} − \texttt{all} is
directionally consistent but not resolved} --- all three seeds favour \texttt{all}, yet the speaker × seed
interval spans zero, which is what a 5-speaker test set should be expected to produce at this effect size.

\textbf{Speakers, not seeds, are the binding source of uncertainty, which reverses the usual worry.} On
\texttt{upper} − \texttt{middle} the contrast varies by 0.85 pp across the three seeds with all five speakers held
fixed, while resampling speakers as well widens the interval to 3.98 pp.~(Those two are different
statistics, so we do not quote their quotient as an uncertainty ratio.) Running more seeds would buy this
study little; more test speakers would buy it a great deal.

\textbf{Table 16.} Paired speaker bootstrap on both corpora, B = 10,000, speakers resampled with replacement.
A single shared set of resamples per corpus scores every variant, so speaker difficulty cancels within a
replicate. Negative Δ favours the first variant.

{\def\LTcaptype{none} % do not increment counter
\begin{longtable}[]{@{}
  >{\raggedright\arraybackslash}p{(\linewidth - 4\tabcolsep) * \real{0.3333}}
  >{\raggedright\arraybackslash}p{(\linewidth - 4\tabcolsep) * \real{0.3333}}
  >{\raggedright\arraybackslash}p{(\linewidth - 4\tabcolsep) * \real{0.3333}}@{}}
\toprule\noalign{}
\begin{minipage}[b]{\linewidth}\raggedright
Comparison
\end{minipage} & \begin{minipage}[b]{\linewidth}\raggedright
CDSD Δ (95\% CI)
\end{minipage} & \begin{minipage}[b]{\linewidth}\raggedright
UASpeech Δ (95\% CI)
\end{minipage} \\
\midrule\noalign{}
\endhead
\bottomrule\noalign{}
\endlastfoot
\texttt{upper} − \texttt{middle} & −0.83 [−2.06, \textbf{+1.86}] \emph{n.s.} & −2.21 [−4.37, −0.39] \\
\texttt{upper} − \texttt{lower} & −5.64 [−10.87, −2.43] & −17.44 [−24.44, −10.05] \\
\texttt{lower} − \texttt{middle} & +4.81 [+1.10, +10.23] & +15.23 [+8.79, +21.33] \\
\texttt{upper} − \texttt{all} & −1.31 [−2.16, −0.25] & +1.66 [−0.51, +4.46] \emph{n.s.} \\
\texttt{all} − \texttt{middle} & +0.48 [−1.10, +3.13] \emph{n.s.} & −3.87 [−6.72, −1.24] \\
\texttt{upper} − base & −2.18 [−6.78, +4.60] \emph{n.s.} & −7.54 [−11.53, −3.83] \\
\texttt{all} − base & −0.87 [−5.63, +5.93] \emph{n.s.} & −9.20 [−14.93, −4.37] \\
\texttt{middle} − base & −1.35 [−5.51, +3.80] \emph{n.s.} & −5.33 [−8.25, −2.88] \\
\texttt{lower} − base & +3.46 [−3.74, +12.26] \emph{n.s.} & +9.90 [+5.17, +14.97] \\
\end{longtable}
}

Intervals marked \emph{n.s.} include zero. \textbf{On UASpeech every contrast resolves except \texttt{upper} − \texttt{all}},
which is the one this paper most wants to read and the one that stays open there. The CDSD column now
carries speaker × seed intervals throughout (§6.1.1). Marginal per-variant CIs are omitted deliberately:
at 5--6 speakers they span 33--54 pp and say nothing.

\textbf{On what is primary, stated plainly because an earlier version got it wrong.} That version declared
\texttt{upper} − \texttt{middle} the single primary contrast --- the one the probe predicted before any adapter was
trained --- and then rested its headline on \texttt{upper} − \texttt{all}, which had not been declared. We now nominate
the \textbf{cross-model interaction} as primary (§6.9.1) \textbf{on the grounds of its role}: it is what
Contribution 1 asserts and the question this paper sets out to answer. That it also happens to be the
only quantity here resolving under micro CER, speaker-equal weighting and an exact test alike is a
property we report, not the reason for the designation --- choosing a primary contrast because it came
out well would be selection on outcome.
\texttt{upper} − \texttt{middle} and \texttt{upper} − \texttt{all} are reported as the two pre-specified main effects. Every other
entry in Table 16 is \textbf{secondary and exploratory}, reported without multiplicity adjustment across the
nine contrasts per corpus; with nine contrasts, some intervals excluding zero are expected by chance.

\textbf{Two results replicate with intervals excluding zero on both corpora}: \texttt{upper} beats \texttt{lower}, and
\texttt{lower} is worse than \texttt{middle}. A third --- \texttt{upper} beats \texttt{middle} --- resolves on UASpeech (−2.21,
CI [−4.37, −0.39]) but \textbf{not} on CDSD once training seeds are resampled (−0.83, CI [−2.06, +1.86];
§6.1.1), though the CDSD point estimate is on the same side in all three seeds. We therefore report
\texttt{upper} \textgreater{} \texttt{middle} as directionally consistent across both corpora and resolved on one of them. \textbf{No contrast is resolved
in opposite directions on the two corpora.} \textbf{They are not in agreement either}: the \texttt{upper} − \texttt{all}
\emph{point estimate} reverses sign (−1.31 pp on CDSD, +1.66 pp on UASpeech), which is why we treat the
parameter-efficiency result as single-corpus --- an absence of evidence at five test speakers, licensing no
reverse claim.

\textbf{UASpeech settles the question §6.1 could not.} On CDSD every adaptation-versus-base contrast included
zero; on UASpeech all four resolve, so adapting the upper band, all encoder blocks, or even the middle
band each beats no adaptation, and \textbf{adapting the lower band is reliably worse than not adapting at all}
(+9.90 pp CER, +16.22 pp WER, and more than double the mild-speaker CER: 8.84 → 19.68). Two further
observations: \texttt{all} is the best variant on UASpeech and second-best on CDSD, so uniform insertion is a
strong and safe default even where targeted placement wins; and \texttt{lower} inflicts its greatest relative
damage on the \textbf{mild} speakers, whose base CER of 8.84 an adapter has least excuse to disturb.

\begin{center}\rule{0.5\linewidth}{0.5pt}\end{center}

\subsection{6.6 Second architecture: the identical sweep on Whisper-small}\label{second-architecture-the-identical-sweep-on-whisper-small}

§6.5 varies the corpus while holding the architecture fixed; this section does the reverse, and it is the
comparison the paper's first contribution rests on. Its conclusions should be taken as CDSD-specific until
§6.9. The sweep runs on Whisper-small using the \textbf{same} CDSD manifest, speaker split, integrity gate,
LoRA recipe, seed discipline and evaluation code as §6.1, with bands defined proportionally over its 12
blocks (\texttt{lower} 1--4, \texttt{middle} 5--8, \texttt{upper} 9--12, \texttt{all} 1--12).

\textbf{Table 17.} CER (\%) on the same 5,388 held-out utterances from the same 6 unseen speakers, \textbf{mean ± SD
over three training seeds (42/43/44)}; the unadapted base involves no training and reproduced 77.567
identically on all three passes.

{\def\LTcaptype{none} % do not increment counter
\begin{longtable}[]{@{}lll@{}}
\toprule\noalign{}
Variant & CER & vs unadapted \\
\midrule\noalign{}
\endhead
\bottomrule\noalign{}
\endlastfoot
unadapted base & 77.57 & --- \\
\textbf{\texttt{all} (1--12)} & \textbf{53.56 ± 0.01} & \textbf{−24.01} \\
\texttt{middle} (5--8) & 55.16 ± 0.13 & −22.41 \\
\texttt{lower} (1--4) & 56.28 ± 0.07 & −21.29 \\
\texttt{upper} (9--12) & 56.95 ± 0.08 & −20.61 \\
\end{longtable}
}

\textbf{Table 18.} Whisper-small band contrasts on CDSD, three seeds. Three estimators are reported for each
contrast: the speaker × seed bootstrap used throughout the paper (\textbf{micro} CER, pooling edit counts over
resampled speakers), the same bootstrap on \textbf{macro} CER (speakers weighted equally), and an \textbf{exact
paired sign-flip} p-value over all 2⁶ = 64 speaker sign assignments. §6.10 explains why all three are
shown. Negative Δ favours the first-named variant.

{\def\LTcaptype{none} % do not increment counter
\begin{longtable}[]{@{}
  >{\raggedright\arraybackslash}p{(\linewidth - 10\tabcolsep) * \real{0.1667}}
  >{\raggedright\arraybackslash}p{(\linewidth - 10\tabcolsep) * \real{0.1667}}
  >{\raggedright\arraybackslash}p{(\linewidth - 10\tabcolsep) * \real{0.1667}}
  >{\raggedright\arraybackslash}p{(\linewidth - 10\tabcolsep) * \real{0.1667}}
  >{\raggedright\arraybackslash}p{(\linewidth - 10\tabcolsep) * \real{0.1667}}
  >{\raggedright\arraybackslash}p{(\linewidth - 10\tabcolsep) * \real{0.1667}}@{}}
\toprule\noalign{}
\begin{minipage}[b]{\linewidth}\raggedright
Contrast
\end{minipage} & \begin{minipage}[b]{\linewidth}\raggedright
Micro Δ
\end{minipage} & \begin{minipage}[b]{\linewidth}\raggedright
95\% CI
\end{minipage} & \begin{minipage}[b]{\linewidth}\raggedright
Macro Δ
\end{minipage} & \begin{minipage}[b]{\linewidth}\raggedright
95\% CI
\end{minipage} & \begin{minipage}[b]{\linewidth}\raggedright
Exact p
\end{minipage} \\
\midrule\noalign{}
\endhead
\bottomrule\noalign{}
\endlastfoot
\textbf{\texttt{all} − \texttt{upper}} & \textbf{−3.40} & [−5.03, −1.84] & \textbf{−3.63} & [−4.94, −2.14] & \textbf{0.031} \\
\texttt{all} − \texttt{lower} & \textbf{−2.72} & [−4.42, −1.29] & \textbf{−3.05} & [−4.71, −1.56] & \textbf{0.031} \\
\texttt{middle} − \texttt{upper} & −1.80 & [−4.38, \textbf{+0.27}] & \textbf{−2.01} & [−4.07, −0.04] & 0.156 \\
\texttt{all} − \texttt{middle} & −1.60 & [−3.24, \textbf{+0.46}] & −1.61 & [−3.05, \textbf{+0.11}] & 0.125 \\
\end{longtable}
}

\textbf{\texttt{all} − \texttt{upper} resolves under all three estimators}, and its exact p sits at the attainable floor
for six speakers --- every speaker favours \texttt{all}. It is not unique in that: \texttt{all} − \texttt{lower} in the same
table resolves under all three as well, as does \texttt{upper} − \texttt{lower} on Qwen3-ASR × CDSD (Table 26). What
is true of \texttt{all} − \texttt{upper} alone is that it is the \emph{placement} contrast this cell's conclusion rests on
and it survives every estimator, where its Qwen3-ASR counterpart does not. Seed SDs on this
cell are an order of magnitude smaller than on the Qwen3-ASR sweep (\texttt{all} 0.013, others 0.06--0.13,
against 0.31--1.15 pp): Whisper-small's adaptation outcome on CDSD is far less seed-sensitive.

\textbf{Figure 4} places the two sweeps side by side, each normalised to its own unadapted baseline, with
per-speaker points overlaid and the no-adaptation line marked.

\begin{figure}[htbp]
\centering
\includegraphics[width=\linewidth]{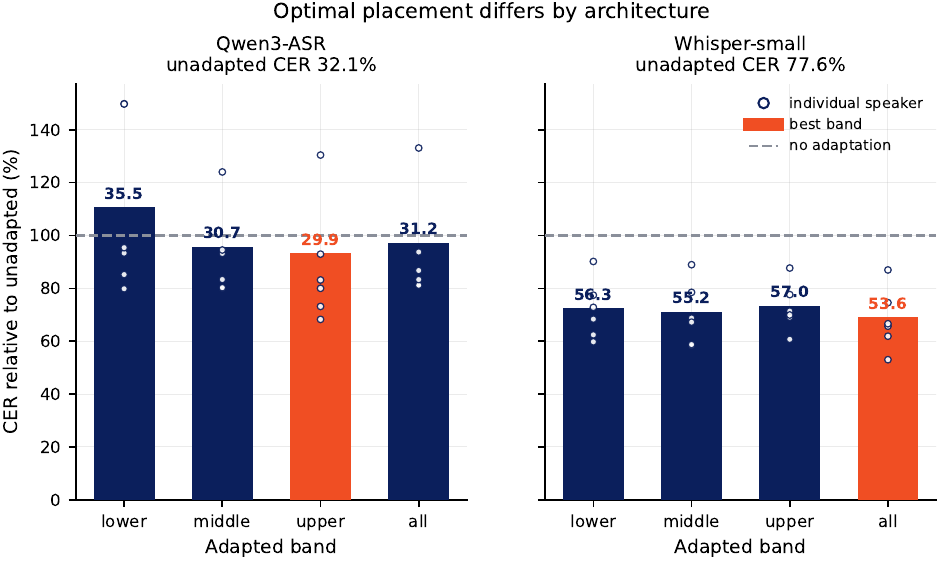}
\caption{Four-band LoRA sweep on CDSD, same manifest, split, gate, recipe, parameter budget and seed discipline, encoder swapped. CER is shown relative to each architecture's own unadapted baseline (32.06\% against 77.57\%), since a shared absolute axis would be unreadable; absolute CER is printed on each bar. Open circles are individual test speakers. The dashed line is no adaptation. The best band is orange: \texttt{upper} on Qwen3-ASR, \texttt{all} on Whisper-small, where \texttt{upper} ranks worst of the four. On Qwen3-ASR \texttt{lower} sits above the line --- worse than not adapting at all.}
\label{fig:fig4_band_placement}%
\end{figure}

\textbf{Alt text (Figure 4).} Two panels side by side, one per architecture, each plotting the four LoRA placement bands --- lower, middle, upper and all --- as change in CER relative to that architecture's own unadapted baseline, with error bars. On Qwen3-ASR the upper band gives the largest improvement and the lower band the smallest. On Whisper-small the ordering differs: all-block adaptation gives the largest improvement and the upper band the smallest. The two panels therefore rank the bands differently, which is the cross-model interaction the paper tests directly.

\textbf{What this establishes, stated no more strongly than the intervals allow.}

\begin{enumerate}
\def\labelenumi{\arabic{enumi}.}
\tightlist
\item
  \textbf{The Qwen3-ASR answer does not transfer, and neither model resolves this pair.} On Qwen3-ASR,
  \texttt{upper} − \texttt{middle} = −0.83, CI [−2.06, \textbf{+1.86}] --- directional in \texttt{upper}'s favour across all three
  seeds but \textbf{not resolved} (§6.1.1). On Whisper-small the same contrast in the same orientation is
  \textbf{+1.80, CI [−0.27, +4.38]} --- the opposite sign, and likewise unresolved under micro CER, though it
  does resolve under speaker-equal weighting (Table 18 lists it as \texttt{middle} − \texttt{upper} = −1.80). So the
  pair points in opposite directions on the two models while resolving on neither under the primary
  metric: suggestive of a sign flip, not a demonstration of one.
\item
  \textbf{\texttt{all} is best on Whisper-small, resolved against \texttt{upper} but not against \texttt{middle}.} \texttt{all} − \texttt{upper}
  = −3.40, CI [−5.03, −1.84]; \texttt{all} − \texttt{middle} = −1.60, CI [−3.24, +0.46], which spans zero. On
  Qwen3-ASR \texttt{upper} beat \texttt{all}, so the two architectures resolve the \texttt{upper}/\texttt{all} pair in opposite
  directions --- the strongest single piece of evidence in the paper that placement is not a property of
  the disorder, though on the evidence of the fourth cell specific to this corpus. The narrower claim
  that whole-encoder adaptation beats \emph{every} equal-budget band is not supported: \texttt{middle} is not
  separable from \texttt{all} here.
\item
  \textbf{\texttt{upper} --- the winner on Qwen3-ASR --- ranks worst of the four bands here.}
\item
  \textbf{The prior work's sufficiency claim does not hold on our split.} They report the 5--8 band within
  2.89\% relative of full-encoder adaptation in their v2 tables; we measure \texttt{middle} 3.0\% relatively worse than \texttt{all}, and cannot separate the two (Table 18).
\end{enumerate}

\textbf{A confound we cannot remove, and which bounds all of the above.} Whisper-small's unadapted CER on this
corpus is 77.57 against Qwen3-ASR's 32.06 --- still less than half as competent, though the gap is
considerably narrower than it first appeared. \textbf{This figure supersedes an earlier 103.57 that was partly
an artefact of our own decoding}, and the correction is documented in §6.6.1 because it changes what can
be claimed here. Under a repetition-controlled decode only one of the six test speakers still exceeds
100\% CER (135.1; it was 308.19), so the corpus-level ``insertions outnumber the reference'' signature of
hallucination does \textbf{not} hold for this model on this corpus, and we no longer rest any argument on it.
A less competent model may still improve from a different direction, visible in \texttt{lower}: it \emph{harms}
Qwen3-ASR (+3.46 pp) but helps Whisper-small (−21.29 pp). The two architectures therefore differ in
baseline competence as well as in depth, frame rate, pre-training corpus, scale and decoder type
(§9, Limitation 16). What this experiment does establish is that the difference is \textbf{real and not a
cross-paper artefact} --- and it falsifies the knee rule, since Whisper-small's PER minimum at layer 10
lies inside its \texttt{upper} band, which the rule predicts should be productive. It is the worst of the four.

\subsubsection{6.6.1 The unadapted Whisper baseline is decode-sensitive, and we report both settings}\label{the-unadapted-whisper-baseline-is-decode-sensitive-and-we-report-both-settings}

Our first measurement of Whisper-small's unadapted CDSD CER returned 103.57, and being above 100\% invites
reading it as evidence of hallucination. That number was produced with greedy search and a flat
200-token generation budget on utterances averaging 2.4 s --- ample room for a repetition loop, and every
looped character scores as an insertion. Re-running \textbf{every} Whisper condition in this paper with
repetition control (\texttt{no\_repeat\_ngram\_size} 3, beam 5) and a duration-scaled token cap gives the figures
reported throughout. A flat small cap is not the remedy either: CDSD reaches 18.3 s and UASpeech 19.5 s,
so a fixed budget tuned on short utterances would trade an insertion artefact for a deletion one.

The correction is strongly non-uniform, and in the direction the diagnosis predicts: the unadapted base
gains 26.01 pp while the adapted bands gain between 1.85 and 4.63 pp, because an adapted model already
emits well-formed output and has little looping to suppress. \textbf{No placement conclusion in this paper
changes}, on either corpus; what changes is the size of every gap involving an unadapted or weakly
adapted condition. Two verdicts do move, and both are reported at their corrected values: \texttt{all} −
\texttt{middle} here (was resolved, now spans zero) and \texttt{lower} − base on UASpeech CER (§6.8, was spanning zero,
now resolved). We report this rather than silently restating the numbers because the same decode default
is used widely, and a baseline inflated this way invites the conclusion that a model is broken when it is
partly the harness that is.

\begin{center}\rule{0.5\linewidth}{0.5pt}\end{center}

\subsection{6.7 Placement or capacity? A parameter-matched 2×2}\label{placement-or-capacity-a-parameter-matched-22}

Every \texttt{upper} versus \texttt{all} contrast reported so far varies two things at once: \texttt{upper} places adapters in
8 of 24 blocks and \texttt{all} in all 24, but both use r = 64, so \texttt{all} also carries three times the trainable
capacity. That leaves open the reading that \texttt{all} wins nothing because it is over-parameterised for 33,617
utterances. This subsection separates the two factors on CDSD × Qwen3-ASR, the cell in which \texttt{upper} beats
\texttt{all}.

\textbf{Design.} Placement × capacity, fully crossed, with rank chosen per cell to equalise the
trainable-parameter count along the capacity axis. Each adapted block contributes four LoRA modules of
r × 2 × 1024 parameters, so a configuration's budget is proportional to (blocks × rank):

\begin{itemize}
\tightlist
\item
  \textbf{High capacity, exact match.} 8 blocks × r = 192 and 24 blocks × r = 64 both give 1,536 rank-blocks
  and \textbf{12,582,912} trainable parameters --- an exact match, not an approximate one.
\item
  \textbf{Low capacity, 98.4\% match.} 8 blocks × r = 64 gives 4,194,304 parameters; 24 blocks × r = 21 gives
  4,128,768. Integer rank makes an exact match impossible at 24 blocks.
\end{itemize}

\textbf{All four cells were trained at seeds 42, 43 and 44 with α = 2r throughout}, with manifest, split,
gate, optimiser, schedule, epoch count, merge procedure and decode path identical to §6.1 and to each
other. Two design points matter and are stated because an earlier version of this section got the second
one wrong.

\emph{Seeds.} An earlier version ran this 2×2 at seed 42 only. Since its headline margin was about 1.5 pp ---
inside what a seed moves on this corpus (§6.1.1) --- every cell was retrained at three seeds and every
interval below is a speaker × seed bootstrap.

\emph{Scaling.} An earlier version left LoRA α at the default 128 for \textbf{every} rank, so effective scaling
α/r was \textbf{not} matched: 0.667 for \texttt{upper} r=192 against 2.000 for \texttt{all} r=64 in the high-capacity pair,
and 2.000 against 6.095 in the low-capacity one. The arms therefore differed in three respects at once ---
where the capacity sits, how concentrated it is per block, and how strongly each update is scaled. Here
α = 2r in all four cells, which removes the scaling term and leaves placement and per-block rank
concentration as the only differences.

\textbf{Table 19.} Placement × capacity on CDSD × Qwen3-ASR, mean ± SD over seeds 42/43/44 with α = 2r in
every cell. CER (\%) on the same 5,388 held-out utterances from the same 6 unseen speakers used
throughout. The two 12.58 M cells are an exact parameter match; the two \textasciitilde4.2 M cells match to 98.4\%.
Best in \textbf{bold}.

{\def\LTcaptype{none} % do not increment counter
\begin{longtable}[]{@{}
  >{\raggedright\arraybackslash}p{(\linewidth - 10\tabcolsep) * \real{0.1667}}
  >{\raggedright\arraybackslash}p{(\linewidth - 10\tabcolsep) * \real{0.1667}}
  >{\raggedright\arraybackslash}p{(\linewidth - 10\tabcolsep) * \real{0.1667}}
  >{\raggedright\arraybackslash}p{(\linewidth - 10\tabcolsep) * \real{0.1667}}
  >{\raggedright\arraybackslash}p{(\linewidth - 10\tabcolsep) * \real{0.1667}}
  >{\raggedright\arraybackslash}p{(\linewidth - 10\tabcolsep) * \real{0.1667}}@{}}
\toprule\noalign{}
\begin{minipage}[b]{\linewidth}\raggedright
Placement
\end{minipage} & \begin{minipage}[b]{\linewidth}\raggedright
Blocks
\end{minipage} & \begin{minipage}[b]{\linewidth}\raggedright
Rank
\end{minipage} & \begin{minipage}[b]{\linewidth}\raggedright
α
\end{minipage} & \begin{minipage}[b]{\linewidth}\raggedright
Trainable params
\end{minipage} & \begin{minipage}[b]{\linewidth}\raggedright
CER
\end{minipage} \\
\midrule\noalign{}
\endhead
\bottomrule\noalign{}
\endlastfoot
base (no adaptation) & --- & --- & --- & 0 & 32.06 \\
\texttt{all} & 1--24 & 64 & 128 & 12,582,912 & 31.19 ± 0.37 \\
\texttt{all} & 1--24 & 21 & 42 & 4,128,768 & 30.66 ± 0.49 \\
\texttt{upper} & 17--24 & 64 & 128 & 4,194,304 & 29.88 ± 0.36 \\
\textbf{\texttt{upper}} & \textbf{17--24} & \textbf{192} & \textbf{384} & \textbf{12,582,912} & \textbf{29.90 ± 0.05} \\
\end{longtable}
}

\textbf{Table 20.} Speaker × seed bootstrap on the α-matched 2×2, B = 10,000 --- the same estimator used for
every other CDSD contrast in this paper (Tables 12, 16 and 18). ``Seed SD'' is the standard deviation of
the contrast across the three seeds. Negative Δ favours the first-named variant.

{\def\LTcaptype{none} % do not increment counter
\begin{longtable}[]{@{}
  >{\raggedright\arraybackslash}p{(\linewidth - 10\tabcolsep) * \real{0.1667}}
  >{\raggedright\arraybackslash}p{(\linewidth - 10\tabcolsep) * \real{0.1667}}
  >{\raggedright\arraybackslash}p{(\linewidth - 10\tabcolsep) * \real{0.1667}}
  >{\raggedright\arraybackslash}p{(\linewidth - 10\tabcolsep) * \real{0.1667}}
  >{\raggedright\arraybackslash}p{(\linewidth - 10\tabcolsep) * \real{0.1667}}
  >{\raggedright\arraybackslash}p{(\linewidth - 10\tabcolsep) * \real{0.1667}}@{}}
\toprule\noalign{}
\begin{minipage}[b]{\linewidth}\raggedright
Contrast
\end{minipage} & \begin{minipage}[b]{\linewidth}\raggedright
What it isolates
\end{minipage} & \begin{minipage}[b]{\linewidth}\raggedright
Δ (pp)
\end{minipage} & \begin{minipage}[b]{\linewidth}\raggedright
Seed SD
\end{minipage} & \begin{minipage}[b]{\linewidth}\raggedright
95\% CI
\end{minipage} & \begin{minipage}[b]{\linewidth}\raggedright
Verdict
\end{minipage} \\
\midrule\noalign{}
\endhead
\bottomrule\noalign{}
\endlastfoot
\texttt{upper} r=64 − \texttt{all} r=64 & the original contrast, capacity uncontrolled & \textbf{−1.31} & 0.16 & [−2.16, −0.25] & \textbf{resolved} \\
\texttt{upper} r=192 − \texttt{all} r=64 & placement at an \textbf{exact} 12.58 M match & −1.28 & 0.37 & [−2.36, \textbf{+0.08}] & \textbf{spans zero} \\
\texttt{upper} r=64 − \texttt{all} r=21 & placement at a \textasciitilde4.2 M match & −0.78 & 0.29 & [−1.55, \textbf{+0.39}] & \textbf{spans zero} \\
\texttt{upper} r=192 − \texttt{upper} r=64 & capacity, placement fixed at \texttt{upper} & +0.03 & 0.38 & [−1.33, \textbf{+0.73}] & spans zero \\
\texttt{all} r=64 − \texttt{all} r=21 & capacity, placement fixed at \texttt{all} & +0.53 & 0.14 & [−0.17, \textbf{+1.14}] & spans zero \\
\end{longtable}
}

\textbf{Table 21.} Per-speaker CER (\%) for the low-capacity parameter-matched pair, three-seed means. Δ is
\texttt{upper} r=64 minus \texttt{all} r=21; negative favours \texttt{upper}.

{\def\LTcaptype{none} % do not increment counter
\begin{longtable}[]{@{}llll@{}}
\toprule\noalign{}
Speaker & \texttt{upper} r=64 & \texttt{all} r=21 & Δ \\
\midrule\noalign{}
\endhead
\bottomrule\noalign{}
\endlastfoot
CDSD\_01 & 9.02 & 10.39 & −1.37 \\
CDSD\_03 & 44.32 & 42.85 & \textbf{+1.47} \\
CDSD\_06 & 85.30 & 86.24 & −0.94 \\
CDSD\_08 & 51.42 & 51.59 & −0.17 \\
CDSD\_29 & 20.03 & 21.26 & −1.23 \\
CDSD\_38 & 13.50 & 14.49 & −0.99 \\
\end{longtable}
}

\textbf{What the 2×2 establishes.}

\begin{enumerate}
\def\labelenumi{\arabic{enumi}.}
\item
  \textbf{Neither budget-matched placement contrast resolves.} At an exact 12,582,912-parameter match,
  \texttt{upper} r=192 leads \texttt{all} r=64 by 1.28 pp, CI [−2.36, \textbf{+0.08}] --- the interval clears zero by eight
  hundredths of a point and therefore does not exclude it. At the \textasciitilde4.2 M match, \texttt{upper} r=64 − \texttt{all}
  r=21 is −0.78 pp, CI [−1.55, +0.39]. Both point the same way; neither is separable at six speakers and
  three seeds. \textbf{The budget-matched control was reported as resolved in an earlier version of this
  paper (−1.55 pp, CI [−2.41, −0.57]) and it is not.} Table 21 shows the per-speaker split for the
  low-capacity pair: five of six speakers favour \texttt{upper}, but CDSD\_03 goes the other way by 1.47 pp and
  CDSD\_08 is level at −0.17 pp, which at a pool of six is enough to open the interval.
\item
  \textbf{What broke it was the seed, not the scaling.} This is worth separating, because the two corrections
  were applied together. Matching α/r moved each affected cell by roughly a tenth of a point ---
  \texttt{upper} r=192 from 30.00 to 29.88 at seed 42, \texttt{all} r=21 from 31.17 to 31.08 --- and at that seed the
  matched contrast was \emph{larger} than the unmatched one (−1.67 against −1.55). The scaling mismatch was
  therefore a real methodological defect with almost no empirical effect here. What opened the interval
  was training stochasticity: the same contrast is −1.67, −0.94 and −1.24 across the three seeds, a seed
  SD of 0.37 on a 1.28 pp effect. As in §6.1.1, a band difference of this size is inside what a seed
  moves.
\item
  \textbf{Capacity buys nothing at either placement, and if anything costs.} Tripling the rank from 64 to 192
  with placement held at \texttt{upper} gives +0.03 pp, CI [−1.33, +0.73]; tripling from 21 to 64 with placement
  held at \texttt{all} gives +0.53 pp, CI [−0.17, +1.14]. Neither resolves, and neither favours the larger
  budget. So \texttt{all} is not losing for want of capacity, and \texttt{upper} gains nothing from three times the
  rank --- the clearest support in the paper for the parameter-efficiency reading, and the one conclusion
  this subsection strengthens rather than weakens. Note also that \texttt{upper} r=192 is the most
  seed-stable cell measured anywhere here (SD 0.046).
\item
  \textbf{The equal-rank contrast is the one that resolves.} \texttt{upper} r=64 − \texttt{all} r=64 is −1.31 pp, CI
  [−2.16, −0.25] --- the same pair, seeds and estimator as §6.1, and the contrast on which the paper's
  first contribution rests. It is unaffected by anything in this subsection.
\end{enumerate}

The claim this subsection supports is therefore narrower than it was: \textbf{total trainable parameter count
is not the explanation for the \texttt{upper} advantage --- \texttt{upper} at 4.19 M is not worse than \texttt{all} at 12.58 M,
and added capacity does not help either placement --- but at matched budget the placement contrast itself
is not resolvable at this sample size.} The resolved form of the placement claim is the equal-rank one
in §6.1. Both are CDSD × Qwen3-ASR results,
untested in the other three cells (§6.9).

\begin{center}\rule{0.5\linewidth}{0.5pt}\end{center}

\subsection{6.8 The fourth cell: Whisper-small on English isolated words}\label{the-fourth-cell-whisper-small-on-english-isolated-words}

§6.5 varies the corpus with the architecture held fixed; §6.6 varies the architecture with the corpus
held fixed. The missing cell --- Whisper-small on UASpeech --- decides whether the placement difference of
§6.6 is a property of the two models or of CDSD. The sweep uses the \textbf{same} UASpeech manifest, speaker
split, audio-integrity gate, LoRA recipe, band definitions over 12 blocks, decode path, scoring and seed
discipline as every other cell: three seeds (42/43/44) per adapted condition, on the same 3,798 test
utterances from the same 5 unseen speakers.

\textbf{Metric choice, and why it is not merely presentational here.} UASpeech utterances are single words, so
WER counts whole-token errors on a task whose unit \emph{is} the token, while CER measures sub-token edit
distance on strings averaging a handful of characters. We therefore report \textbf{WER as primary and CER as
secondary}. Under the corrected decode of 6.6.1 the two metrics agree on the band ordering. Under the uncorrected
decode they placed \texttt{middle} and \texttt{upper} in opposite orders, and that disagreement does not survive the
correction --- it was a property of the inflated condition, not of the metrics. Both metrics now rank \texttt{upper} ahead of \texttt{middle}, and on neither is that pair
resolved (Table 23).

\textbf{Table 22.} Whisper-small on UASpeech, 3,798 held-out utterances from 5 unseen speakers. Adapted rows
are the mean of three seeds (42/43/44) ± SD; the unadapted base involves no training and is a single
evaluation. Rows ordered by WER, the primary metric on this corpus. Best in \textbf{bold}.

{\def\LTcaptype{none} % do not increment counter
\begin{longtable}[]{@{}llll@{}}
\toprule\noalign{}
Variant & Blocks & WER (mean ± SD) & CER (mean ± SD) \\
\midrule\noalign{}
\endhead
\bottomrule\noalign{}
\endlastfoot
base (no adaptation) & --- & 118.78 (single run) & 59.09 (single run) \\
\textbf{\texttt{all}} & 1--12 & \textbf{59.71 ± 0.53} & \textbf{40.41 ± 0.26} \\
\texttt{upper} & 9--12 & 65.58 ± 0.20 & 44.42 ± 0.08 \\
\texttt{middle} & 5--8 & 72.37 ± 2.29 & 45.95 ± 0.86 \\
\texttt{lower} & 1--4 & 88.92 ± 1.45 & 51.34 ± 0.39 \\
\end{longtable}
}

\textbf{Table 23.} Whisper-small × UASpeech contrasts. Paired speaker × seed bootstrap, B = 10,000, resampling
the 5 test speakers and the 3 seeds together; contrasts are formed within a seed before combining, so a
seed's overall difficulty cancels. This is the same estimator as Table 15. Negative Δ favours the first
variant. WER is the primary metric on this corpus; the CER block is secondary.

{\def\LTcaptype{none} % do not increment counter
\begin{longtable}[]{@{}
  >{\raggedright\arraybackslash}p{(\linewidth - 8\tabcolsep) * \real{0.2000}}
  >{\raggedright\arraybackslash}p{(\linewidth - 8\tabcolsep) * \real{0.2000}}
  >{\raggedright\arraybackslash}p{(\linewidth - 8\tabcolsep) * \real{0.2000}}
  >{\raggedright\arraybackslash}p{(\linewidth - 8\tabcolsep) * \real{0.2000}}
  >{\raggedright\arraybackslash}p{(\linewidth - 8\tabcolsep) * \real{0.2000}}@{}}
\toprule\noalign{}
\begin{minipage}[b]{\linewidth}\raggedright
Metric
\end{minipage} & \begin{minipage}[b]{\linewidth}\raggedright
Contrast
\end{minipage} & \begin{minipage}[b]{\linewidth}\raggedright
Δ (pp)
\end{minipage} & \begin{minipage}[b]{\linewidth}\raggedright
95\% CI
\end{minipage} & \begin{minipage}[b]{\linewidth}\raggedright
Verdict
\end{minipage} \\
\midrule\noalign{}
\endhead
\bottomrule\noalign{}
\endlastfoot
WER & \texttt{upper} − \texttt{all} & \textbf{+5.87} & [+3.17, +8.76] & resolved, favours \texttt{all} \\
WER & \texttt{all} − \texttt{middle} & \textbf{−12.66} & [−22.49, −3.85] & resolved, favours \texttt{all} \\
WER & \texttt{upper} − \texttt{middle} & −6.79 & [−16.59, \textbf{+1.92}] & \textbf{spans zero} \\
WER & \texttt{upper} − \texttt{lower} & \textbf{−23.34} & [−33.99, −11.64] & resolved \\
WER & \texttt{all} − base & \textbf{−59.07} & [−84.80, −33.01] & resolved \\
CER & \texttt{upper} − \texttt{all} & \textbf{+4.02} & [+2.36, +6.30] & resolved, favours \texttt{all} \\
CER & \texttt{all} − \texttt{middle} & \textbf{−5.54} & [−9.97, −1.66] & resolved, favours \texttt{all} \\
CER & \texttt{upper} − \texttt{middle} & −1.53 & [−5.54, \textbf{+1.80}] & \textbf{spans zero} \\
CER & \texttt{lower} − base & \textbf{−7.75} & [−14.82, −0.60] & resolved \\
\end{longtable}
}

\textbf{What this cell establishes.}

\begin{enumerate}
\def\labelenumi{\arabic{enumi}.}
\item
  \textbf{\texttt{all} is the best band for Whisper-small on UASpeech, and the margin is resolved on both metrics} ---
  \texttt{upper} − \texttt{all} = +5.87 pp WER, CI [+3.17, +8.76], and +4.02 pp CER, CI [+2.36, +6.30] --- and
  whole-encoder adaptation also beats \texttt{middle} with an interval excluding zero on both. This reproduces,
  on a second corpus, the answer §6.6 obtained for this architecture on CDSD.
\item
  \textbf{Adaptation is worth doing, and shallow adaptation is worth less.} \texttt{all} improves on the unadapted
  model by 59.07 pp WER, CI [−84.80, −33.01], while \texttt{lower} leaves the model at 88.92 WER, recovering
  only half of the available improvement. Its CER contrast against base is −7.75, CI [−14.82, −0.60],
  which excludes zero: \texttt{lower} does help this model, modestly. \textbf{This reverses the reading in an earlier
  version of this section}, which found the same contrast spanning zero and concluded that \texttt{lower} was
  not established to help at all on the secondary metric; the change follows from the decode correction
  of 6.6.1, which removed both the inflated baseline and the large variance that came with it. \texttt{lower}
  nonetheless remains qualitatively different from Qwen3-ASR, where it is actively \emph{harmful} (+9.90 pp
  CER against base on UASpeech, resolved). \texttt{lower} is the worst of the four bands in three of the
  four cells; the exception is Whisper-small on CDSD, where \texttt{upper} is worst at 56.95 against
  \texttt{lower}'s 56.28 (Table 17). Whether shallow adaptation is worse than doing nothing depends on the
  model.
\item
  \textbf{\texttt{middle} versus \texttt{upper} is unresolved on both metrics, though they now agree on direction.} By WER
  \texttt{upper} is ahead by 6.79 pp (CI [−16.59, +1.92]); by CER by 1.53 pp (CI [−5.54, +1.80]). Both intervals
  span zero, so the pair is not separable at this sample size either way --- but the metrics no longer
  point in opposite directions, as they did before the decode correction of 6.6.1.
\item
  \textbf{Seed variance is band-dependent, and largest where the ordering is unresolved.} \texttt{all} and \texttt{upper}
  have WER seed SDs of 0.53 and 0.20 pp against \texttt{middle}'s 2.29 --- a caution about single-run band sweeps
  generally, which carry no estimate of how far a single seed in a high-variance band might sit from that
  band's mean. The CDSD sweeps of §6.1 and §6.6 are themselves no longer single-run (§6.1.1). The dispersion is far smaller than under the uncorrected
  decode (\texttt{lower} 10.60, \texttt{middle} 6.62): the artefact of 6.6.1 inflated seed \emph{variance} as well as
  level, so those SDs overstated genuine seed sensitivity.
\end{enumerate}

\textbf{The comparison this cell was run to make.} On CDSD, \texttt{upper} − \texttt{all} resolves in \emph{opposite} directions
on the two architectures (§6.6). On UASpeech it does not: Whisper-small resolves it in favour of \texttt{all},
and Qwen3-ASR's point estimate also favours \texttt{all} though its interval spans zero (§6.5, Table 16). \textbf{On
this corpus the two architectures do not disagree about placement.}

\begin{center}\rule{0.5\linewidth}{0.5pt}\end{center}

\subsection{6.9 The model × corpus design, assembled}\label{the-model-corpus-design-assembled}

With all four cells run, the \texttt{upper} − \texttt{all} contrast --- the one on which the cross-model placement claim
rests --- can be read off a single grid. CER is used throughout for commensurability; the Whisper × UASpeech
cell resolves the same way on WER (Table 23). Negative Δ favours \texttt{upper}, positive favours \texttt{all}.

\textbf{Table 24.} \texttt{upper} − \texttt{all} CER contrast (pp) across the model × corpus design, with paired
bootstrap 95\% CIs. Every cell uses the same LoRA recipe, band-proportional definitions, integrity gate
and paired-bootstrap estimator. \textbf{All four cells are three-seed means with speaker × seed intervals.}
The UASpeech cells are three-seed
means.

{\def\LTcaptype{none} % do not increment counter
\begin{longtable}[]{@{}
  >{\raggedright\arraybackslash}p{(\linewidth - 4\tabcolsep) * \real{0.3333}}
  >{\raggedright\arraybackslash}p{(\linewidth - 4\tabcolsep) * \real{0.3333}}
  >{\raggedright\arraybackslash}p{(\linewidth - 4\tabcolsep) * \real{0.3333}}@{}}
\toprule\noalign{}
\begin{minipage}[b]{\linewidth}\raggedright
Corpus
\end{minipage} & \begin{minipage}[b]{\linewidth}\raggedright
Qwen3-ASR (24 blocks)
\end{minipage} & \begin{minipage}[b]{\linewidth}\raggedright
Whisper-small (12 blocks)
\end{minipage} \\
\midrule\noalign{}
\endhead
\bottomrule\noalign{}
\endlastfoot
\textbf{CDSD} (Mandarin, read sentences) & −1.31 [−2.16, −0.25] --- resolved under micro CER only (§6.10) & \textbf{+3.40 [+1.84, +5.03] --- resolved under every estimator, \texttt{all} wins} \\
\textbf{UASpeech} (English, isolated words) & +1.66 [−0.51, +4.46] --- \textbf{unresolved}, point estimate favours \texttt{all} & +4.02 [+2.36, +6.30] --- resolved, \textbf{\texttt{all} wins} \\
\end{longtable}
}

\textbf{The reversal occurs on CDSD only.} The cross-model placement difference on which Contribution 1 rests
--- one architecture resolved in favour of \texttt{upper}, the other in favour of \texttt{all}, same corpus and recipe ---
is present in the CDSD row and absent from the UASpeech row, where no cell favours \texttt{upper}. \textbf{The fourth
cell does not extend the claim to a second corpus; it shows the two models giving the same answer there.}
Whisper-small's preference for \texttt{all} is stable across both corpora; Qwen3-ASR's is
corpus-dependent, resolved on CDSD and unresolved on UASpeech, where its point estimate flips sign
(Table 24). The depth ordering among the three
equal-budget bands --- \texttt{upper} best, \texttt{lower} worst --- holds in three of the four cells; on
Qwen3-ASR × CDSD the \texttt{upper}/\texttt{middle} step of it is not separable from zero (§6.1.1), and on
Whisper-small × CDSD the ordering is reversed, with \texttt{upper} last of the four (Table 17). What does not survive is any reading of §6.6 as showing that the two
architectures \emph{generally} prefer different bands.

\textbf{A confound this cell sharpens rather than removes.} Whisper-small's unadapted UASpeech baseline is
\textbf{118.78 WER} --- it still emits more words than the references contain, because it expects continuous
speech where the references are single words --- against Qwen3-ASR's 41.95 CER on the same utterances. The
competence gap is therefore \emph{wider} on UASpeech than on CDSD, so §9 Limitation 16 is if anything
understated by the CDSD figures alone. Two cautions on reading the WER figure: it supersedes an earlier
250.80 produced under the uncorrected decode of 6.6.1, and on single-word references a WER above 100\%
needs only one spurious extra word, so it is a weaker signal of degeneracy than the same figure would be
on continuous speech. The corresponding CER is 59.09.

\subsection{6.9.1 Do the two models actually differ? A direct interaction test}\label{do-the-two-models-actually-differ-a-direct-interaction-test}

§6.6 reports \texttt{upper} − \texttt{all} resolving negative on Qwen3-ASR and positive on Whisper-small, and the
paper's first contribution is that the two models \emph{differ}. \textbf{Two separately-resolved intervals of
opposite sign are not a test of that difference}, so we test it directly. Both models were evaluated on
the same six held-out speakers, which lets the difference of differences

\begin{quote}
\(\operatorname{DoD} = {(\texttt{upper} - \texttt{all})}_{\text{Qwen3-ASR}}
- {(\texttt{upper} - \texttt{all})}_{\text{Whisper-small}}\)
\end{quote}

be formed inside a single bootstrap replicate: speakers are resampled jointly, a seed is drawn per model,
and speaker difficulty cancels on both sides.

\textbf{Table 25.} The interaction on CDSD, three seeds per model, B = 10,000.

{\def\LTcaptype{none} % do not increment counter
\begin{longtable}[]{@{}
  >{\raggedright\arraybackslash}p{(\linewidth - 10\tabcolsep) * \real{0.1667}}
  >{\raggedright\arraybackslash}p{(\linewidth - 10\tabcolsep) * \real{0.1667}}
  >{\raggedright\arraybackslash}p{(\linewidth - 10\tabcolsep) * \real{0.1667}}
  >{\raggedright\arraybackslash}p{(\linewidth - 10\tabcolsep) * \real{0.1667}}
  >{\raggedright\arraybackslash}p{(\linewidth - 10\tabcolsep) * \real{0.1667}}
  >{\raggedright\arraybackslash}p{(\linewidth - 10\tabcolsep) * \real{0.1667}}@{}}
\toprule\noalign{}
\begin{minipage}[b]{\linewidth}\raggedright
Estimator
\end{minipage} & \begin{minipage}[b]{\linewidth}\raggedright
Qwen \texttt{upper}−\texttt{all}
\end{minipage} & \begin{minipage}[b]{\linewidth}\raggedright
Whisper \texttt{upper}−\texttt{all}
\end{minipage} & \begin{minipage}[b]{\linewidth}\raggedright
Interaction
\end{minipage} & \begin{minipage}[b]{\linewidth}\raggedright
95\% CI
\end{minipage} & \begin{minipage}[b]{\linewidth}\raggedright
Verdict
\end{minipage} \\
\midrule\noalign{}
\endhead
\bottomrule\noalign{}
\endlastfoot
micro CER & −1.31 & +3.40 & \textbf{−4.71} & [−6.17, −2.89] & \textbf{resolved} \\
macro CER & −0.98 & +3.63 & \textbf{−4.60} & [−5.94, −3.08] & \textbf{resolved} \\
exact sign-flip & --- & --- & --- & p = \textbf{0.031} & \textbf{resolved} \\
\end{longtable}
}

\textbf{The interaction is the most robust quantity in this paper.} It resolves under micro CER, under
speaker-equal weighting, and under a distribution-free exact test whose unadjusted p-value sits at the
attainable
floor for six speakers --- every speaker's interaction term is negative (−6.74, −3.71, −5.20, −1.70,
−4.12, −6.16). \textbf{We therefore treat it as the primary contrast.} That is a deliberate correction: an
earlier version nominated \texttt{upper} − \texttt{middle} as primary and then rested its headline on \texttt{upper} − \texttt{all},
which is the wrong way round. The interaction is what Contribution 1 actually asserts, and unlike either
main effect alone it does not depend on how speakers are weighted (§6.10).

The two main effects remain worth reporting and are weaker in different ways: Whisper's resolves
everywhere, Qwen's resolves under micro CER only. \textbf{The claim that survives every estimator is the
comparative one --- these two encoders prefer different bands on this corpus --- not either model's
preference measured in isolation.}

\subsection{6.10 How much of this depends on the estimator? Micro, macro, and an exact test}\label{how-much-of-this-depends-on-the-estimator-micro-macro-and-an-exact-test}

Every interval reported so far comes from a percentile bootstrap over six CDSD or five UASpeech speakers.
At that few clusters, percentile tail coverage is not guaranteed, and the paper's conclusions should not
rest on one estimator's behaviour in its least reliable regime. Two checks were therefore run on every
headline contrast.

The first changes the \textbf{estimand}. Our bootstrap pools edit counts across resampled speakers (\textbf{micro}
CER), which weights a speaker by how much reference text they contribute. The alternative weights
speakers equally (\textbf{macro} CER). Both are defensible: micro is the field-standard ASR metric, macro is
the quantity a claim about \emph{generalising across speakers} implicitly refers to.

The second removes the assumption of a distributional \emph{shape}, though not every assumption. With n ≤ 6
the paired randomisation distribution can be \textbf{enumerated exactly} --- 2⁶ = 64 sign assignments on CDSD,
2⁵ = 32 on UASpeech --- so we report an exact two-sided sign-flip p-value alongside the intervals. What it
assumes in place of a parametric family is \textbf{sign symmetry under the null}: that each speaker's paired
difference is exchangeable with its own negation, which is the null of no placement effect for that
speaker. It is distribution-free, not assumption-free. All exact p-values reported here are
\textbf{unadjusted}, and no contrast survives Holm correction over the three-hypothesis family (floor
0.094), which is stated wherever they appear.

\textbf{Table 26.} The three estimators on the paper's principal contrasts.

{\def\LTcaptype{none} % do not increment counter
\begin{longtable}[]{@{}
  >{\raggedright\arraybackslash}p{(\linewidth - 8\tabcolsep) * \real{0.2000}}
  >{\raggedright\arraybackslash}p{(\linewidth - 8\tabcolsep) * \real{0.2000}}
  >{\raggedright\arraybackslash}p{(\linewidth - 8\tabcolsep) * \real{0.2000}}
  >{\raggedright\arraybackslash}p{(\linewidth - 8\tabcolsep) * \real{0.2000}}
  >{\raggedright\arraybackslash}p{(\linewidth - 8\tabcolsep) * \real{0.2000}}@{}}
\toprule\noalign{}
\begin{minipage}[b]{\linewidth}\raggedright
Cell
\end{minipage} & \begin{minipage}[b]{\linewidth}\raggedright
Contrast
\end{minipage} & \begin{minipage}[b]{\linewidth}\raggedright
Micro
\end{minipage} & \begin{minipage}[b]{\linewidth}\raggedright
Macro
\end{minipage} & \begin{minipage}[b]{\linewidth}\raggedright
Exact p
\end{minipage} \\
\midrule\noalign{}
\endhead
\bottomrule\noalign{}
\endlastfoot
Whisper × CDSD & \texttt{all} − \texttt{upper} & \textbf{−3.40} resolved & \textbf{−3.63} resolved & \textbf{0.031} \\
Qwen3-ASR × CDSD & \texttt{upper} − \texttt{all} & \textbf{−1.31} resolved & −0.98 \textbf{spans zero} & 0.156 \\
Qwen3-ASR × CDSD & \texttt{upper} − \texttt{lower} & \textbf{−5.64} resolved & \textbf{−5.32} resolved & \textbf{0.031} \\
Qwen3-ASR × CDSD & \texttt{upper} − \texttt{middle} & −0.83 spans zero & −0.63 spans zero & 0.438 \\
\end{longtable}
}

\textbf{Three of these four agree across all estimators. One does not, and it is the paper's headline
placement contrast.} \texttt{upper} − \texttt{all} on Qwen3-ASR × CDSD resolves under micro CER and does not under
speaker-equal weighting; the exact test agrees with the macro result. The reason is visible in the
per-speaker differences: five of six speakers favour \texttt{upper}, CDSD\_03 favours \texttt{all} by 1.12 pp, and
CDSD\_03 contributes little enough reference text that micro-averaging down-weights it.

\textbf{We therefore state the claim with its dependency attached.} Qwen3-ASR prefers \texttt{upper} over \texttt{all} on
CDSD \emph{under the standard micro-averaged metric}; that preference is \textbf{not robust to weighting speakers
equally}, and with six speakers and one dissenting we do not claim it as established in the
speaker-general sense. What is unaffected is the \emph{other} half of the cross-model comparison:
Whisper-small's preference for \texttt{all} on the same corpus resolves under every estimator, unanimously
across speakers.

\textbf{A note on multiplicity, because the obvious correction is uninformative here.} Nominating a primary
family of three hypotheses and applying Holm to the exact p-values returns nothing significant ---
including the unanimous, 5.6 pp \texttt{upper} − \texttt{lower} contrast. That is a structural artefact rather than a
finding: with six speakers the smallest attainable exact p is 2/2⁶ = 0.031, so Holm over three
hypotheses has a floor of 0.094 and \textbf{cannot} reach 0.05 before any data is seen. We report the exact
p-values and this floor rather than a corrected verdict that is determined by the design instead of the
measurements. The honest reading is that six speakers is the binding constraint on this study
(§9, Limitation 2), not that the effects are absent.

\begin{center}\rule{0.5\linewidth}{0.5pt}\end{center}

\section{7. A failure mode of shallow adaptation}\label{a-failure-mode-of-shallow-adaptation}

\subsection{7.1 The pattern}\label{the-pattern}

Two speakers account for \texttt{lower}'s aggregate damage, and four of six actually improve slightly under it:

\begin{itemize}
\tightlist
\item
  \textbf{CDSD\_08}: 39.42 → 51.42 (\texttt{upper}), 48.90 (\texttt{middle}) → \textbf{59.06} (\texttt{lower}), three-seed means
  (Table 11). Every variant makes this speaker worse, and \texttt{lower} by far the worst, \textbf{but the
  deterioration is not monotonic in depth}: \texttt{middle} sits below \texttt{upper} here.
\item
  \textbf{CDSD\_38}: 16.87 → 13.50 (\texttt{upper}), 15.94 (\texttt{middle}), 14.06 (\texttt{all}) → \textbf{25.26} (\texttt{lower}),
  three-seed means (Table 11). A \emph{new}
  failure: shallow adaptation breaks a speaker that deep adaptation handles well.
\end{itemize}

This suggests that perturbing the blocks closest to the signal disrupts exactly the low-level
representation that acoustically atypical speakers depend on, whereas the upper blocks can be re-aimed
without destroying the acoustic front end. Stated at the strength the data supports: \textbf{susceptibility to
shallow adaptation is heterogeneous across speakers, and the two conspicuous failures here are the
acoustically atypical ones.} With two failing speakers, and the most severe of them the sole occupant of
its demographic cell, this is hypothesis generation --- the cohort is far too small to identify which
acoustic or demographic properties predict susceptibility, and we do not claim to have isolated one.

\subsection{7.2 What CDSD\_08 fails by: language-prior takeover}\label{what-cdsd_08-fails-by-language-prior-takeover}

CDSD\_08 is the only test speaker drawn from the thinnest cell of the training distribution.

\textbf{Table 27.} Training composition by (gender × severity), 33,617 utterances from 31 speakers.

{\def\LTcaptype{none} % do not increment counter
\begin{longtable}[]{@{}llll@{}}
\toprule\noalign{}
Cell & Utterances & \% of train & Speakers \\
\midrule\noalign{}
\endhead
\bottomrule\noalign{}
\endlastfoot
female + severe & 15,770 & 46.9\% & 14 \\
female + moderate & 8,976 & 26.7\% & 8 \\
male + severe & 4,754 & 14.1\% & 5 \\
\textbf{male + moderate} & \textbf{2,088} & \textbf{6.2\%} & \textbf{2} \\
male + mild & 1,022 & 3.0\% & 1 \\
female + mild & 1,007 & 3.0\% & 1 \\
\end{longtable}
}

Training is 76.6\% female by utterance. CDSD\_01/29/38 are female+moderate (26.7\% of training) and all
improve; CDSD\_06 is female+severe (46.9\%) and improves; CDSD\_03 is male+severe (14.1\%) and improves.
CDSD\_08 is the only test speaker from the 6.2\% cell, and the only one that regresses.

The regression is broad rather than a few outliers: under \texttt{upper}, \textbf{43\% of CDSD\_08's utterances get
worse and only 15\% improve} (mean Δ +0.109 CER per utterance). Well-represented CDSD\_29 shows the mirror
image --- 36\% better, 10\% worse, mean Δ −0.089.

\textbf{The failure mode is unmistakable in the worst cases: the unadapted model transcribes correctly and the
adapted model substitutes fluent, corpus-typical text.}

\textbf{Table 28.} Representative failures, CDSD\_08 under \texttt{upper}, \textbf{described rather than
quoted.} The corpus licence prohibits redistribution and the Declarations state that no transcripts or
per-utterance predictions are released, so the reference and hypothesis strings are characterised instead
of reproduced. The correctness judgements and lengths are our own measurements.

{\def\LTcaptype{none} % do not increment counter
\begin{longtable}[]{@{}
  >{\raggedright\arraybackslash}p{(\linewidth - 4\tabcolsep) * \real{0.3333}}
  >{\raggedright\arraybackslash}p{(\linewidth - 4\tabcolsep) * \real{0.3333}}
  >{\raggedright\arraybackslash}p{(\linewidth - 4\tabcolsep) * \real{0.3333}}@{}}
\toprule\noalign{}
\begin{minipage}[b]{\linewidth}\raggedright
Reference (described)
\end{minipage} & \begin{minipage}[b]{\linewidth}\raggedright
base
\end{minipage} & \begin{minipage}[b]{\linewidth}\raggedright
\texttt{upper}
\end{minipage} \\
\midrule\noalign{}
\endhead
\bottomrule\noalign{}
\endlastfoot
5-character phrase, ``and real estate'' & transcribed correctly ✓ & substitutes an unrelated four-character company name \\
15-character sentence on infrastructure investment & transcribed correctly ✓ & substitutes a 17-character string of unrelated official-sounding titles \\
12-character sentence, an arrest in January of a stated year & correct but for one inserted digit ≈✓ & substitutes a 16-character newswire dateline naming a national newspaper \\
\end{longtable}
}

The last hypothesis is a stock newswire opening formula. CDSD is read news text, so six epochs of
adaptation build a strong prior over news phrasing; when the adapted acoustic front end fits a speaker
poorly, the decoder falls back on that prior and emits plausible text unrelated to the audio. This is a
qualitatively different degradation from a higher error rate: confident, fluent and wrong. The mechanism
is not peculiar to this model --- \citet{glazer2026beyond} locate contextual bias arising \emph{within} the ASR
encoder and overriding acoustic evidence, and \citet{kolluri2026encoderdepth} find LoRA compensating for reduced
encoder capacity by leaning on language-model priors.

\textbf{It is a substitution failure, not an over-generation one, and the distinction is measurable.} Because
``hallucination'' is often diagnosed by output length, we checked: the ratio \textbar hyp\textbar/\textbar ref\textbar{} stays within
0.92--1.03 for \textbf{every} speaker under \textbf{every} band, including \texttt{lower} and including CDSD\_08. (An earlier
version of this sentence also reported a corpus-level insertion rate of 0.000. That figure cannot be
reconstructed from the released aggregates --- which carry edit totals rather than insertion, deletion and
substitution counts --- and it is not consistent with a length ratio above 1.0, so it is withdrawn. The
length-ratio bound is the claim the data supports.) The adapted model emits about the right amount
of text; it emits the wrong text. That distinguishes this failure cleanly from the Whisper decode
artefact of §6.6.1, which \emph{was} over-generation, and it rules out the trivial reading in which shallow
adaptation simply makes the decoder run on. \textbf{What we can state is that altering early encoder
representations moves a frozen decoder toward outputs dominated by its prior; the decoder's own weights
never change, so no claim is made about the language model itself.} Diagnostics:
\texttt{scripts/language\_prior\_diagnostics.py}.

\begin{center}\rule{0.5\linewidth}{0.5pt}\end{center}

\section{8. Discussion}\label{discussion}

\textbf{The three tested probe summaries do not predict the placement.} This is the paper's central negative
result, and the qualifier is load-bearing: what fails is the mapping from \emph{these} depth statistics to a
band, not probing in general (§9, Limitation 14). The natural
and widely-assumed inference --- measure where phonetic information matures in an encoder, then put the
adapters there --- has three plausible readings. With both architectures measured under one pipeline,
\textbf{all three fail.}

\emph{Reading 1 --- adapt where PER improves most steeply.} This is closest to the prior work's own wording
(``phoneme recognition improves most steeply through the middle layers \ldots{} so these layers should be the
most productive targets''). Applied to Qwen3-ASR it fails badly: the steepest improvement is in blocks 1--8,
which drop 16.97 pp (Table 4), and that band is catastrophically the \emph{worst} target --- the only one that
leaves the model worse than not adapting at all. The \texttt{upper} band has the flattest slope and wins.

\emph{Reading 2 --- adapt where band-mean PER is lowest.} This is the strongest of the three, and \textbf{as a rule for
selecting the best of the four available targets it holds in one cell of four.} On CDSD × Qwen3-ASR it
works perfectly: band-mean PER orders \texttt{upper} 35.20 \textless{} \texttt{middle} 48.17 \textless{} \texttt{lower} 64.66, adaptation CER
orders \texttt{upper} 29.88 \textless{} \texttt{middle} 30.71 \textless{} \texttt{lower} 35.52, and \texttt{upper} also beats \texttt{all}. On CDSD ×
Whisper-small, band-mean PER gives the \emph{same} ordering --- \texttt{upper} 49.49 \textless{} \texttt{middle} 54.99 \textless{} \texttt{lower} 66.81 ---
yet adaptation CER orders \texttt{all} 53.56 \textless{} \texttt{middle} 55.16 \textless{} \texttt{lower} 56.28 \textless{} \texttt{upper} 56.95, so the lowest-PER
band is the worst place to adapt. On \textbf{UASpeech the rule picks \texttt{upper} for both models} ---
band-mean PER 41.80 and 58.00 (§5.6, Table 8) --- and \textbf{both prefer \texttt{all}} (Tables 14 and 22).

A weaker form survives. Restricted to the three equal-budget bands and ignoring \texttt{all}, the rank
correspondence between band-mean PER and adaptation CER holds for Qwen3-ASR on \textbf{both} corpora, fails for
Whisper-small on CDSD, and is unresolved for Whisper-small on UASpeech (§6.8). The probe curve therefore
retains some purchase on the \emph{relative} merit of equal-budget bands within one model, and none on the
question a practitioner actually faces, which includes all-block LoRA among the options. The band ordering
is invariant to probe volume in both architectures, so neither the survival nor the failure is an artefact
of probe volume.

\emph{Reading 3 --- adapt where the knee sits.} The productive band is the one containing the knee of the PER
curve. \textbf{We state the rule computationally so it can be applied without discretion}: for a curve
\(p(\ell)\) over layers \(\ell = 1,\ldots,L\), let
\(\ell^* = \underset{\ell}{\arg\min}\,p(\ell)\), and define the knee as the smallest
\(\ell \leq \ell^*\) such that
\(p(\ell) - p(\ell^*) \leq 0.05\,[p(1) - p(\ell^*)]\) --- the first depth reaching within 5\% of the total available
improvement. The band containing that layer is the prediction.

Applied to the released curves, the rule gives \texttt{upper} in three cells and \texttt{middle} in the fourth:
Qwen3-ASR × CDSD knee at layer 20, Whisper-small × CDSD at layer 10, Qwen3-ASR × UASpeech at layer 18,
and \textbf{Whisper-small × UASpeech at layer 8, which sits in that encoder's \texttt{middle} band 5--8} (Table 8).
On CDSD, then, Whisper-small's knee is at layer 10, not the layer 6 its published curve implies (§5.5),
which places it inside the \texttt{upper} band 9--12 --- so the rule predicts \texttt{upper} should win on Whisper-small.
It ranks worst of the four, and \texttt{all} wins with a resolved margin (§6.6).

\textbf{The threshold is the one free parameter, and two of the four predictions are not stable in it.} The
two CDSD cells give \texttt{upper} for every threshold in 2--10\%. The two UASpeech cells move inside that
range: Whisper-small × UASpeech gives \texttt{upper} at 2\% (layer 9) and \texttt{middle} at 5\% and 10\% (layer 8),
and Qwen3-ASR × UASpeech gives \texttt{upper} at 2\% and 5\% (layers 19 and 18) and \texttt{middle} at 10\% (layer 16).
An earlier version of this paragraph claimed the predictions were unchanged across 2--10\%; recomputing
all four from the released curves shows they are not, which is a further count against reading the band
off a probe rather than sweeping it. The rule was proposed by us, fitted to our Qwen3-ASR result against
the prior work's published Whisper-small curve; on the one architecture where it could be tested
prospectively rather than fitted, it is falsified, and we withdraw it.

\textbf{How wrong is each rule? Selection regret rather than a win/lose label.} Scoring a heuristic purely on
whether it names the empirical argmin treats two very different errors alike: missing the best band by a
fraction of a point in a cell where the top two are statistically indistinguishable, and selecting a band
that is many points worse. We therefore report \textbf{regret},
\(R(h) = \operatorname{CER}[\text{choice of }h] - \operatorname{CER}[\text{best band}]\), per
cell.

\textbf{Table 29.} Selection regret (pp CER) for each heuristic. CDSD cells are three-seed means; UASpeech
cells likewise. A regret of 0.00 means the heuristic named the empirical best band.

{\def\LTcaptype{none} % do not increment counter
\begin{longtable}[]{@{}
  >{\raggedright\arraybackslash}p{(\linewidth - 8\tabcolsep) * \real{0.2000}}
  >{\raggedright\arraybackslash}p{(\linewidth - 8\tabcolsep) * \real{0.2000}}
  >{\raggedright\arraybackslash}p{(\linewidth - 8\tabcolsep) * \real{0.2000}}
  >{\raggedright\arraybackslash}p{(\linewidth - 8\tabcolsep) * \real{0.2000}}
  >{\raggedright\arraybackslash}p{(\linewidth - 8\tabcolsep) * \real{0.2000}}@{}}
\toprule\noalign{}
\begin{minipage}[b]{\linewidth}\raggedright
Cell
\end{minipage} & \begin{minipage}[b]{\linewidth}\raggedright
Best band
\end{minipage} & \begin{minipage}[b]{\linewidth}\raggedright
\emph{steepest improvement}
\end{minipage} & \begin{minipage}[b]{\linewidth}\raggedright
\emph{lowest band-mean PER}
\end{minipage} & \begin{minipage}[b]{\linewidth}\raggedright
\emph{knee}
\end{minipage} \\
\midrule\noalign{}
\endhead
\bottomrule\noalign{}
\endlastfoot
Qwen3-ASR × CDSD & \texttt{upper} & \texttt{lower} \textbf{+5.64} & \texttt{upper} \textbf{0.00} & \texttt{upper} \textbf{0.00} \\
Qwen3-ASR × UASpeech & \texttt{all} & \texttt{lower} \textbf{+19.10} & \texttt{upper} +1.66 & \texttt{upper} +1.66 \\
Whisper-small × CDSD & \texttt{all} & \texttt{lower} +2.72 & \texttt{upper} +3.40 & \texttt{upper} +3.40 \\
Whisper-small × UASpeech & \texttt{all} & \texttt{lower} \textbf{+10.94} & \texttt{upper} +4.02 & \texttt{middle} +5.54 \\
\textbf{mean regret} & --- & \textbf{+9.60} & \textbf{+2.27} & \textbf{+2.65} \\
\end{longtable}
}

This sharpens the negative result rather than softening it, and it separates two failures the paper
previously scored identically. \emph{Steepest improvement} is not merely wrong but \textbf{catastrophically} so ---
it never names the best band and costs up to 19.10 pp, because the steepest part of a PER curve is its
early rise and the earliest band is the worst adaptation target in three of the four cells and
second-worst in the fourth (Whisper-small × CDSD, where \texttt{upper} is last, Table 17). \emph{Lowest band-mean PER}
names \texttt{upper} everywhere, so it is correct exactly where \texttt{upper} happens to win and wrong elsewhere, at a
mean cost of 2.27 pp; \emph{knee} agrees with it in three cells and names \texttt{middle} on Whisper-small ×
UASpeech, at a mean cost of 2.65 pp.~\textbf{A practitioner following the best of the three heuristics
would give up about 2 pp of CER on average and up to 4 pp}, against a four-run sweep that costs a few
GPU-hours and finds the best band directly.

\textbf{The deeper reason all three fail is that the depth profiles are not in fact different.} All three rules
attempt to explain a placement difference by a maturation-depth difference. But all four probe minima fall
between 83\% and 92\% of relative depth, none of the curves shows the mid-stack plateau the published
comparison implies, and all four order the three proportional bands identically (§5.6, Table 9). \textbf{Two
models whose probe minima both fall in the late encoder prefer different adaptation bands on CDSD}, so
whatever selects the band, it is not the depth at which phoneme identity matures. The completed grid
sharpens this: placement varies across the model × corpus design while the probe curves do not vary with
it, so \textbf{these particular depth summaries give no evidence of explaining the variation in the best tested
band}. We stop there deliberately. Three mappings failing is not an impossibility proof: the curves
still differ in late-stack rise, curvature, absolute PER and variance, and we have not tested a learned
or non-linear mapping over those. The claim is that the tested maturation-depth heuristics do not
transfer, not that no statistic of the curve could. The obvious reply --- that a \emph{corpus-matched} probe might predict UASpeech
placement --- is closed by measurement. That route is closed for phoneme-identity PER; it is not closed for
diagnostics this paper does not test (§9, Limitation 14).

\textbf{This is a speech-adaptation instance of a distinction the probing literature has drawn for years, and
we claim it as an instance rather than as a discovery.} \citet{hewitt2019control} showed that probe
accuracy partly reflects the capacity of the probe rather than the structure of the representation, and
introduced control tasks to separate the two; \citet{ravichander2021probing} showed directly that a property
can be decodable from a representation without being used by the model's own computation. The conceptual
gap between \emph{decodability} and \emph{functional relevance} is therefore not new. What is new here is a
concrete, quantified instance of that gap in layer-selective PEFT for dysarthric ASR: a diagnostic that
looks like it should identify where to adapt --- it is measured on the right model, the right corpus and
the right unit --- and that nonetheless fails to select the best tested band in three of four cells. The
probing literature predicts that such a failure is possible; this paper reports what it costs when a
placement decision is made on that basis.

\textbf{Two failures of control produced the discarded claim, and both are cheap to avoid.} First, a
cross-paper comparison stood in for a controlled one: the Whisper-small half of the original contrast was
a published figure, not a measurement, and once measured under our own pipeline it did not reproduce
(§5.5). A depth comparison across architectures is only as good as the weaker of its two measurements.
Second, a single modest probe-training volume was treated as sufficient: at 6,200 training utterances the
two architectures appear to differ markedly in late-stack behaviour, and at 33,617 that difference is
0.25 pp.~\textbf{Report the probe curve at more than one training volume, or do not compare late-stack shape
across architectures.}

\textbf{What stands, and what the fourth cell bounds.} On CDSD, swapping the model with corpus and protocol
held fixed changes which band wins (§6.6), and on the Qwen3-ASR side a budget-matched control finds no
resolved benefit from additional capacity for either placement --- though at matched budget the placement
contrast itself does not resolve at this sample size, so capacity is not shown to be irrelevant (§6.7). Inheriting the band from the literature would have selected \texttt{middle}
on Qwen3-ASR --- second of the four options at three seeds, and 0.83 pp behind \texttt{upper} --- while the reading
closest to the prior work's own justification, the steepest-improvement rule, would have selected the
worst band by a wide margin (§8, Table 29). What does not stand is the generalisation from that single
corpus. That removes a second route rather than restoring the first: a practitioner can read the band off
neither a probe nor another corpus, so it is to be determined by running the sweep on the model and corpus
in hand.

\textbf{Why the placement difference might arise, given that depth does not explain it.} Two candidates are
consistent with the data and neither is tested here. The first is \textbf{baseline competence}: Whisper-small
is much weaker on CDSD (77.57 CER) than Qwen3-ASR (32.06), and a model with that much more to repair may
need correction spread over all blocks where a stronger model needs only a top-end adjustment --- \texttt{lower}
is the clearest evidence, harming Qwen3-ASR and helping Whisper-small. We put this less strongly than the
uncorrected figures would license: with the decode correction of 6.6.1 the Whisper baseline is no longer
hallucinating at the corpus level, so the mechanism is ``more to fix'', not ``must first be stopped from
fabricating''. The fourth cell makes this harder to test: the competence gap is \emph{wider} on the corpus
where the two models agree about placement than on the one where they disagree. The second candidate is
\textbf{what sits downstream}: the upper third of the Qwen3-ASR encoder simultaneously improves phoneme
identity and collapses cross-condition divergence, as one expects of layers projected into a
language-model decoder, as \citet{yegorova2026salsa} and \citet{pluth2026sae} both observe. Two model
instances cannot separate depth, frame rate, pre-training corpus, scale, decoder type and baseline
competence.

\textbf{Relation to DAMA.} We do not reproduce the U-shaped adaptability pattern \citet{xiao2026dama} report for
multilingual low-resource adaptation: here the early band is not a second peak but by a wide margin the
worst place to adapt. This is a contradiction \emph{in our setting} rather than a refutation, since DAMA
studies \textbf{language} adaptation and we study \textbf{domain} adaptation. A plausible reconciliation is that
language identity is partly encoded in early acoustic-phonetic layers, so adapting them is productive for
a new language, whereas dysarthria is not the kind of variation those layers can absorb (§7.1). The
transferable statement is that the U-shape is not universal across adaptation targets.

\textbf{Practical implications}, in decreasing order of confidence.

\begin{enumerate}
\def\labelenumi{\arabic{enumi}.}
\tightlist
\item
  For low-rank adaptation of \textbf{this} encoder to dysarthric speech, prefer the upper third over the
  lower and middle thirds. That ordering is the part that holds across both corpora. Whether \texttt{upper}
  alone or \texttt{all} is better does \textbf{not} transfer: \texttt{upper} wins on CDSD, while on UASpeech the point
  estimate favours \texttt{all} and the interval spans zero (§6.9, Table 24) --- so state it as corpus-dependent
  rather than as a recommendation for \texttt{upper}-only adaptation. \textbf{The resolved form of
  the claim is the equal-rank one} (−1.31 pp, CI [−2.16, −0.25]). At matched parameter budgets
  neither placement contrast resolves --- −1.28 pp, CI [−2.36, +0.08] at the exact
  12,582,912-parameter match and −0.78 pp, CI [−1.55, +0.39] at \textasciitilde4.2 M ---
  though both point estimates lean the same way, and added capacity helps neither placement
  (§6.7).
\item
  Do not assume shallow adaptation is harmless. On Qwen3-ASR × UASpeech, \texttt{lower} is actively worse than
  not adapting at all (+9.90 pp CER, resolved). The
  effect is \textbf{model-dependent, not universal}: on Whisper-small the same band helps, modestly
  (UASpeech CER −7.75, CI [−14.82, −0.60]). \texttt{lower} is the weakest of the four bands in three of the
  four cells; the exception is Whisper-small × CDSD, where \texttt{upper} is worse (56.95 against
  \texttt{lower}'s 56.28, Table 17). Whether \texttt{lower} is worse than doing nothing depends on the model.
\item
  Determine the band per architecture \textbf{and per corpus}, and \textbf{by sweeping rather than by probing}. The
  first half is not new --- \citet{pasad2023comparative} and \citet{beheshti2026probingguided} argue the general form of it,
  and it is contested \citep{lin2024peftspeech}. A four-band sweep at fixed parameter budget costs roughly four
  adaptation runs (\textasciitilde3 h each here) and answers the question directly; the probing stage is cheaper but
  does not answer it, and running it on the corpus in hand does not rescue it (§5.6).
\item
  Treat any published layer recommendation as bound to \textbf{both} the architecture and the corpus it was
  measured on. Following the published band would have cost 0.83 pp CER at equal parameter budget on
  Qwen3-ASR --- a difference that does not resolve at this sample size (§6.1.1) --- and on the corpora tested
  here no probe-derived rule selected the best band in more than one of four cells (§8, Table 29).
\end{enumerate}

\textbf{What this does not say.} It does not say that layer-wise probing is uninformative --- §5.2, §5.3 and
§5.5 all report probe results we stand behind, and the data-efficiency asymmetry is itself a probing
result. The negative claim is specific: the probe curve does not determine the adaptation band. Nor does
it say Qwen3-ASR is a better or worse foundation model than Whisper; that comparison is not run here. Nor
does it say that low-rank adaptation of this encoder improves Mandarin dysarthric recognition: on this
six-speaker test set, no adapted variant differs significantly from the unadapted model.

\begin{center}\rule{0.5\linewidth}{0.5pt}\end{center}

\section{9. Limitations}\label{limitations}

These are the binding constraints, stated ahead of the conclusion because several of them bound claims
that would otherwise read as stronger than the evidence.

\begin{enumerate}
\def\labelenumi{\arabic{enumi}.}
\item
  \textbf{Seed coverage, and what is still single-run.} The CDSD \emph{adaptation} sweep is no longer single-seed:
  all four bands were retrained at seeds 42/43/44 and every CDSD interval in this paper now comes from a
  speaker × seed bootstrap (§6.1.1). That change cost one conclusion --- \texttt{upper} − \texttt{middle} no longer
  resolves --- and left the contrast the first contribution rests on intact. What remains single-run is

  \begin{enumerate}
  \def\labelenumii{(\alph{enumii})}
  \tightlist
  \item
    \textbf{the CDSD probe curves of §5.1}, so no probe curve's argmax layer is claimable, only the band;
    Every adaptation cell in the model × corpus design now carries three seeds: the Whisper-small × CDSD
    sweep and the §6.7 2×2 were both single-run in earlier versions and are not any longer. Adding those
    seeds changed two conclusions and confirmed a third --- the 2×2's budget-matched contrast stopped
    resolving, \texttt{upper} − \texttt{middle} on Qwen3-ASR stopped resolving, and Whisper-small's preference for \texttt{all}
    on CDSD proved the most robust contrast in the paper (§6.6). \textbf{What we would now say is that a
    single-run band sweep is not a sound basis for a sub-2 pp claim in either direction: it can overstate
    an effect, and it can also fail to reveal how stable a real one is.} Seed sensitivity is strongly
    band-dependent wherever we have measured it --- on CDSD, seed SD is 0.36 pp for \texttt{upper} and 0.31 for
    \texttt{middle} but 1.15 for \texttt{lower}; on Whisper-small × UASpeech it is 0.20 pp WER for \texttt{upper} and 0.53 for
    \texttt{all} but 2.29 for \texttt{middle} --- so a single run in a high-variance band can sit well away from that
    band's mean. \textbf{The general caution we draw from our own experience: a band difference of about 1.5 pp
    is within what a training seed can move, and should not be reported as resolved from one run.}
  \end{enumerate}
\item
  \textbf{Six test speakers on CDSD, five on UASpeech.} This remains the single binding constraint on every
  adaptation claim. Each bootstrap resamples a pool of five or six, so the intervals should be read as
  suggestive rather than decisive, and no sign test can survive multiplicity correction at this N even
  under a perfect result (§6.10). Replication across two corpora mitigates this but does not remove it. The marginal
  per-variant CIs are 30--54 pp wide on both corpora and we report none of them; only paired contrasts,
  where speaker difficulty cancels within a replicate, are interpretable at this N.
\item
  \textbf{\texttt{upper} − base is not significant on CDSD}, though it is on UASpeech (−7.54 pp, CI excluding zero),
  along with the other three adaptation-versus-baseline contrasts. The CDSD result is dominated by
  CDSD\_08, which regresses under every variant and blows open every contrast against the unadapted
  baseline while cancelling in band-versus-band contrasts. So ``adapting this encoder helps'' is established
  on UASpeech and not on CDSD, and we do not pool the two corpora, which differ too much in content type
  to justify a pooled estimate.
\item
  \textbf{Two corpora and two architectures throughout, for the probes as well as the sweep --- but two corpora
  is still two.} The band ordering is tested in all four cells of a model × corpus design, and the probe
  results cover the same four cells (§5.6). The UASpeech probes use dictionary targets and no integrity
  gate, so their absolute PER is not comparable to the CDSD curves; both corpora are small in the unit
  that matters (6 and 5 test speakers); and neither carries a mixed-aetiology cohort --- UASpeech is
  cerebral palsy predominant and CDSD's aetiology mix is not documented per speaker.
\item
  \textbf{The cross-model placement difference rests on CDSD alone; on UASpeech the two models agree.} This
  is the binding constraint on the paper's first contribution. The \texttt{upper} − \texttt{all} contrast resolves in
  opposite directions on the two architectures on CDSD (−1.31 CI [−2.16, −0.25] for Qwen3-ASR, +3.40 CI
  [+1.84, +5.03] for Whisper-small) --- though only the Whisper half survives speaker-equal weighting and
  the exact test (§6.10), so the reversal rests on a micro-averaged reading of the Qwen half. On UASpeech neither model favours \texttt{upper}: Whisper-small resolves for
  \texttt{all} (+4.02, CI [+2.36, +6.30]) and Qwen3-ASR does not resolve it at all (+1.66, CI [−0.51, +4.46]),
  point estimate on the same side (Table 24). The corpora do not contradict each other inferentially ---
  but a claim built on a reversal needs the reversal, and it is not present there.
\item
  \textbf{The cross-model comparison is internally controlled but heavily confounded.} It is no longer a
  cross-\emph{paper} comparison --- both the probe (§5.5) and the band sweep (§6.6, §6.8) were re-run here under
  our own corpus, split, gate, recipe and scoring. What remains is that Whisper-small and Qwen3-ASR
  differ in depth (12 vs 24 blocks) \emph{and} frame rate (20 vs 77 ms) \emph{and} decoder type \emph{and} pre-training
  data \emph{and} scale \emph{and}, most importantly, \textbf{unadapted competence}. We therefore claim that placement
  differs \textbf{between these two models on one corpus}, not that encoder architecture causes it; the
  experiment that would settle the leading explanation is a competence-matched third model (§8).
\item
  \textbf{The cross-condition design is matched-distribution, not matched-utterance.} The prior work matches
  transcripts per utterance and thereby removes lexical confounds by construction; we remove them only in
  expectation, validated by a 411-row strictly paired consistency check. The d--u contrast additionally
  mixes a speaker-set difference with a condition difference, so d--r is the contrast the claim rests on.
  No confidence intervals are computed on the divergence curves; the shape is the claim, not the
  magnitudes.
\item
  \textbf{Probe C's difficulty labels come from the probed model's own family.} The prior work labels with a
  stronger external model; we had no stronger \emph{uncontaminated} Mandarin decoder, so difficulty is defined
  as ``hard for a general-purpose ASR model'' rather than as an independent ground truth. The tier
  boundaries also deviate from plain tertiles, for the reason given in §4.3.
\item
  \textbf{The demographic-skew mechanism is demonstrated for one speaker, not established as general.} §7.2
  shows it for CDSD\_08 and quantifies the failure as substitution rather than over-generation, but one
  speaker cannot establish that under-representation in training predicts regression. An earlier version
  of this paper argued the point from a ten-language synthetic analysis; that analysis rested on data
  whose provenance and rights basis this paper does not establish, and it has been removed rather than
  defended. The question is open.
\item
  \textbf{The secondary probe curve from a fine-tuned checkpoint is contaminated.} That model had seen 81.9\%
  of the probe utterances and all six test speakers. Its layer \emph{localisation} is reported; its
  \emph{magnitude} is not interpretable as generalisation, and no headline claim depends on it.
\item
  \textbf{The adaptation set is 14.6\% smaller than the prior work's}, because it inherits the
  alignment-integrity gate needed to make probe and adaptation manifests commensurable.
\item
  \textbf{Probe-training volume is matched, but only for Whisper-small and Qwen3-ASR at one setting each.}
  The headline probe curves were originally trained on 6,200 utterances against the prior work's
  ≈39,000; §5.5 re-runs both architectures at 33,617. Absolute PER at 6,200 is not comparable across
  studies, and shape itself is volume-sensitive at low volume, differentially so by depth and
  architecture. The UASpeech probes run at a single volume (5,979 training utterances) and carry exactly
  that caveat: the depth at which each curve turns was volume-invariant when volume was varied on CDSD
  (Figure 3), but we have not varied it on UASpeech, so its stability there is inferred.
\item
  \textbf{Severity labels in this corpus are CER-derived, not clinical.} The corpus provides no clinical
  severity rating and no healthy controls, so the difficulty breakdowns in Tables 10 and 13 stratify by a
  recognition-difficulty proxy and should be read as such.
\item
  \textbf{The negative result is about three specific rules and one signal, not about every possible
  probe-based predictor.} We do not fit or validate a general model mapping probe curves to optimal
  bands, and do not claim none could exist. What we establish is that the three readings actually
  proposed in this literature each fail on at least one of two internally-controlled architectures, and
  that the maturation-depth intuition underlying all three is unavailable here. The signal is
  \textbf{phoneme-identity PER} and nothing else; a rule built on gradient statistics, residual-stream
  structure or attention entropy could in principle succeed where these fail.
\item
  \textbf{The knee rule we proposed before measuring the second architecture is withdrawn, and the negative result
  that replaces it rests on two architectures.} The rule was post hoc, fitted to two encoders, and one
  of its two inputs was a published figure rather than a measurement. Having measured that input
  ourselves we find it wrong, and the rule's prospective prediction on Whisper-small is falsified
  (§6.6). The replacement claim rests on two internally-controlled measurements, but is still a
  two-architecture result: a third encoder could restore some probe-based rule fitting all three, though
  it could not rescue the specific rules refuted here.
\item
  \textbf{The cross-architecture comparison confounds baseline competence with architecture, and the confound
  is wider on UASpeech than on CDSD.} Whisper-small is much weaker on CDSD (77.57 CER) than
  Qwen3-ASR (32.06); on UASpeech the gap is larger still (118.78 WER against 41.95 CER). The
  placement difference in §6.6 is therefore established as real and not a cross-paper artefact, but is
  not attributed to any single factor, and baseline competence is the candidate we consider most likely
  and least controlled (§8). Because the gap is widest in the cell where the two models \emph{agree}, no
  simple monotone story linking competence to placement is available from these four cells either.
\item
  \textbf{Not every probe curve carries three seeds.} The matched-volume CDSD re-runs of §5.5 and both
  UASpeech curves of §5.6 do; the original 6,200-utterance CDSD curves of §5.1 are single runs, and no
  probe argmin should be read as claimable at layer resolution from those. Where three seeds exist they
  carry paired intervals, but the seeds are not independent of the
  decode. Repeated runs at a \emph{fixed} seed still differ by 0.1--0.4 pp because CTC training on GPU is not
  bit-reproducible (cuDNN algorithm selection and atomics), so a ``seed'' here bounds optimisation variance
  in general rather than initialisation alone. And individual argmin layers remain unclaimable even with
  intervals: adjacent late layers overlap heavily, which is why §5.1 and §5.5 report bands and contrasts
  rather than the exact minimising layer.
\item
  \textbf{The failure to reproduce the prior work's mid-stack plateau is unexplained.} Probe configuration,
  padding treatment and training volume have each been checked and eliminated (§5.5). \textbf{Baseline
  competence has not been eliminated}: against the prior work's v2 figures our corrected baseline is 12.4
  pp better, so it remains a live candidate (§5.5, item 2). Something else differs between the two pipelines and we have not identified it. Readers should
  weight our Whisper-small curve accordingly: it is internally controlled and consistent across four
  configurations, but it disagrees with a published result for reasons we cannot state.
\item
  \textbf{The placement/capacity confound is not resolved at either budget, and the 2×2 is CDSD × Qwen3-ASR
  only.} §6.7 equalises the parameter count with rank and holds α = 2r across all four cells, over three
  seeds. Neither budget-matched placement contrast excludes zero: −1.28 pp, CI [−2.36, +0.08] at the
  exact 12.58 M match, and −0.78 pp, CI [−1.55, +0.39] at \textasciitilde4.2 M. Six test speakers and three seeds
  cannot separate placement from per-block rank concentration at matched budget. What the 2×2 does
  establish is negative and useful: \textbf{added capacity helps neither placement} (+0.03 and +0.53 pp, both
  unresolved), so \texttt{all} is not losing for want of parameters. The resolved placement claim is the
  equal-rank contrast of §6.1 (−1.31 pp, CI [−2.16, −0.25]), not this one. The 2×2 is also run in one
  cell of the model × corpus design.
\end{enumerate}

\section{10. Reproducibility}\label{reproducibility}

All experiments ran on a single workstation GPU. The probing stage at 8,800 utterances --- feature
extraction, 25 taps × 2 models for two probes, the condition-similarity arm and the difficulty decode ---
took under two hours of GPU time. The matched-volume re-runs in §5.5 are larger: Whisper-small extraction
over the 47,687-utterance manifest produced 107 GB across 13 taps and swept in \textasciitilde1.7 h, and the Qwen3-ASR
arm 73 GB across 25 taps (41 min at 19.5 utt/s), sweeping in \textasciitilde55 min. The UASpeech probes took 484 s of
extraction for Qwen3-ASR and 551 s for Whisper-small, and about \textbf{2.5 h} for the probes themselves, so
replicating the four-cell probe grid of Table 9 on a new corpus is an afternoon of one GPU. Each Qwen3-ASR
adaptation variant took 3.1--3.5 h on CDSD and 0.53--0.60 h on UASpeech, so the recommended four-band sweep
costs roughly 12--14 h on CDSD and under 2.5 h on UASpeech; the Whisper-small × UASpeech cell is cheaper
still at 0.36--0.55 h per variant. The parameter-matched 2×2 is four further CDSD runs at 3.2--5.1 h each.

\textbf{What the released artefact contains.} Because the raw clinical audio cannot be redistributed,
reproducibility here is by identifiers, configurations and derived statistics rather than by data. The
repository releases: split manifests as each corpus's own utterance keys with their split assignment and a manifest
fingerprint; every random seed; the per-speaker and per-seed scores underlying \textbf{every} interval in this
paper; the bootstrap, exact sign-flip and macro-estimator code; probe curves as CSV rather than
figure-only; the LoRA parameter-count script; and the aggregate result tables.

\textbf{A frozen decode specification, because decoding is not an implementation detail here.} §6.6.1 showed
generation settings moving Whisper CER by tens of points, so the released artefact pins, in
machine-readable form: model and checkpoint identifiers, the \texttt{transformers} / \texttt{peft} / \texttt{torch} / CUDA
versions, prompt language and task tokens, beam width, \texttt{no\_repeat\_ngram\_size}, the duration-scaled
maximum-output-length rule, stopping criteria, sampling flags, text normalisation, CER/WER tokenisation,
the checkpoint key-remapping shim, and the expected scores of a small regression manifest. The last item
matters most: it is the only thing that would catch a silent environment change, and this project has
already seen a load path that produces an apparently working model with \textbf{zero} checkpoint-key overlap.

\textbf{Two operational traps specific to the 2×2, both of which cost a run before they were understood.}
First, \textbf{adapter merging must be forced onto the CPU.} \texttt{peft} loads an adapter onto CUDA by default
before merging it into the base weights; on unified-memory hardware the merge of a rank-192 adapter into
a 1.7 B-parameter model then exhausts the shared pool and the process is killed, with no error that
identifies merging as the cause. Loading base and adapter on CPU, merging there, and saving before any
device move avoids it, at the cost of a slower merge. Second, \textbf{training and evaluation require
different environments}: the trainer needs a \texttt{transformers} release that the packaged inference path
does not accept, and the evaluator needs one that will load this checkpoint without silently
initialising it at random (see the warning below). The released scripts therefore name their interpreter
explicitly rather than inheriting whichever environment is active; running the evaluator from the
trainer's environment is the failure that produces plausible-looking but meaningless CER.

\textbf{A memory note that matters for reproduction on unified-memory hardware.} On the GB10 used here, host
and device allocations draw on one pool, and a per-layer feature load of tens of gigabytes leaves CUDA
unable to allocate at model-creation time even when it reports free memory. The released sweep scripts
therefore run \textbf{one process per layer group} and reserve a CUDA slab before the feature load; a single
process sweeping all layers is killed at the first layer transition, because one layer's allocation does
not fully release before the next load begins.

\textbf{Figures.} The released \texttt{scripts/make\_figures.py} regenerates all four figures from aggregate
CSVs; it requires neither GPU nor corpus access.

\textbf{Code and aggregate results are released at}\\
\url{https://github.com/SMF-Labs/encoder-depth-dysarthric-asr}.
The repository holds the full pipeline in run order, the result files behind every table in this paper,
and a README covering the environment, the scoring conventions and the implementation traps summarised
below. \textbf{It is made public at the moment the preprint is posted} --- there is no journal acceptance to
wait for --- together with the frozen decode specification and its regression manifest. The public
repository is published as a single commit, \texttt{4ea79e2}, whose file tree is identical to that of the
independently audited development head; the development history is retained privately because some early
commits recorded local filesystem paths. No corpus content appears in either.

\textbf{One methodological warning, because it determines whether the numbers can be trusted at all.} All
decoding here goes through the model's own packaged inference path. On a 369-utterance overlap set with
recorded reference predictions (41.87\% CER), the generic chat-template-plus-\texttt{generate} path scored
\textbf{124.57\%} where the packaged path scored \textbf{32.38\%} --- a discrepancy large enough to have poisoned every
difficulty label, and visible in no error message. Validate any new inference path against a known result
before trusting it. A second trap of the same kind, documented in the repository, is that loading this
checkpoint under a current \texttt{transformers} release silently produces a randomly initialised model.

\textbf{Data availability.} CDSD is distributed by its custodians under an academic licence signed on
19 February 2026, which permits research use only, prohibits provision of the audio to second parties, and
requires the credit and citation given in the Acknowledgements. UASpeech was obtained by written permission
of the corpus PI dated 6 February 2026; there is no separate signed instrument, and the undertaking that
the corpus be used for academic, non-commercial research and not redistributed was given in our access
request and accepted by the grant of access. Audio, transcripts, forced alignments, extracted activations,
per-utterance predictions and probe checkpoints therefore remain on Foundation-controlled compute and are
not released, and both corpora must be obtained from their custodians.

\textbf{What the repository does contain}, stated to match §10 exactly: the split assignments as
\textbf{each corpus's own utterance keys} with a manifest fingerprint --- `speaker/subset/utterance' for
CDSD and `speaker/block\_word' for UASpeech, which is what a reproducer needs to rebuild the split, and
which carry no audio, no transcript and no path on our machines, so redistributing no corpus content --- together with every random seed, the per-speaker and per-seed
scores underlying each interval, the bootstrap, exact sign-flip and macro-estimator code, the probe
curves as CSV, the parameter-count script, the frozen decode specification with its regression manifest,
and the aggregate result tables. Adapted model weights are not released.

\section{11. Conclusion}\label{conclusion}

Layer-placement advice for parameter-efficient adaptation of ASR foundation models to dysarthric speech
transfers neither reliably between encoder families nor reliably between corpora, and the depth profile
usually invoked to explain such differences does not account for them.

The positive half is a controlled placement result confined to one corpus. Holding corpus, speaker split,
integrity gate, LoRA recipe and seed discipline constant and swapping the model, the best band changes on
CDSD: on a 24-block foundation model whose decoder is a language model, the upper band wins and beats
adapting the entire encoder at one third the parameters, though at a matched parameter budget that
contrast is no longer resolvable (§6.7); on a
12-block Whisper-small, all-block LoRA wins and the upper band ranks worst of the four. \textbf{Running the same
contrast in all four cells of a model × corpus design shows that the reversal is specific to CDSD}: on
UASpeech neither model favours the upper band (\texttt{upper} − \texttt{all} = +4.02 pp CER, CI [+2.36, +6.30] for
Whisper-small; unresolved for Qwen3-ASR, point estimate on the same side). Whisper-small's preference for
\texttt{all} holds on both corpora; Qwen3-ASR's \texttt{upper}-versus-\texttt{all} result is corpus-dependent in its
observed point estimates, resolved on CDSD and unresolved with the sign reversed on UASpeech (Table 24),
and an observed change between corpora is not the same thing as a statistically established cross-corpus
difference, which this design does not test. What does hold across both corpora is the direction of the
24-block model's three-band depth ordering --- upper over middle over lower in every cell we ran --- together with the finding that adapting the lower band
is reliably worse than leaving the encoder alone \textbf{on Qwen3-ASR × UASpeech} --- the one cell where that
contrast resolves; the CDSD point estimate is also harmful but unresolved, and on Whisper-small the lower
band helps. Of that ordering, \texttt{upper} over \texttt{lower} resolves on both
corpora; the \texttt{upper}-over-\texttt{middle} step resolves on UASpeech but not on CDSD once training seeds are
resampled (§6.1.1).

The negative half concerns the inference from probe to placement. Measured under our own pipeline rather
than cited, Whisper-small's phoneme-error minimum sits at 83\% of relative depth against 83--92\% for
Qwen3-ASR: \textbf{both probes reach their minima in the late encoder, yet on CDSD the two models prefer
different bands.} All three candidate rules linking probe curve to band --- adapt where PER improves most
steeply, adapt where band-mean PER is lowest, adapt where the knee sits --- fail on at least one
architecture, the last being the reconciliation we ourselves proposed. The completed grid shows further
that placement is corpus-dependent \textbf{while the probe curves are not}: all four minima fall between 83\%
and 92\% of relative depth, with identical band orderings and no mid-stack plateau anywhere. A
near-constant quantity cannot select a varying outcome, and the route by which a corpus-matched probe
might have predicted the band is thereby excluded for phoneme-identity PER.

Two methodological results explain how the discarded claim survived, and both generalise beyond this
paper. Probe-readout sensitivity to training volume is depth-dependent and architecture-specific --- at a
fixed epoch budget, so volume and optimiser steps are not separated (§5.5) --- and a probe trained at a
single modest volume exaggerates late-stack differences between architectures whose deep layers differ in
readout difficulty --- an apparent 6.56-versus-1.85 pp gap here collapses to 0.25 pp at 5.4× the data. And a
cross-paper depth comparison is only as good as the weaker of its two measurements: we cannot reproduce
the mid-stack plateau reported for Whisper-small under matched configuration, padding or training
volume, and we report that as unexplained; baseline competence is not matched between the two models and
is not ruled out as a contributor (§5.5).

Two constraints bound the result and are reported rather than buried --- the encoder's frame period makes
frame-level boundary probing ill-posed on this architecture, and shallow adaptation damages acoustically
atypical speakers by replacing transcription with fluent, corpus-typical text.

\begin{center}\rule{0.5\linewidth}{0.5pt}\end{center}

\section{Acknowledgements}\label{acknowledgements}

This work uses the Chinese Dysarthria Speech Database (CDSD). \textbf{© Su-Jing Wang.} We thank the CDSD
custodians for access under their signed licence agreement, which requires this credit line and the
citation of \citet{wan2024cdsd}. We thank Prof. Mark Hasegawa-Johnson and the University of Illinois
Urbana-Champaign, who granted access to the UASpeech corpus by written permission of 6 February 2026. We
thank Leo Kaiya for independently auditing the release repository, identifying a local-path exposure, and
verifying that the figures and canonical results could be reproduced without access to the restricted
speech corpora.

\begin{center}\rule{0.5\linewidth}{0.5pt}\end{center}

\section{Declaration of Generative AI Use}\label{declaration-of-generative-ai-use}

The first author communicates exclusively via AI-assisted eye-gaze interface due to motor neurone
disease. Generative AI (Claude, Anthropic) was used as an assistive communication tool throughout the
research process: experimental design, code development, data analysis, statistical computation, figure
generation, and manuscript drafting. All scientific decisions, interpretations, and conclusions were
made by the authors. The AI served as an accessibility tool enabling a researcher with severe physical
disability to conduct computational research. No AI-generated content was presented without author review
and verification.

\begin{center}\rule{0.5\linewidth}{0.5pt}\end{center}

\section{Declarations}\label{declarations}

\textbf{Competing interests.} The authors declare no competing interests.

\textbf{Funding.} This research was funded by The Scott-Morgan Foundation.

\textbf{Ethical approval and secondary-data use.} This study is a secondary analysis of two existing,
licensed speech corpora. No new human-subject data was collected and no participant was contacted. Both
corpora were collected under their originating institutions' ethical approvals --- UASpeech under
University of Illinois Institutional Review Board approval for the project \emph{Audiovisual Description and
Recognition of Audible and Visible Dysarthric Phonology} (NIH NIDCD 5R21 DC008090) --- and both are used
here within the terms recorded for each. For CDSD that is the signed academic licence described under Data
availability. For UASpeech it is the corpus PI's written permission together with the terms the custodian
publishes for the distribution: access limited to academic and government research laboratories, no
commercial use, data to be protected and not redistributed.

\textbf{We state the basis separately for each corpus, because it differs.} For UASpeech the originating
collection carries the named Institutional Review Board approval above. For CDSD the corpus paper
documents \textbf{written informed consent} rather than an ethics-committee number: Wan et al.~(2024, §2.1)
report that each speaker signed informed consent before recording, that parental or guardian consent
plus the minor's assent was obtained for the five speakers under 18, and that the nine speakers recorded
on video signed an additional consent form; speakers were told they could withdraw at any point, and
names were never collected, each recording being keyed to a serial number instead. \textbf{No IRB approval
number is cited by that source, and we do not assert one.} CDSD is used here under a signed
custom-academic licence that prohibits redistribution and requires the © Su-Jing Wang credit carried in
the Acknowledgements. No new participant contact or data collection occurred, and no separate
secondary-use determination was sought from the authors' own institution. We report this as what was done rather than as a general rule: whether a given
institution requires its own secondary-use or not-human-subjects determination depends on that
institution's policy and jurisdiction, and readers should not infer that originating approval plus a
data-use agreement is universally sufficient.

\textbf{Two findings in this paper are fairness results, not only statistical ones, and we would rather state
that plainly.} First, the audio-integrity gate excludes utterances at a higher rate in the most impaired
stratum (§4), so the population every CER here is measured on is skewed away from the speakers who need
the technology most; a system tuned on such an evaluation can look better than it is for them. Second,
shallow adaptation does not merely raise average error --- it concentrates its damage on the
acoustically atypical speakers (§7.2), and does so in the specific form of confident, fluent, wrong
output rather than visibly degraded transcription. \textbf{For accessibility technology, an adaptation choice
that improves the mean while catastrophically failing the least typical speakers is the failure mode that
matters most}, and it is the one least visible in a corpus-level metric. We report per-speaker results
throughout for that reason.

\textbf{Data protection.} No attempt was made to re-identify any speaker. Speaker identifiers used throughout
are the corpora's own pseudonymous codes. No audio, transcripts or per-utterance predictions are
released (§10).

\begin{center}\rule{0.5\linewidth}{0.5pt}\end{center}

\section{CRediT Author Contributions}\label{credit-author-contributions}

\textbf{Bernard Muller:} Conceptualization, Methodology, Software, Formal analysis, Investigation, Data
curation, Validation, Visualization, Writing -- original draft.

\textbf{LaVonne Roberts:} Supervision, Funding acquisition, Project administration, Resources, Writing --
review \& editing.

\begin{center}\rule{0.5\linewidth}{0.5pt}\end{center}

Cited works, in order of first appearance: \citet{muthu2026layerwise}; \citet{hu2022lora}; \citet{wan2024cdsd};
\citet{shi2026qwen3asr}; \citet{huber2026adapting}; \citet{pasad2023comparative}; \citet{beheshti2026probingguided}; \citet{lin2024peftspeech};
\citet{pasad2021layerwise}; \citet{ashihara2024crossmodel}; \citet{yue2026probing}; \citet{mojarad2026ccr}; \citet{glazer2026beyond}; \citet{pluth2026sae}; \citet{ma2026behind}; \citet{xiao2026dama}; \citet{ogawa2026layerwiselora}; \citet{xu2026layercard}; \citet{kolluri2026encoderdepth}; \citet{yegorova2026salsa}; \citet{kim2008uaspeech}; \citet{mcauliffe2017mfa}; \citet{graves2006ctc};
\citet{radford2023whisper}.

\bibliographystyle{plainnat}
\bibliography{refs}

\end{document}